\documentclass[11pt]{article}
\usepackage[margin=1in]{geometry}
\usepackage{graphicx}
\usepackage[T1]{fontenc}
\usepackage{lmodern}
\usepackage{hyperref}
\usepackage{xcolor}
\usepackage{float}

\usepackage{parskip}
\renewcommand{\title}[1]{{\noindent\large\bfseries#1\medskip\\}}
\renewcommand{\author}[2]{{\noindent #1 \medskip\\ \noindent \small #2 \medskip\\}}
\usepackage{natbib}
\usepackage{tcolorbox}
\usepackage{amssymb}
\usepackage{amsmath}
\usepackage{longtable}
\usepackage{booktabs}
\usepackage{import}
\usepackage{setspace}
\usepackage{lineno}

\usepackage{amsthm}
\usepackage{tikz}       
\usepackage{multirow}

\usetikzlibrary{positioning,arrows.meta,calc}
\usepackage{enumitem}
\usepackage[font=normal,
            labelfont=bf,
            labelsep=newline,
            format=plain,
            justification=raggedright,
            singlelinecheck=false]{caption}
\usepackage{subcaption}
\newcommand{\apacap}[2]{\textit{#1}\\[2pt]\textit{Note.}\enskip #2}

\definecolor{uqblue}{HTML}{4682B4}   
\definecolor{syngreen}{HTML}{2E8B57} 
\definecolor{redorange}{HTML}{CD853F} 

\usepackage{booktabs,threeparttable}

\usepackage{placeins}

\usetikzlibrary{arrows.meta,positioning,fit,calc,matrix,decorations.pathreplacing}

\tikzset{
  var/.style={circle,draw,thick,minimum size=7mm,inner sep=0pt,font=\small},
  target/.style={var,very thick},
  dep/.style={-Latex,thick},
  conddep/.style={-Latex,thick,dashed},
  weakdep/.style={-Latex,thick,densely dotted},
  blocked/.style={thick,densely dashed,opacity=0.35},
  condset/.style={draw,thick,dashed,rounded corners=2mm,inner sep=3pt},
  embedset/.style={draw,thick,dash pattern=on 3pt off 2pt,rounded corners=2mm,inner sep=3pt},
  jbrace/.style={decorate,decoration={brace,amplitude=4pt},thick},
  paneltitle/.style={font=\bfseries\small,anchor=west},
  note/.style={font=\scriptsize,align=left},
}

\begin{document}


\title{Scalable partial information decomposition for symptom networks via supervised embeddings}
\author{
\small
Cillian Hourican,\textsuperscript{1}
Eric Dignum,\textsuperscript{1}
Rick Quax,\textsuperscript{1}
Debraj Roy,\textsuperscript{1}
}
{
\small
1. Computational Science Lab, Institute for Informatics, University of Amsterdam \\
}

\begin{abstract}
\noindent\textbf{Background.} Pairwise relationships among mental-health symptoms are routinely summarised as scalar edge weights, which cannot express whether two symptoms carry overlapping information about a third or information that appears only in combination. Partial information decomposition (PID) addresses this gap but is computationally intractable beyond a few sources.

\noindent\textbf{Method.} We introduce embedding-based PID (ePID), a scalable pipeline that compresses all non-focal symptoms into a low-cardinality discrete embedding and computes a tractable two-source PID, yielding source-unique, remainder-unique, redundant, and synergistic components for each ordered source-target pair. We benchmarked 13 candidate embeddings on synthetic Bayesian networks calibrated to PHQ-9 and on 83 real-world datasets across five PID measures.

\noindent\textbf{Results.} A supervised Agglomerative Conditional Information Bottleneck (ACIB) embedding recovered the reference decomposition most accurately of the 13 embeddings tested, and did so for every PID measure yielding non-negative atoms once four or more symptoms were compressed (synergy recovery $r=0.92$). The two instruments then diverged sharply. In PHQ-9 networks (UK Biobank, $N=154{,}291$; Xinxiang student sample, $N=24{,}292$) the surrounding symptom context carried most pairwise dependence through redundant and remainder-unique channels; synergy contributed 6 to 9\%, and no directed edge was synergy-dominated in either cohort. In the 28-item Interpersonal Reactivity Index, 45\% of source pairs were. The identical pipeline, applied without parameter changes, therefore returned opposite profiles for the two instruments, each consistent with how that instrument was constructed.

\noindent\textbf{Conclusions.} By separating overlapping from interaction-dependent information, ePID provides a scalable, model-agnostic complement to standard symptom-network methodology to distinguish redundant and synergistic contributions to observed correlations.
\end{abstract}

\bigskip
\noindent\textbf{Public Significance Statement}

\noindent
Symptom networks map how mental-health symptoms relate to one another. These relationships are complex, yet standard methods report only whether two symptoms are associated---not whether they carry the same information about a third symptom or only matter in combination. We developed ePID, a pipeline that splits each pairwise relationship into overlapping (redundant) information, shared by both symptoms, and joint-only (synergistic) information, available only when the two occur together. Applied to the PHQ-9 (depressive symptoms) and the Interpersonal Reactivity Index (empathy), ePID returned profiles matching how each questionnaire was built. The PHQ-9 is designed so its items measure a single severity dimension, and ePID found its information almost entirely overlapping. The Interpersonal Reactivity Index is designed around four distinct facets of empathy, and ePID found much of its information available only from combinations of items. The method recovered each design without being told anything about it, which is what makes the profiles credible. Such profiles could, with further validation, help indicate which symptoms to assess or address together rather than in isolation.


\newpage
\section{Introduction}
Network models represent multivariate systems as nodes and their statistical associations as edges, offering a framework in which the structure of interest emerges from the observed relationships themselves rather than from an assumed latent cause. This approach has been widely adopted in psychopathology, where symptom networks represent mental-health symptoms as nodes and conditional dependencies as edges~\citep{borsboom2013network,borsboom2017network}. Rather than treating symptoms as passive indicators of a single disease entity, the network perspective treats them as mutually reinforcing elements whose interaction patterns constitute the disorder itself~\citep{guloksuz2017application,malgaroli2021networks}. In the most common formulation, edges quantify pairwise conditional associations between symptoms after adjusting for the remaining symptoms, yielding a sparse network often interpreted as a ``direct'' dependency structure.

For example, after a stressful event, an individual may experience insomnia that induces fatigue, which might affect concentration, in turn contributing to depressed mood and feelings of worthlessness~\citep{cramer2016major}. This perspective has shifted the conceptualisation of mental disorders from a centralised latent-disease model to a decentralised network model. However, commonly employed estimators of symptom networks share a fundamental limitation that no choice of pairwise measure can overcome.

Each edge summarises the association between two symptoms as a single number. Even when that association is estimated without assuming a particular functional form---a useful safeguard---a scalar edge still cannot express how several symptoms act \emph{together} to inform a third. That is the gap we address.

Consider how symptoms behave in context. Two symptoms may act together, so that their combined presence is informative even though neither tells us much alone; or they may act as overlapping sources, each carrying largely the same information about a third symptom. For instance, the sleep and fatigue items of the PHQ-9 remain correlated after conditioning on overall depression severity, so that observing one adds little once the other is known~\citep{horton2016rasch,fried2017moving}; by contrast, other symptom pairs may carry information about suicidal ideation jointly that neither carries alone, a possibility that pairwise screening has rarely tested~\citep{franklin2017risk}. A single number per pair cannot tell these cases apart. They can, however, be operationalised as distinct kinds of information: overlapping information that two symptoms share about a target (\emph{redundancy}) and information that becomes available only when they are considered jointly (\emph{synergy}), alongside the information each contributes uniquely.

This distinction has practical implications under an interventionist reading---one that, we stress, presupposes a causal structure and is vulnerable to hidden confounding, so the readings below are strong causal assumptions rather than established conclusions. If two source symptoms are highly redundant, changing the state of only one would change little for the target, as the other continues to convey the same information. Conversely, if their joint information is primarily synergistic, then whether disrupting either source suffices to break the joint channel depends on the underlying causal system: that the joint configuration carries the information does not by itself license the intervention. These interpretations are hypothesis-generating and would require longitudinal or experimental validation; we revisit them in light of the empirical findings in the Discussion.

Information theory supplies the tools to make this operationalisation precise. A source variable provides \emph{information} about a target when observing the source reduces uncertainty about the target's state~\citep{shannon1948mathematical}. Conditional mutual information (CMI) uses this idea to generalise partial correlation (with which it coincides as a conditional-independence criterion under joint Gaussianity) into a model-free (with respect to functional form) measure of conditional dependence. Partial information decomposition (PID) goes further, decomposing the information that two or more \emph{sources} provide about a designated \emph{target} into the non-overlapping components, or \emph{atoms}~\citep{williams2010nonnegative}, sketched above---redundant, unique, and synergistic---so that a conditional association can be read as overlapping or interaction-only rather than as a single undifferentiated number.

The obstacle to applying PID is scale. The number of atoms grows super-exponentially with the number of sources, from 4 atoms for 2 sources to over 7 million for 6, following the Dedekind numbers $M(N)$~\citep{kleitman1969dedekind}. Estimating so many atoms requires high-dimensional joint distributions that are severely undersampled at typical clinical sample sizes, which in practice restricts most analyses to the triplet level (two sources and one target).

Although triplet-level decompositions are tractable, they miss interactions that emerge only when three or more sources act together. In symptom networks the relevant comparator for a given symptom is its full context---all remaining symptoms---so triplet decompositions cannot answer the system-level attribution question of primary interest: whether a symptom contributes information about a target \emph{beyond what is already present in the rest of the symptom profile}.

To address this, we introduce an embedding-based PID pipeline, which we refer to as \emph{ePID}. For a given pair of symptoms, ePID first compresses the rest of the network into a compact summary variable---a small number of categories learned to preserve as much information as possible about the \emph{target} symptom specifically---and then asks how the focal symptom (i.e., the source in the pair) relates to the target compared with that summary. Because such a decomposition is defined entirely by information about the target, a summary optimised to preserve the remainder's target-relevant information retains what it depends on, while what the summary discards is, by construction, largely irrelevant to the target; we make this approximation, and the information it loses, precise in the Methods. This target-supervised compression is specific to ePID; partial correlation and CMI, by contrast, are not directed at any particular target. The procedure yields four components for each ordered symptom pair: information that only the focal symptom provides (source-unique), information that only the rest of the network provides (remainder-unique), overlapping information (redundancy), and information that emerges only when the focal symptom and the summary are considered together (synergy). The resulting directed edges reflect the source--target orientation that the decomposition requires.

A complementary strand of psychometric network research tackles higher-order structure with Moderated Network Models (MNM), which add multiplicative interaction terms so that the association between two symptoms can vary with a third~\citep{haslbeck2021moderated}. The contrast with ePID is essentially parametric versus non-parametric: MNM provides a parametric test of whether a chosen moderator significantly modifies a given edge, whereas ePID asks, without specifying interactions in advance, how the information underlying each edge is distributed across redundant, unique, and synergistic channels relative to the rest of the symptom profile. We develop the fuller methodological contrast in the Discussion.

\begin{figure*}[!t]
  \centering
  \includegraphics[width=\textwidth]{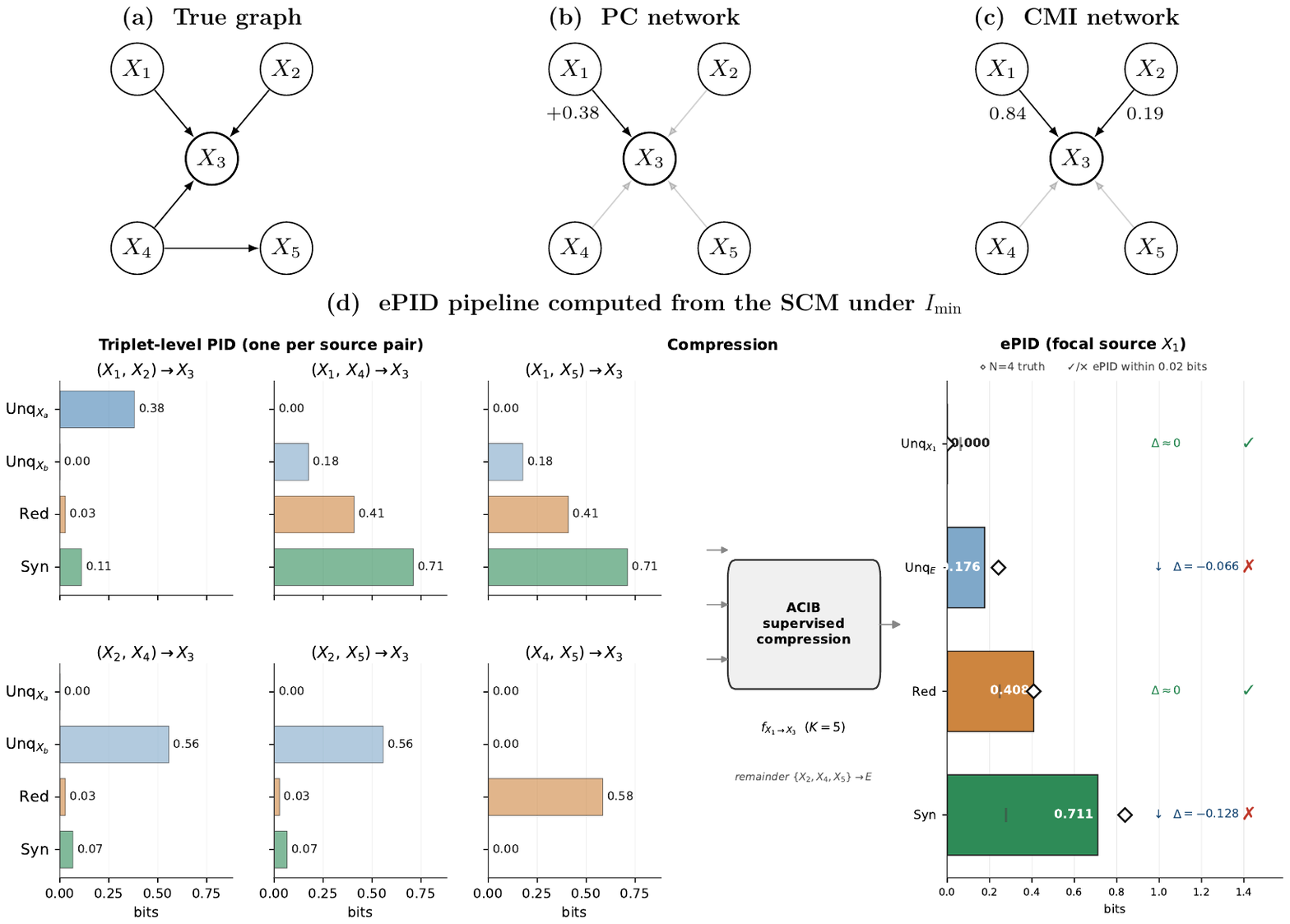}
  \caption{\apacap{Methods comparison on the toy structural causal model}{The toy system is a five-variable mixture structural causal model ($X_1, X_2, W \sim \mathrm{Uniform}\{0,\ldots,4\}$ independent; $X_4 = X_5 = W$; $X_3 = h(X_1, X_2)$ with probability $0.5$ else $X_3 = W$, where $h$ is the non-monotonic stylised mapping defined in Appendix~\ref{app:stylised_example}). Top row: \textbf{(a)} True graph: $X_1, X_2, X_4 \to X_3$ with $X_4 = X_5$ marked. \textbf{(b)} PC network: only the $X_1 \to X_3$ edge survives ($\rho_{ij \mid \text{rest}} = +0.38$); $X_4$, $X_5$ edges are undefined under naive PC because the deterministic equality $X_4 = X_5$ makes the conditioning covariance singular. \textbf{(c)} CMI network: $X_1 \to X_3$ and the non-monotonic $X_2 \to X_3$ edge (0.84 and 0.19 bits respectively); $X_4$, $X_5$ contributions vanish because conditioning on the remaining system already includes the duplicate $X_5 = X_4$, giving each zero conditional entropy given the rest and hence $I(X_4; X_3 \mid \text{rest}) = 0$; the edges disappear from vanishing conditional mutual information, not from the conditioning rule itself, and the same attenuation arises under near-redundancy when the sources are highly but not perfectly correlated. Bottom row: \textbf{(d)} the ePID pipeline. Six triplet-level two-source PIDs (one per source pair targeting $X_3$, shown at lower opacity to signal pair-dependent attribution) collapse via ACIB-supervised compression of the remainder $\{X_2, X_4, X_5\}$ into a four-atom ePID for focal source $X_1$. All atom values are computed exactly under $I_{\min}$ (Williams \& Beer 2010); the corresponding BROJA decomposition is shown in Appendix Figure~\ref{fig:method_comparison_broja}. PC $=$ partial correlation; CMI $=$ conditional mutual information; PID $=$ partial information decomposition; ePID $=$ embedding-based PID; ACIB $=$ Agglomerative Conditional Information Bottleneck.}}
  \label{fig:method_comparison}
\end{figure*}

This study estimates and compares three approaches to symptom network estimation that progressively relax modelling assumptions and move from pairwise to higher-order characterisations (Figure~\ref{fig:method_comparison}). Partial correlations (PC), in their Pearson (PPC) and Spearman (SPC) variants, provide a baseline for conditional linear and monotonic associations; the two variants yield highly similar networks on the ordinal symptom items analysed here (Appendix~\ref{app:cmi_pid_comparison}, Figure~\ref{fig:PPC_vs_PSP}), and we therefore present them jointly as PC. CMI generalises to arbitrary conditional dependence while remaining at the pairwise level. To characterise how that dependence is composed, we apply PID via the ePID pipeline, yielding for each source--target pair a decomposition into the four atoms above. The degree to which these methods agree or diverge is itself informative about the statistical structure of the symptom system under study.

We evaluate ePID in two stages. First, we benchmark multiple embedding strategies in simulation settings calibrated to PHQ-9 response distributions, then extend the evaluation across 83 real-world datasets spanning five PID measures to assess generalisability, quantifying how much information the compression loses (the approximation error). Second, we apply the approach to item-level depressive-symptom networks in two datasets (UK Biobank~\citep{sudlow2015uk,davis2019mhq_reanalysis} and the Xinxiang student sample~\citep{Su2024_TemporalDynamics,Su2024_TemporalDynamics_data}), using harmonised PHQ-9 items (excluding suicidal ideation) to assess cross-cohort stability, and to the Interpersonal Reactivity Index~\citep{Davis1983_IRI}, a multi-facet empathy instrument that serves as a contrasting psychometric profile. In each application we first treat the agreement between partial correlations and CMI as an assumption check on whether conditional structure is largely monotonic at the item level, and then use ePID to add an interaction-aware interpretation layer, separating redundant from synergistic channels relative to the full symptom context.

By combining model-free measures of dependence with a scalable decomposition method, we characterise what kinds of statistical structure underlie observed symptom networks and assess whether, and to what extent, interaction-aware models are warranted for a given instrument and population. To our knowledge this is the first application and systematic validation of supervised-embedding PID to symptom networks, and although we develop it here for psychometric instruments, the pipeline is generic and should apply to other multivariate systems in which network-context higher-order structure is of interest---a generalisation we do not yet validate empirically. We emphasise throughout that these patterns describe statistical associations in cross-sectional data rather than causal mechanisms, and should be read as constraints and hypotheses for future longitudinal and experimental work; their empirical magnitudes, and their behaviour across cohorts and instruments, are reported in the Results.

\section{Methodology}
The methodology proceeds as follows. We first introduce notation and then describe two correlation-based methods: Pearson and Spearman partial correlations (PPC and SPC). Next, we explain conditional mutual information (CMI), which relaxes the linearity and monotonicity assumptions of PPC and SPC respectively. These three methods capture only pairwise associations.

Partial information decomposition (PID) goes further by decomposing the information that multiple source variables provide about a target into redundant, synergistic, and unique components, illustrated with a stylised example (Appendix~\ref{app:stylised_example}). To make this decomposition tractable beyond a handful of sources, we introduce an embedding-based PID method (ePID) that compresses the remaining variables into a low-dimensional proxy while retaining information about the target (Section~\ref{subsec:embedding_symptoms}). We then describe the simulation and multi-dataset benchmarks used to validate this approximation, define the normalisation and significance-testing framework, and outline the empirical-analysis procedures applied to the two PHQ-9 cohorts and the IRI empathy instrument: permutation testing with false discovery rate control for edge significance, entropy-normalised edge ranking for network visualisation, and cross-cohort stability assessment between the two PHQ-9 cohorts.

\subsection{Notation}
Let $\mathbf{X} = \{X_1, X_2, \ldots, X_p\}$ denote $p$ discrete random variables. Each $X_i$ can represent the presence or severity of symptom $i$ in an individual, for example, or other covariates such as education or neighbourhood characteristics. The cardinality of a discrete variable $X_i$ is denoted $|X_i|$ and equals the number of categories. For any ordered pair $(X_i, X_j)$, we write $Z_{ij} = \mathbf{X} \setminus \{X_i,X_j\}$ for the remaining variables used as a conditioning set. For PID, we consider the triplets $\{X_i,X_j,X_k\}$ to be directed, where we refer to $\{X_i,X_j\}$ (inputs) as sources and $\{X_k\}$ (output) as the target.

For embedding-based decomposition, we consider an ordered source--target pair $(X_i, X_k)$ and denote the remaining variables by
\[
Z_{i\to k} := \{X_\ell : \ell \in \{1,\ldots,p\} \setminus \{i,k\}\}.
\]
In words, $Z_{i\to k}$ collects every symptom other than the focal source $X_i$ and the target $X_k$: the rest of the network, whose joint state we will compress into a low-dimensional embedding.
We construct a supervised discrete embedding of this remainder,
\[
E_{i\to k} := f_{i\to k}(Z_{i\to k}),
\]
where $f_{i\to k}$ is fit to predict the target $X_k$ from $Z_{i\to k}$ and $E_{i\to k}$ has finite cardinality $|E_{i\to k}|$. We then compute a two-source PID of $(X_i, E_{i\to k}; X_k)$, interpreting $E_{i\to k}$ as a compressed representation of the network context (the ``remainder embedding'').

\subsection{Data}
\label{subsec:data}
We analyse Patient Health Questionnaire (PHQ-9) item responses from two datasets: the UK Biobank online mental health questionnaire~\citep{davis2019mhq_reanalysis} and a large online sample of Chinese university students (Xinxiang Medical University; Su et al.\ 2024)~\citep{Su2024_TemporalDynamics,Su2024_TemporalDynamics_data}. The PHQ-9 is a widely used and validated instrument for screening and severity assessment of depressive symptoms in clinical and population settings~\citep{kroenke2001phq9,spitzer1999phq}. Each item asks how often, in the last two weeks, respondents were bothered by a symptom, with ordinal response categories coded as 0 (``Not at all''), 1 (``Several days''), 2 (``More than half the days''), and 3 (``Nearly every day'').

For the primary analyses, we focus on eight PHQ-9 items (anhedonia, depressed mood, sleep problems, fatigue/low energy, appetite change, feelings of worthlessness or excessive guilt, concentration problems, and psychomotor changes). We exclude the suicidality item (thoughts of self-harm) because it is extremely sparse in both datasets, leading to unstable estimation in conditional and higher-order analyses. To reduce sparsity in the remaining items and harmonise measurement across datasets, we merge response categories 2 and 3 and analyse a three-level ordinal scale \(\{0,1,2\}\) for each retained item. Because categories 2 and 3 are the rarest, this adjacent merge preserves ordinality but yields an asymmetric scale on which the lowest-severity level is well sampled and the highest is compressed into a single cell, which reduces the attainable entropy, and hence the absolute mutual information in bits, for high-severity symptom combinations. Because we report atoms as shares of the target entropy (Section~\ref{sec:normalisation}), however, this reduction acts largely as a common rescaling that cancels in the shares: recomputing a direct two-source PID at two, three, and four levels yields a near-constant synergy share of $0.193$, $0.188$, and $0.190$ respectively, with the redundancy share stable at $0.534$ to $0.542$ and hence dominant throughout. We therefore treat the merge as likely conservative for synergy rather than as a fixed downward bias, and return to this point in Section~\ref{subsec:limitations}. After preprocessing, the UK Biobank dataset comprised $N=154{,}291$ participants and the Xinxiang student sample comprised $N=24{,}292$ respondents.

To assess whether the ePID pipeline generalises beyond depressive-symptom networks to instruments with different higher-order structure, we additionally analyse responses to the Interpersonal Reactivity Index (IRI), a widely used multidimensional measure of dispositional empathy~\citep{Davis1983_IRI}. The IRI comprises 28 items distributed across four 7-item subscales: Perspective-Taking (PT) and Fantasy (FS), indexing cognitive-empathy facets, and Empathic Concern (EC) and Personal Distress (PD), indexing affective-empathy facets~\citep{Davis1983_IRI}. Responses are scored on a five-point Likert scale; for the present analyses we discretise to a three-level ordinal scale to harmonise cardinality with the PHQ-9 protocol. The dataset analysed here comprises $N = 1{,}973$ respondents from the Open-Source Psychometrics Project (\url{https://openpsychometrics.org/_rawdata/}). We apply ePID using the same protocol as for PHQ-9 ($n_{\mathrm{sources}} = 5$ ACIB-embedded remainder context, $I_{\mathrm{mmi}}$ redundancy measure), supplemented by within-subscale and cross-subscale analyses reported in Section~\ref{subsec:iri_results}.

\subsection{Correlation Networks}
We represent the symptom system as an undirected graphical model in which nodes are symptoms and edges represent conditional dependence after adjusting for the remaining symptoms. Under a Gaussian (or Gaussian-copula) graphical model, the absence of an edge corresponds to conditional independence given all other symptoms, while a nonzero partial correlation indicates a conditional association within that model class.

The (population) partial correlation between $X_i$ and $X_j$ given $Z_{ij}$, denoted $\rho_{ij \mid Z_{ij}}$, is defined as the correlation between the residuals obtained by regressing $X_i$ and $X_j$ on $Z_{ij}$, respectively.

To relax marginal normality and linearity assumptions, we compute rank-based partial correlations (SPC). Specifically, each variable $X_k$ is transformed by its empirical marginal distribution to obtain ranks $R(X_k)$, and partial correlations are computed on the transformed variables. This procedure estimates conditional associations under a Gaussian copula graphical model: dependencies are encoded by a latent multivariate normal structure while marginal distributions are left unrestricted. Rank-based partial correlations therefore capture conditional monotonic dependence, irrespective of marginal scale or skewness.

Partial correlations take values in $[-1,1]$ and define an undirected network in which edges represent conditional associations between symptom pairs.

\subsection{Information-Theoretic Networks}
To quantify conditional dependence without restricting the functional form of associations, we construct information-theoretic networks based on mutual information.

For a discrete random variable \(X\), Shannon entropy \(H(X)\) quantifies uncertainty about its outcome on a logarithmic scale and provides a natural uncertainty baseline for categorical data. For a variable with $|X|$ categories, $H(X) \le \log_2 |X|$. Mutual information between two variables $X_i$ and $X_j$,
\[
I(X_i; X_j) = H(X_i) - H(X_i \mid X_j),
\]
measures the reduction in uncertainty about one variable given knowledge of the other. Mutual information is symmetric, non-negative, and makes no assumptions about linearity or monotonicity.

To isolate direct associations in a network setting, we compute conditional mutual information (CMI),
\[
I(X_i; X_j \mid Z_{ij}) = H(X_i \mid Z_{ij}) - H(X_i \mid X_j, Z_{ij}),
\]
which measures the remaining dependence between $X_i$ and $X_j$ after conditioning on all other variables. Unlike correlation-based measures, CMI detects arbitrary conditional dependence. At the population level, \(I(X_i;X_j\mid Z_{ij})=0\) if and only if \(X_i\) and \(X_j\) are conditionally independent given \(Z_{ij}\) for discrete variables, so CMI provides a model-free criterion for conditional independence. In finite samples, plug-in estimates of CMI are positively biased; we address this via permutation testing (Section~\ref{sec:significance_testing}).

Correlation-based measures and CMI estimate the same conditional dependence graph but differ in the class of dependencies they can detect. Partial correlations detect only linear or monotonic conditional associations, whereas CMI provides a necessary and sufficient test of conditional independence for discrete data. Because mutual information is in turn bounded above by the relevant entropies, we report a normalised CMI that can be interpreted as the fraction of remaining uncertainty explained (Section~\ref{sec:normalisation}).

\subsection{Higher-order interactions}
\label{sec:higher_order}
Having established the pairwise measures, we now move from pairwise dependence to interactions among more than two variables. We proceed in three steps: (i) define the two-source PID and its redundancy function, (ii) introduce embedding-based PID (ePID), which compresses the multivariate remainder of the symptom system into a supervised discrete embedding $E_{i\to k}$ for each focal source--target pair, and (iii) map the resulting atoms onto directed network visualisations. A stylised three-variable example (Appendix~\ref{app:stylised_example}) builds intuition before the formal definitions.

Multivariate extensions of mutual information quantify whether dependence among groups of variables is primarily overlapping or interaction-specific. A core diagnostic is that conditioning can reveal dependencies invisible to marginal analysis: in an exclusive-or (XOR) relationship between two independent, uniformly distributed binary inputs, where the output indicates whether the inputs differ, each input alone carries no information about the output, whereas the joint state is fully informative. We apply partial information decomposition (PID) to obtain non-negative unique, redundant, and synergistic components, using an embedding of the remaining symptoms to make the decomposition tractable at PHQ-9 scale.

Several complementary frameworks provide scalable summaries of higher-order dependence without estimating a full multivariate PID. 
System-level measures such as the O-information (and related local or dynamical variants) quantify whether multivariate dependence is dominated by redundancy or synergy, but do not attribute information to specific source--target relations~\citep{rosas2019oinformation,scagliarini2023gradients}.
Decision-theoretic and entropy-centric decompositions offer principled alternatives to the standard PID lattice: the redundancy bottleneck recasts redundancy as a constrained optimisation over compressed representations of the sources~\citep{kolchinsky2024redundancy_bottleneck}, while partial entropy decomposition operates on joint entropy rather than mutual information~\citep{ince2017ped}. Both can scale via convex optimisation or algebraic structure. Finally, synergy-backbone approaches aim to identify irreducible synergistic subsets in large systems while avoiding the full lattice~\citep{varley2024synergybackbone}.
These methods highlight a common trade-off between scalability and attributional resolution; here we retain PID's directional, source--target interpretability while scaling to the full symptom context via ePID's remainder embedding (Section~\ref{subsec:embedding_symptoms}).

As a concrete anchor, consider a stylised three-variable system (stress, sleep problems, and concentration) in which concentration tracks sleep under normal conditions but, when sleep is very poor, depends on stress in a non-monotonic way. A partial correlation recovers only the sleep-to-concentration link; conditional mutual information additionally flags the stress-to-concentration dependence; and PID makes the structure explicit, attributing about $71\%$ of the information about concentration uniquely to sleep, $7\%$ to redundancy, and $21\%$ to synergy between stress and sleep (the interaction-only signal that partial correlation misses entirely and conditional mutual information detects only indirectly). The full worked construction is given in Appendix~\ref{app:stylised_example}. This pattern is not an artefact of the construction: in the real IRI empathy data, the analogous configuration (sources Fantasy and Personal Distress, target Perspective-Taking) is synergy-dominant in $69.4\%$ of informative cross-subscale triplets, with a mean synergy fraction of about $21\%$, albeit only about $0.006$ bits in total. We now formalise these atoms.

\subsubsection{Two-source partial information decomposition}
\label{subsec:pid_two_source}

Partial information decomposition (PID) refines mutual information by decomposing the information that two sources $(X_i,X_j)$ provide about a target $X_k$ into four non-negative components (information ``atoms''): redundant, unique-to-$X_i$, unique-to-$X_j$, and synergistic information. For discrete variables, we write
\begin{equation}
I(X_i,X_j; X_k) = \mathrm{Red} + \mathrm{Unq}_{X_i} + \mathrm{Unq}_{X_j} + \mathrm{Syn},
\end{equation}
subject to the consistency relations
\begin{align}
I(X_i; X_k) &= \mathrm{Red} + \mathrm{Unq}_{X_i}, \\
I(X_j; X_k) &= \mathrm{Red} + \mathrm{Unq}_{X_j}.
\end{align}
That is, the four atoms partition the joint mutual information $I(X_i,X_j;X_k)$ into non-overlapping non-negative pieces, and individually they must sum to the marginal informations $I(X_i;X_k)$ and $I(X_j;X_k)$ via the consistency constraints.
Here $\mathrm{Red}$ is information that both sources provide about the target, $\mathrm{Unq}_{X_i}$ and $\mathrm{Unq}_{X_j}$ are information that one source provides and the other does not, and $\mathrm{Syn}$ is information that only becomes available when the two sources are considered jointly (interaction-only effects).

Different PID measures pin down these four atoms in different ways. Most, including the minimum mutual information (MMI) redundancy measure we use in all embedding-based analyses, define the redundant information first and obtain the others from it; the BROJA measure (below) instead defines the unique information first, as the smallest unique information consistent with the source and target marginals. MMI satisfies self-redundancy, symmetry, and monotonicity and has been widely applied in discrete systems.
MMI defines redundancy as the minimum information that any single source provides about the target, $\mathrm{Red}_{\mathrm{MMI}} = \min\{I(X_i; X_k),\, I(X_j; X_k)\}$, and derives the remaining atoms via the consistency relations above.
This definition is intuitive (redundancy cannot exceed what any one source individually reveals) and is computationally efficient for low-cardinality systems.
A known limitation of MMI is that it can overestimate redundancy when two sources carry different information about the target of equal magnitude, because the minimum over marginal mutual informations equates equal MI values with shared content regardless of whether the sources encode different aspects of the target~\citep{bertschinger2014broja}. Consequently, MMI may underestimate synergy.
Alternative redundancy definitions address this limitation from different angles. For example, the BROJA measure~\citep{bertschinger2014broja} defines redundancy as the smallest amount of unique information consistent with the observed source--target marginals, expressed as a decision-theoretic optimisation; but each introduces additional computational cost or conceptual trade-offs.
To assess sensitivity to the redundancy definition, we additionally compute decompositions using four alternative measures as robustness checks: the $I_{\min}$ (Williams--Beer) measure~\citep{williams2010nonnegative}, the original PID redundancy that takes the pointwise minimum specificity across sources; a rescaled redundancy measure $I_{\mathrm{rr}}$~\citep{goodwell2017temporal} that normalises by source entropy; the pointwise partial information measure $I_{\pm}$~\citep{finn2018pointwise}, which decomposes information at the level of individual joint outcomes rather than averaged distributions; and the G\'acs--K\"orner common-information measure $I_{\wedge}$~\citep{gacs1973common}, which retains only the information that two sources jointly encode in a perfectly aligned (deterministic) way. Qualitative patterns in our empirical analyses (in particular the ordering of edges by synergy fraction) were consistent across all five measures.

Computing $I_{\pm}$ and $I_{\wedge}$ for five or more sources via conventional lattice enumeration is computationally prohibitive. The PID redundancy lattice, the hierarchy that organises all non-trivial ways source subsets can overlap in their information about a target, contains $7{,}579$ antichains (the lattice nodes that index distinct PID atoms) at $N = 5$ sources. We bypass this bottleneck using the fast M\"obius transform (FMT)~\citep{jansma2025fastmobius}, which precomputes the lattice structure and obtains each target's PID atoms via a single matrix--vector multiplication, reducing per-target computation from hours to minutes.

\subsubsection{Embedding the remaining sources for symptom networks}
\label{subsec:embedding_symptoms}

In principle, PID can be extended to more than two sources, but the number of information atoms grows super-exponentially with the number of sources (from 4 atoms at two sources to over 7,500 at five; Section~\ref{subsec:pid_two_source}), quickly becoming computationally and statistically intractable. Two factors compound this. First, the decomposition is not computed once but re-estimated for every directed source--target pair, in each dataset, across the five redundancy measures, and across the calibrated synthetic ensemble used for validation, so the nominal per-target cost is incurred many times over. Second, stable estimation of a high-order joint distribution requires adequate counts in each joint cell: the eight non-target symptoms together with the target (nine three-level variables) span $3^{9} \approx 2 \times 10^{4}$ joint states, leaving on the order of one observation per state in the Xinxiang student sample ($N \approx 24{,}000$) and about eight in UK Biobank ($N \approx 154{,}000$), far too sparse for reliable higher-order estimates. To approximate the high-order structure while retaining interpretability, we adopt an embedding-based strategy. Intuitively, instead of estimating the full joint distribution of all remaining symptoms, we replace them with a single compact summary variable---a learned grouping of the possible remainder configurations that keeps what is informative about the target and discards the rest---so the decomposition need only handle the focal symptom together with this summary. Throughout this section and in all empirical analyses, the two-source decomposition uses the MMI redundancy measure $I_{\mathrm{mmi}}$ defined in Section~\ref{subsec:pid_two_source}. The four alternative measures introduced there are reported as robustness checks and as a benchmark axis in the multi-dataset evaluation, whose results are in Appendix~\ref{subsec:results_multidataset}.

For each ordered symptom pair $(X_i, X_k)$ with $i \neq k$, we treat $X_k$ as the target and $X_i$ as a focal source.
The remaining variables $\mathbf{X} \setminus \{X_i,X_k\}$ are compressed into a single discrete embedding
\[
E_{i\to k} \;=\; f_{i\to k}\!\left(\mathbf{X} \setminus \{X_i,X_k\}\right).
\]
This embedding provides a compact proxy for the joint configuration of the remaining symptom network when paired with the focal source in a two-source PID.

We consider multiple constructions of $f_{i\to k}$, including (i) \emph{target-directed} embeddings that learn $f_{i\to k}$ using $X_k$ and (ii) \emph{unsupervised} embeddings that compress $\mathbf{X} \setminus \{X_i,X_k\}$ without access to $X_k$.
All candidate embeddings are evaluated in a simulation benchmark against tractable multivariate ground truth (Section~\ref{subsec:embedding_benchmark}).

Our primary embedding, which we term the Agglomerative Conditional Information Bottleneck (ACIB), groups the joint states of the remainder symptoms into a small number of clusters and then iteratively merges the most similar clusters, choosing each merge so that the conditional information that the remainder carries about the target (given the focal source) is preserved as much as possible. The result is a low-cardinality discrete embedding $E_{i\to k}$ that summarises the rest of the symptom network without discarding the part of its dependence with the target that the focal source does not already explain.

Formally, ACIB adapts conditional information-bottleneck clustering~\citep{gondek2003conditional} with agglomerative merging~\citep{slonim1999agglomerative} to the PID setting. It operates in two phases: first, joint states of the remainder are assigned to clusters by minimising a conditional Kullback--Leibler-divergence distortion; second, the closest cluster pair is greedily merged while constraining the relative loss in conditional mutual information, $1 - I(X_i; X_k \mid E_{i\to k}) / I(X_i; X_k \mid Z_{i\to k})$, to remain below a specified tolerance. Because this constraint operates on conditional rather than unconditional mutual information, the compression preserves the source--target relationship that PID subsequently decomposes.

ACIB is fitted separately for each $(X_i, X_k)$ pair, using $\mathbf{X} \setminus \{X_i, X_k\}$ as inputs. It has three hyperparameters: the conditional-information loss tolerance, the cardinality cap $K_{\max}$, and a Dirichlet smoothing constant, which we set to $5\%$ relative information loss, $K_{\max}=12$, and $0.5$ throughout. A dedicated sensitivity analysis (Appendix~\ref{app:acib_sensitivity}) shows that fidelity is essentially insensitive to the loss tolerance and the smoothing constant and that $K_{\max}=12$ is statistically indistinguishable from neighbouring values; ACIB was selected over alternative embeddings for its superior fidelity at $N \geq 4$ sources (Section~\ref{subsec:embedding_benchmark}).

Given the embedded system $(X_i, E_{i\to k}; X_k)$, we compute a two-source PID using the MMI measure described above. The resulting atoms,
\[
\mathrm{Unq}_{i\to k}, \quad \mathrm{Unq}_{E_{i\to k}}, \quad \mathrm{Red}_{i\to k}, \quad \mathrm{Syn}_{i\to k},
\]
are interpreted as follows for each target $X_k$:
\begin{itemize}
  \item $\mathrm{Unq}_{i\to k}$: information about $X_k$ that is carried uniquely by symptom $X_i$ beyond what is available from the embedded remainder $E_{i\to k}$;
  \item $\mathrm{Unq}_{E_{i\to k}}$: information about $X_k$ that is carried uniquely by the rest of the network (via $E_{i\to k}$) and not by $X_i$;
  \item $\mathrm{Red}_{i\to k}$: information about $X_k$ that is shared between $X_i$ and the embedded remainder;
  \item $\mathrm{Syn}_{i\to k}$: information about $X_k$ that becomes available only when $X_i$ and the embedded remainder are considered together.
\end{itemize}
These atoms provide an approximate decomposition of the information that all other symptoms collectively carry about $X_k$ into components attributable uniquely to $X_i$, uniquely to the remainder, shared, and synergistic channels. Because the embedding is many-to-one, the decomposition is approximate rather than exact for the original multivariate system; its accuracy is characterised in the simulation study below.

In ePID, the remainder-unique atom $\mathrm{Unq}_{E_{i\to k}}$ quantifies predictability of the target from the remainder of the symptom system alone. In contrast, the redundancy and synergy atoms necessarily involve the focal source $X_i$ and therefore quantify \emph{source involvement} in the source$\rightarrow$target dependence. That is, $\mathrm{Unq}_{E_{i\to k}}$ quantifies \emph{remainder-only predictability} of $X_k$, $\mathrm{Red}_{i\to k}$ quantifies \emph{overlap} between the focal symptom and the remainder, and $\mathrm{Syn}_{i\to k}$ quantifies \emph{interaction-only information} that is available only from their joint configuration.

Before computing the two-source PID, each of the three variables (focal source, embedding, and target) is capped to at most five categories: the four most frequent values are retained and all remaining, rarer values are pooled into a single ``other'' category. Unlike the adjacency-preserving PHQ-9 merge of Section~\ref{subsec:data}, this is a purely frequency-based cap that does not preserve ordinal adjacency; for the three-level PHQ-9 items it is therefore a no-op on the source and target and binds only on the embedding (a nominal cluster label, for which order is irrelevant), whereas for higher-cardinality instruments such as the IRI it can pool non-adjacent ordinal levels. This cardinality cap ensures that the joint distribution $(X_i, E_{i\to k}, X_k)$ remains well-sampled at typical clinical sample sizes.

\subsubsection{Simulation benchmark for embedding validation}
\label{subsec:embedding_benchmark}

Because the embedding-based PID is approximate and the ground-truth multivariate PID cannot be estimated from observational data at scale, we validate against calibrated synthetic systems where exact computation is tractable. This benchmark serves two purposes: (i) quantifying how well ePID recovers network-context unique, redundant, and synergistic structure, and (ii) selecting an embedding method that is accurate under empirically realistic marginal distributions. We generate an ensemble of Bayesian networks calibrated to PHQ-9 data and compare the embedding-derived two-source PIDs to the ground-truth multivariate PID.

\paragraph{Bayesian network ensemble.}
Using the UK Biobank PHQ-9 item distributions as a reference, we learn a directed acyclic graph over nine discrete variables for benchmarking purposes. We denote each synthetic Bayesian network by $G$ and write $(G, X_t, N)$ for the configuration in which target $X_t$ and number of sources $N$ are evaluated on network $G$. The empirical analyses exclude suicidal ideation due to sparsity (only $4.3\%$ and $7.6\%$ of respondents endorsed any thoughts of self-harm at severity $\geq 1$ in UK Biobank and the Xinxiang student sample respectively), but including it in the synthetic calibration provides a stress test in the presence of low-prevalence states. Conditional probability tables are estimated from the empirical data and then perturbed with small Dirichlet noise to create 50 distinct Bayesian networks. For each network we draw  $600\,000$ samples by forward sampling. This yields an ensemble of synthetic datasets that preserve the marginal distributions and broad dependency patterns of the PHQ-9 items while allowing exact computation of information-theoretic quantities. Calibration diagnostics confirm that the synthetic ensembles closely match the empirical mutual information structure (mean MI error $6.10\%$; Kolmogorov--Smirnov statistic $0.166$; see Figure~\ref{fig:mi_distribution_comparison}).

\paragraph{Ground-truth multivariate PID.}
For each Bayesian network $G$, each target $X_t$ (one of the nine variables), and each number of sources $N \in \{3,4,5\}$, we select $N$ distinct source variables from the remaining items and compute the full $N$-source PID of $(X_i,\ldots,X_N; X_t)$ using the MMI PID measure. This yields a redundancy lattice whose atoms specify unique, redundant, and synergistic contributions for all combinations of sources.

\paragraph{Embedding-based PID on the same systems.}
For the same $(G, X_t, N)$ configurations, we approximate the multivariate PID using two-source decompositions as follows. For each choice of focal source $X_i$ among the $N$ sources, we embed the remaining $N-1$ sources $\{X_j : j \neq i\}$ into a one-dimensional discrete variable $E$ using each of 13 embedding methods spanning three families:
\begin{itemize}
  \item \emph{Manifold and dimensionality reduction.}
    ACIB, the agglomerative conditional information bottleneck described above~\citep{gondek2003conditional,slonim1999agglomerative};
    multiple correspondence analysis followed by $k$-means clustering of factor scores~\citep{greenacre2017mca,halford2023prince};
    non-negative matrix factorisation followed by $k$-means~\citep{lee1999nmf};
    sliced inverse regression with quantile binning, which estimates a target-supervised low-dimensional projection~\citep{li1991sir,koepke2018sliced};
    partial least squares with quantile binning, which extracts components maximising covariance with the target~\citep{wold1984pls};
    and truncated singular value decomposition followed by $k$-means~\citep{golub2013matrix}.
  \item \emph{Feature selection.}
    Conditional mutual information maximisation (CMIM), which selects features that maximise CMI with the target conditional on previously chosen features~\citep{fleuret2004cmim};
    joint mutual information (JMI), which selects features by their average pairwise joint information with the target~\citep{yang1999jmi};
    a greedy CMI-based selector that adds variables one at a time to maximise the running conditional mutual information~\citep{brown2012featureselection};
    and ReliefF, which scores features by their ability to distinguish nearest hits from nearest misses across classes~\citep{kononenko1994relieff,urbanowicz2018skrebate}.
  \item \emph{Clustering.}
    Agglomerative clustering with Hamming distance over the joint state vector~\citep{murtagh2012hierarchical,finch2005distance};
    $k$-modes, the categorical analogue of $k$-means using mode-based centroid updates~\citep{huang1998kmodes};
    and spectral clustering with a Hamming-distance-based affinity matrix~\citep{ng2001spectral}.
\end{itemize}

We then compute a two-source PID for $(X_i, E; X_t)$ using the MMI measure.

\paragraph{Amalgamation of multivariate PID.}
To compare the two-source PID on $(X_i,E;X_t)$ with the $N$-source PID on $(X_1,\ldots,X_N;X_t)$, we collapse the many atoms of the $N$-source decomposition into four two-source atoms via a structural amalgamation that groups the $N-1$ non-focal sources into a single composite source, so that the projected atoms answer the same question as ePID: how much information about $X_t$ is attributable to the focal source $X_i$ beyond what the remainder provides. The mapping is conservative with respect to synergy, classifying an atom as synergistic only when it necessarily requires contributions spanning both the focal source and the remainder group, so the amalgamated synergy is a lower bound on the true interaction-only information between $X_i$ and the remainder. The grouping preserves the two-source consistency relations. The full antichain classification rule and its derivation are given in Appendix~\ref{app:embed_computational}. Figure~\ref{fig:method_3panel} visualises this amalgamation step on the toy structural causal model used in the methods-comparison figure of the main text.

\paragraph{Error metrics.}

For each PID atom $c \in \{\mathrm{Unq}_{X_i}, \mathrm{Unq}_{E}, \mathrm{Red}, \mathrm{Syn}\}$ and each embedding method, we compute the absolute error
\begin{equation}
\mathrm{AE}_c = \left| \mathrm{PID}^{(N\text{-source})}_c - \mathrm{PID}^{(\text{embed})}_c \right|,
\end{equation}
where the first term is the amalgamated ground-truth value and the second term is the corresponding atom from the two-source PID on $(X_i,E;X_t)$. In words, the absolute error captures the per-atom gap between embedding-based and ground-truth values; smaller values indicate that the embedding more faithfully reproduces the multivariate decomposition for that atom. We summarise errors across all $(G, X_t, N)$ configurations and focal sources. Figure~\ref{fig:embedding_overall_performance} displays the mean absolute error by embedding method and PID atom; Appendix Figure~\ref{fig:embedding_scatter} compares approximated versus ground-truth synergy for one representative method from each family tier (ACIB, JMI, $k$-modes); and Appendix Figure~\ref{fig:embedding_synergy_scaling} shows how synergy error scales with the number of sources.

For the multi-dataset benchmark (Appendix~\ref{subsec:results_multidataset}), we additionally report the total absolute error $\mathrm{TAE} = \sum_{c} \mathrm{AE}_c$ summed across all four PID atoms, and the relative error $\mathrm{RE} = \mathrm{TAE} \,/\, I(X_i,E; X_t)$, where the denominator is the ground-truth total mutual information that the focal source and its embedded remainder jointly carry about the target (equivalently, the sum of the four ground-truth atoms). RE expresses the approximation error as a fraction of the total information being decomposed, enabling comparison across datasets with different baseline information levels.

The simulation study serves two purposes. First, it establishes that embedding-based PID can approximate the multivariate decomposition with low error when suitable embeddings are used. Second, it provides empirical guidance for choosing embeddings in the empirical PHQ-9 analyses, where the ground-truth PID is unknown.

\paragraph{Applicability beyond PHQ-9.}
The ePID pipeline is designed to accommodate datasets with varying numbers of variables, response cardinalities, and sample sizes. For variables with many categories, the general pipeline supports cardinality capping that merges infrequent categories to reduce the dimensionality of joint distributions before embedding and PID computation. This step is exercised in the multi-dataset benchmark (Section~\ref{subsec:multi_dataset_benchmark}), where datasets vary in cardinality. For the PHQ-9 items analysed here, the three-level ordinal scale after category merging is already low-cardinality, making this additional step unnecessary. Computational details for the embedding benchmark, including the subsampling strategy for distance-based methods, are provided in Appendix~\ref{app:embed_computational}. Figure~\ref{fig:pipeline} provides an overview of the full ePID evaluation pipeline, including both the embedding-based path and the ground-truth benchmark path.

\begin{figure}[t]
%

\centering
\resizebox{\textwidth}{!}{%
\begin{tikzpicture}[
  pipenode/.style={
    rectangle, rounded corners=2pt, draw=#1, thick,
    fill=white, text=black, font=\small,
    minimum height=8mm, text width=22mm, align=center,
    inner sep=2pt
  },
  pipenode/.default={black},
  widenode/.style={
    rectangle, rounded corners=2pt, draw=#1, thick,
    fill=white, text=black, font=\small,
    minimum height=8mm, text width=30mm, align=center,
    inner sep=2pt
  },
  widenode/.default={black},
  keynode/.style={
    rectangle, rounded corners=3pt, draw=syngreen, very thick,
    fill=syngreen!12, text=black, font=\small\bfseries,
    minimum height=8mm, text width=22mm, align=center,
    inner sep=2pt
  },
  methodbox/.style={
    rectangle, rounded corners=2pt, draw=syngreen!70!black,
    fill=syngreen!4, font=\tiny, align=left,
    inner sep=3pt, text width=38mm
  },
  pathlabel/.style={
    font=\small\bfseries, text=#1
  },
  pathlabel/.default={black},
  annot/.style={font=\scriptsize, text=black!70, align=center},
  arr/.style={-{Latex[length=2.2mm]}, thick, #1},
  arr/.default={black!70},
  comparr/.style={-{Latex[length=2.2mm]}, thick, dashed, redorange},
]

\node[pipenode=black!60] (data)
  {Input data};

\node[widenode=black!60, below=4mm of data] (preproc)
  {Imputation \&\\[-1pt]discretisation};

\node[annot, below=1pt of preproc] (prepnote)
  {\scriptsize merge cats.\ 2\,\&\,3 $\to$ $k{=}3$};

\draw[arr] (data) -- (preproc);

\coordinate[right=14mm of preproc] (branch);
\draw[arr] (preproc) -- (branch);

\coordinate (branchA) at ($(branch)+(0,2.0)$);
\draw[arr=uqblue] (branch) -- (branchA);

\node[pathlabel=uqblue, anchor=south] at ($(branchA)+(-10mm, 10pt)$)
  {Path\,A: ePID};

\node[pipenode=uqblue, right=8mm of branchA] (pairA)
  {For each pair\\[-1pt]$(X_i, X_k)$};
\draw[arr=uqblue] (branchA) -- (pairA);

\node[keynode, right=10mm of pairA] (embed)
  {Compress\\[-1pt]remainder\\[-1pt]$E_{i\to k}$};
\draw[arr=syngreen!70!black] (pairA) -- (embed);

\node[annot, above=2pt of embed] (embnote)
  {$|E|=k$ states};

\node[pipenode=uqblue, right=10mm of embed] (pidA)
  {2-source PID\\[-1pt]$(X_i, E; X_k)$};
\draw[arr=uqblue] (embed) -- (pidA);

\node[pipenode=uqblue, right=10mm of pidA] (netA)
  {ePID\\[-1pt]network};
\draw[arr=uqblue] (pidA) -- (netA);

\draw[arr=uqblue!50, rounded corners=2pt]
  (netA.north) -- ++(0,0.4) -| node[annot, pos=0.25, above]
  {$\forall\;(i,k)$} (pairA.north);

\node[methodbox, below=6mm of embed, xshift=0mm] (methods) {%
  \textbf{13 embedding methods:}\\[2pt]
  \textit{Dim.\ reduction:} ACIB, SIR, PLS,
  MCA, NMF, SVD (+$k$-means)\\[1pt]
  \textit{Feature selection:} CMIM, JMI,
  greedy CMI, ReliefF\\[1pt]
  \textit{Clustering:} agglomerative,
  $k$-modes, spectral (Hamming)%
};

\draw[syngreen!70!black, thick, densely dashed]
  (embed.south) -- (methods.north -| embed.south);

\coordinate (branchB) at ($(branch)+(0,-2.0)$);
\draw[arr=redorange] (branch) -- (branchB);

\node[pathlabel=redorange, anchor=north] at ($(branchB)+(-10mm, -10pt)$)
  {Path\,B: ground truth};

\node[annot, text=redorange!80!black, anchor=north] at ($(branchB)+(-10mm, -24pt)$)
  {\textit{83 datasets + synthetic BNs}};

\node[pipenode=redorange, right=8mm of branchB] (pairB)
  {For each pair\\[-1pt]$(X_i, X_k)$};
\draw[arr=redorange] (branchB) -- (pairB);

\node[pipenode=redorange, right=10mm of pairB] (fullpid)
  {Full $N$-source\\[-1pt]PID};
\draw[arr=redorange] (pairB) -- (fullpid);

\node[annot, below=1pt of fullpid] (fmtnote)
  {FMT for $N{\geq}5$};

\node[pipenode=redorange, right=10mm of fullpid] (amalg)
  {Amalgamate to\\[-1pt]2-source atoms};
\draw[arr=redorange] (fullpid) -- (amalg);

\node[pipenode=redorange!50!uqblue, right=10mm of amalg,
      text width=20mm] (compare)
  {Compare\\[-1pt]$|\Delta\mathrm{AE}|$};
\draw[arr=redorange] (amalg) -- (compare);

\draw[comparr, rounded corners=3pt]
  (pidA.south) -- ++(0,-0.4) -| (compare.north);

\node[annot, below=2pt of compare]
  {select best\\[-1pt]embedding};

\end{tikzpicture}%
}
\caption{\apacap{Overview of the ePID evaluation pipeline}{An end-to-end view of how raw symptom data become a PID-decomposed network: a scalable embedding-based path used for the main analyses runs in parallel with a small-scale ground-truth path used to validate it. Raw data are preprocessed (imputation, discretisation) and then processed along these two paths. \emph{Path A (top, blue):} for each source--target pair, the remainder symptoms are compressed into a discrete embedding using one of 13 methods, and a two-source PID is computed on (source, embedding, target). \emph{Path B (bottom, orange):} for benchmark validation, the full $N$-source PID is computed via the fast M\"obius transform (FMT) and amalgamated to two-source atoms for comparison with Path A.}}
\label{fig:pipeline}
\end{figure}

\begin{figure*}[!t]
  \centering
  \includegraphics[width=\textwidth]{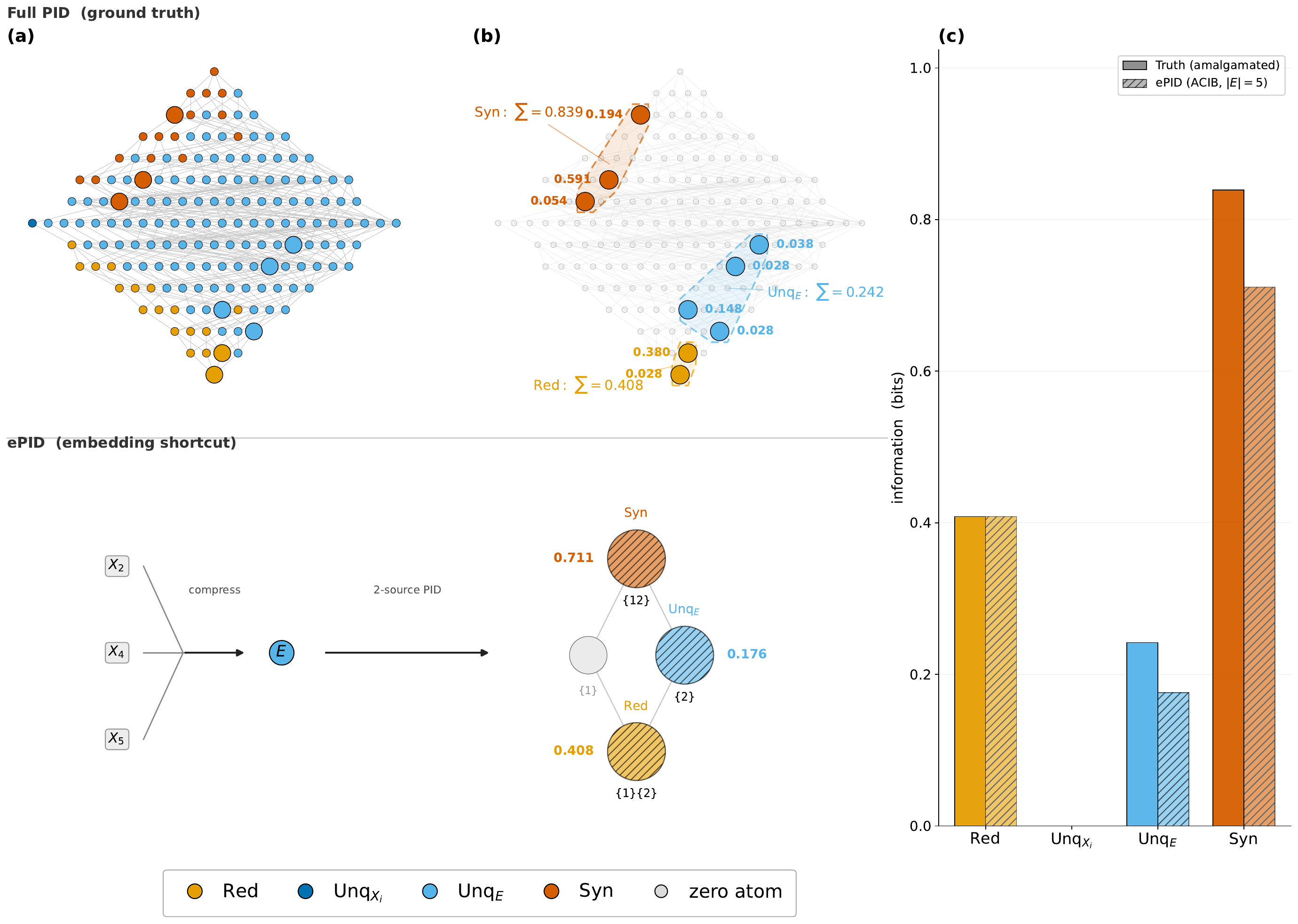}
  \caption{\apacap{ePID on the toy SCM: redundancy lattice, amalgamated atoms, truth-versus-ePID recovery, and the embedding shortcut}{Same SCM as Figure~\ref{fig:method_comparison}; visualises the four-source PID lattice that the main pipeline amalgamates over. \textbf{(a)} Full $N=4$ antichain lattice for $\{X_1, X_2, X_4, X_5\} \to X_3$ under $I_{\min}$: 166 atoms in total, of which 9 carry non-zero information (sum $=$ 1.489 bits). Node area scales with atom value; colour encodes the type-1 amalgamation class. \textbf{(b)} The nine non-zero atoms grouped by their type-1 amalgamation class (synergy $\Sigma = 0.839$, remainder-unique $\Sigma = 0.242$, redundancy $\Sigma = 0.408$ bits), each labelled with its individual value. \textbf{(c)} Amalgamated four-atom view for focal source $X_1$: ground truth versus ePID (ACIB embedding, $\lvert E \rvert = 5$); the arrow marks the amalgamation step that collapses the lattice into four atoms. ACIB recovers $\mathrm{Red}$ exactly, $\mathrm{Syn}$ at $85\%$, and $\mathrm{Unq}_E$ at $73\%$, for $87\%$ of total joint information. The lower row makes the ePID shortcut explicit: rather than enumerating the full lattice, the remaining sources $\{X_2, X_4, X_5\}$ are compressed into a single embedding $E$ and a two-source PID is computed on $(X_1, E; X_3)$, recovering the amalgamated atoms directly (synergy $0.711$, remainder-unique $0.176$, redundancy $0.408$ bits). Grey nodes carry zero information. ACIB $=$ Agglomerative Conditional Information Bottleneck.}}
  \label{fig:method_3panel}
\end{figure*}

\subsection{Multi-dataset embedding benchmark}
\label{subsec:multi_dataset_benchmark}

The synthetic Bayesian network benchmark (Section~\ref{subsec:embedding_benchmark}) provides controlled validation with known ground truth but is limited to a single data-generating mechanism calibrated to PHQ-9. To assess generalisability, we additionally evaluated embedding fidelity across 83 real-world datasets spanning clinical, epidemiological, and psychometric domains.

\paragraph{Dataset assembly and preprocessing.}
We assembled 83 datasets from publicly available repositories spanning clinical, epidemiological, and psychometric domains, with 5--56 variables per dataset and sample sizes ranging from 32 to 445{,}000 observations. Full details of dataset assembly, preprocessing, imputation, and discretisation are provided in Appendix~\ref{app:datasets}. In total, the benchmark comprises 2{,}477{,}784 individual comparisons across 5 PID measures, 83 datasets, 13 embedding methods, and $N \in \{3,4,5\}$ sources.

\paragraph{Embedding hyperparameters.}
The 13 embedding methods were configured as follows: ACIB used a loss tolerance of 5\% relative information loss with maximum embedding cardinality $K_{\max}=12$; clustering methods ($k$-modes, spectral Hamming, agglomerative Hamming) used $K=6$ clusters; SIR and PLS used 2 latent dimensions with $B=4$ bins; MCA, SVD, and NMF used 2 components with $K=6$ $k$-means clusters; ReliefF used 100 neighbours with $k=3$ selected features; JMI and CMIM selected $k=3$ features. For distance-based methods (spectral and agglomerative Hamming), datasets with more than 5{,}000 observations were subsampled to 5{,}000 rows for distance matrix computation. All methods used a fixed random seed of 42 for reproducibility.

\paragraph{Error metrics.}
For each embedding method and dataset, we computed the ground-truth $N$-source PID via the fast M\"obius transform~\citep{jansma2025fastmobius} and amalgamated it to a bivariate decomposition using the conservative amalgamation scheme described in Section~\ref{subsec:embedding_benchmark}. Embedding error was quantified using the total absolute error (TAE) and relative error (RE) defined in Section~\ref{subsec:embedding_benchmark}.

\subsection{Multi-measure robustness framework}
\label{subsec:multi_measure}

PID decompositions are measure-dependent: the choice of redundancy function determines how atoms are partitioned. To assess the robustness of embedding fidelity across different PID axiomatisations, we computed ground-truth $N$-source PIDs using the five redundancy measures introduced in Section~\ref{subsec:pid_two_source} ($I_{\mathrm{mmi}}$, $I_{\min}$, $I_{\mathrm{rr}}$, $I_{\wedge}$, $I_{\pm}$); we adopt $I_{\mathrm{mmi}}$ as the primary measure throughout and treat the remaining four as benchmark axes here.

For $N \geq 5$ sources, $I_{\pm}$ and $I_{\wedge}$ were computed via the fast M\"obius transform~\citep{jansma2025fastmobius} using precomputed antichains and M\"obius matrices for the redundancy lattice, reducing per-target computation from hours to minutes.

For each dataset, target, number of sources $N \in \{3,4,5\}$, and PID measure, we computed the full $N$-source ground-truth PID, amalgamated the resulting atoms onto two-source equivalents using the conservative scheme described in Section~\ref{subsec:embedding_benchmark}, and compared these against the corresponding embedding-based two-source PID. This yields error profiles stratified by embedding method, PID measure, and number of sources, enabling a systematic assessment of which embedding methods are robust across different axiomatisations of redundancy and synergy.

\paragraph{Dataset coverage.}
Coverage was identical across all five measures: complete ground-truth PIDs were available for 83 datasets at $N=3$, 79 at $N=4$, and 77 at $N=5$, the shortfall at $N=4$ arising from four NHANES datasets that lack sufficient variables.

\subsection{Normalisation and effect-size scaling}
\label{sec:normalisation}
Because information-theoretic measures are expressed in bits and are bounded above by the uncertainty of the target, we report normalised effect sizes---a target-normalised CMI ($\mathrm{nCMI}$) and entropy-share PID atoms---to facilitate comparisons across targets and cohorts; the definitions, equations, and the network-visualisation pruning rule are given in Appendix~\ref{app:normalisation}.

\subsection{Significance testing for pairwise measures}
\label{sec:significance_testing}
Statistical significance of pairwise conditional-dependence edges (PC and CMI) was assessed using permutation tests. For each source--target pair $(X_i, X_j)$, the target column $X_j$ was permuted across participants while all other columns (including $X_i$ and the conditioning set $Z_{ij}$) were held fixed. This preserves the joint distribution of the non-target variables while destroying the focal source--target association. The pairwise statistic was recomputed on each of $B = 1{,}000$ permuted datasets to form an empirical null distribution. The one-sided $p$-value was computed as $p = (m + 1) / (B + 1)$, where $m$ is the number of replicates with a test statistic at least as large as the observed value. Multiple testing across all edges was controlled using the Benjamini--Hochberg false discovery rate (FDR) procedure at $q = 0.05$~\citep{benjamini1995fdr}.

\paragraph{Cross-cohort stability.}
To assess replicability of the PID decomposition, we match directed source--target edges that are significant in both UK Biobank and the Xinxiang student sample and compute Spearman rank correlations of PID atom fractions (redundancy, synergy, remainder-unique) across matching edges. Mean absolute differences between dataset-specific fractions are also reported.


\section{Results}

We present results in four stages. First, we benchmark candidate embeddings against ground-truth PID on synthetic Bayesian networks calibrated to PHQ-9. Second, we extend that benchmark across 83 real-world datasets and five PID measures to assess generalisability. Third, we apply the validated pipeline to the empirical PHQ-9 networks in UK Biobank and the Xinxiang student sample, comparing pairwise conditional-dependence estimates from partial correlations and CMI before decomposing each source--target dependence via ePID into redundancy, synergy, and remainder-unique channels. Fourth, we apply the same pipeline to the Interpersonal Reactivity Index as a contrasting psychometric instrument.

\subsection{Embedding benchmark for ePID}
\label{subsec:results_embedding_benchmark}

The central question for the benchmark is whether this compression preserves the resulting PID atoms. This subsection serves to evaluate the quality of preserving PID atoms for different embedding algorithms and determine the optimal embedding algorithm for all subsequent analyses.

To determine which embedding methods provide reliable approximations to high-order PID, we evaluated 13 candidates on an ensemble of 50 Bayesian networks $\{G_1,\ldots,G_{50}\}$ calibrated to the joint distribution of the PHQ-9 items. BN calibration statistics are shown in Figures~\ref{fig:mi_distribution_comparison}, \ref{fig:detailed_validation}.
For each network $G$, each target $X_t$, and each number of sources $N\in\{3,4,5\}$, we computed (i) the exact $N$-source PID using the MMI-based redundancy measure in \texttt{dit}, and (ii) a family of two-source PIDs where one source was a focal variable and the other was a discrete embedding of the remaining $N-1$ sources.
We then amalgamated the ground-truth $N$-source atoms onto the four components of a two-source PID and quantified the absolute error for each atom (source-unique, remainder-unique, redundancy, synergy) and each embedding method (Methods~\ref{subsec:embedding_benchmark}).
This synthetic benchmark uses MMI (minimum mutual information) throughout, both for the ground-truth $N$-source decomposition and for the embedding-based approximation, so comparisons are internally consistent. Because these results characterise approximation fidelity conditional on a given PID measure, they do not speak to the separate question of which measure best captures the information structure of interest; we return to measure dependence in the Discussion (Section~\ref{subsec:limitations}). Cross-measure generalisation, comparing each of five PID measures against its own ground truth, is evaluated separately in Appendix~\ref{subsec:results_multidataset}.

\begin{figure}[htbp]
  \centering
  \includegraphics[width=\textwidth]{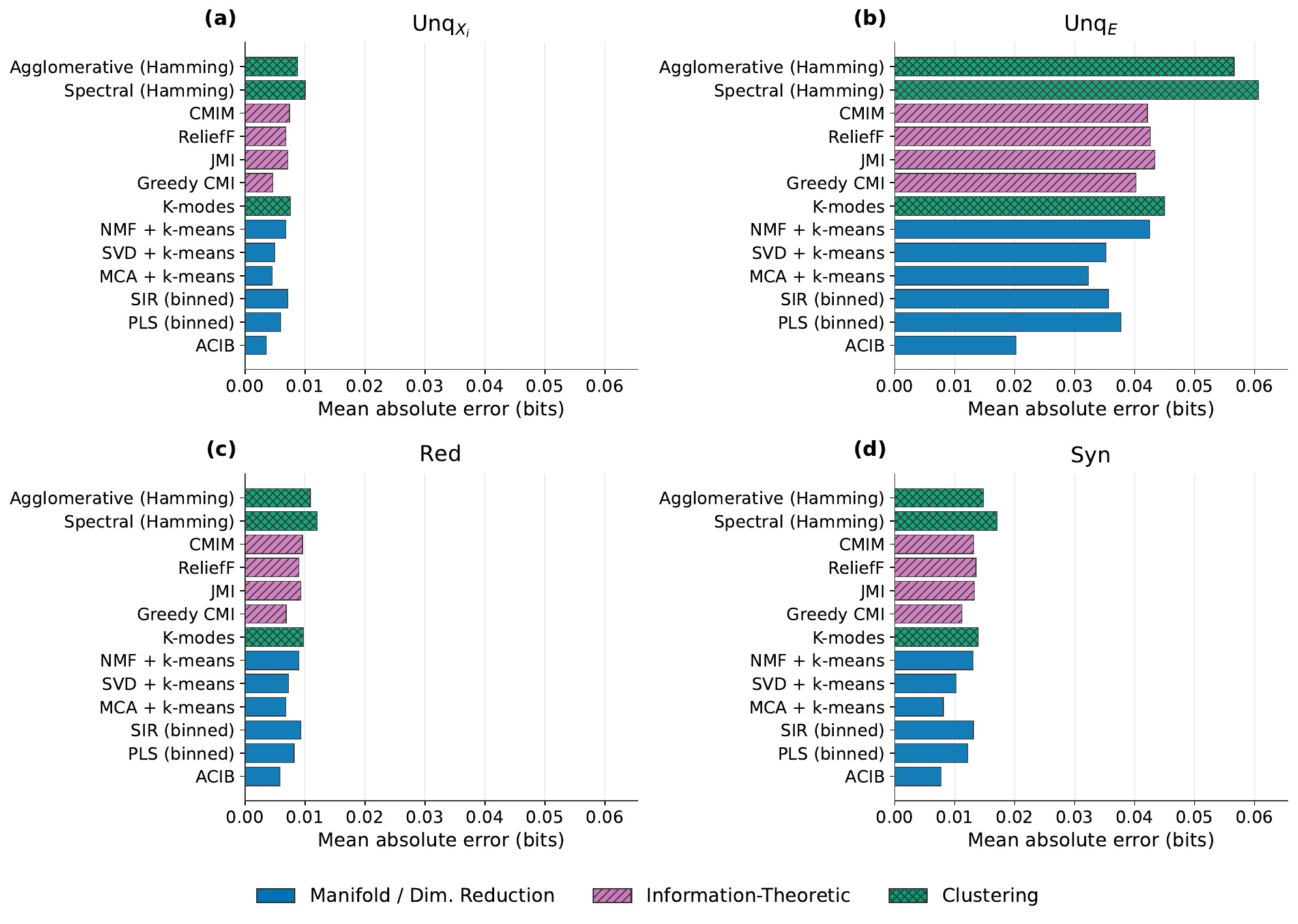}
  \caption{\apacap{Mean absolute error by embedding method and PID atom}{Each panel shows mean absolute error (in bits) between the amalgamated ground-truth $N$-source PID and the embedding-based two-source PID approximation, for one of the four PID atoms: \textbf{(a)} source-unique $\mathrm{Unq}_{X_i}$, \textbf{(b)} remainder-unique $\mathrm{Unq}_{E}$, \textbf{(c)} redundancy $\mathrm{Red}$, \textbf{(d)} synergy $\mathrm{Syn}$. Methods are coloured by family: \emph{manifold / dim-reduction} (blue: ACIB, SIR, PLS, MCA+$k$-means, SVD+$k$-means, NMF+$k$-means), \emph{information-theoretic feature selection} (pink: CMIM, JMI, ReliefF, Greedy CMI), and \emph{clustering} (green: Agglomerative Hamming, Spectral Hamming, $k$-modes). Errors are pooled across 50 synthetic Bayesian networks calibrated to PHQ-9, nine targets, and three source counts ($N \in \{3, 4, 5\}$); within each panel, methods are sorted by mean error. Within the manifold family, ACIB (the recommended default) achieves the lowest error across all four atoms. The clustering family is heterogeneous: $k$-modes performs comparably to manifold methods on $\mathrm{Red}$ and $\mathrm{Unq}_{X_i}$, while Hamming-based clustering collapses $\mathrm{Syn}$ toward zero (see also Figure~\ref{fig:method_ranking}). ACIB $=$ Agglomerative Conditional Information Bottleneck; JMI $=$ Joint Mutual Information; CMIM $=$ Conditional Mutual Information Maximisation.}}
  \label{fig:embedding_overall_performance}
\end{figure}

\paragraph{Method-family stratification}
Errors stratified strongly by embedding family (Figure~\ref{fig:embedding_overall_performance}).
Manifold and dimensionality-reduction methods (ACIB, SIR, PLS, MCA+$k$-means, SVD+$k$-means, NMF+$k$-means) achieved the lowest overall mean absolute error across components ($\approx 0.014$ bits), ahead of information-theoretic feature selection approaches (CMIM, JMI, greedy CMI, ReliefF; mean error $\approx 0.018$ bits) and clustering methods (agglomerative Hamming, spectral Hamming, $k$-modes; mean error $\approx 0.022$ bits).
Among all methods, the Agglomerative Conditional Information Bottleneck (ACIB) embedding showed the highest overall fidelity. At $N=5$, ACIB achieved a mean total absolute error (summed across all four PID atoms) of $0.048$ bits, or about $19\%$ of the total mutual information (TMI); synergy-specific errors remained small in both mean ($0.0095$ bits) and median ($0.008$ bits).

\paragraph{Component-resolved performance.}
Most methods reproduced the source-unique and redundancy atoms with near-numerical precision (median errors $<10^{-10}$ bits for 11 of 13 methods), whereas discrepancies concentrated in the remainder-unique and synergy atoms (bottom panels of Figure~\ref{fig:embedding_overall_performance}). A relative-error view of the same data, in which per-atom errors are normalised by the magnitude of the corresponding ground-truth atom, is reported in Appendix Figure~\ref{fig:component_errors_re}.
For these challenging atoms, ACIB again performed best: the median error for the remainder-unique atom was $0.0163$ bits and the median synergy error was $0.0058$ bits, whereas typical feature-selection methods exhibited larger median synergy errors. We report these errors as fractions of TMI, the exact ground-truth total mutual information for each edge (the sum of its four ground-truth atoms), rather than as fractions of each atom's own true magnitude, because the synergy and remainder-unique atoms are themselves near-zero in many of the calibrated networks: a per-atom denominator is then numerically unstable and can inflate relative errors without bound, whereas TMI provides a stable common scale across atoms. The complementary per-atom relative-error view is the convention used in Appendix Figure~\ref{fig:component_errors_re}, where exactly this instability is visible for the near-zero atoms.
This pattern indicates that the main difficulty in approximating PID lies in preserving higher-order and source-specific information; embeddings that explicitly optimise information preservation fare better than those that optimise only geometric cohesion or pairwise criteria.

\paragraph{Synergy fidelity across method families.}
Per-pair comparison of approximated versus ground-truth synergy illustrates the family-level spread (Appendix Figure~\ref{fig:embedding_scatter}).
For ACIB (manifold/dim-reduction; Pearson $r=0.92$), synergy estimates aligned closely with the identity line, indicating that the one-dimensional embedding captured the range of synergistic information present in the original multi-source systems.
For JMI (feature-selection; $r=0.76$), estimates were more scattered around the identity line but no longer strongly attenuated.
For $k$-modes (clustering; $r=0.75$), estimates aligned reasonably well, illustrating the upper end of clustering performance.
Under this calibration the Hamming-based agglomerative and spectral clustering methods no longer collapse synergy toward zero (agglomerative $r=0.68$, spectral $r=0.62$; see Figure~\ref{fig:method_ranking}); they recover a moderate share of the synergistic structure, though still less than the manifold methods.

Feature selection methods (CMIM, JMI, greedy CMI, ReliefF) showed a distinct failure mode.
For systems with three sources, these methods approximated synergy almost perfectly (errors near machine precision). This is expected: with $N=3$ sources the remainder consists of $N-1=2$ variables, and feature-selection embeddings that retain up to two features can store both losslessly, so the joint state of the remainder is preserved exactly and the resulting two-source PID coincides with the three-source ground truth. The embedding step is therefore active (two sources are still compressed into a single discrete embedding for the PID), but no information is lost in that compression. When the number of sources increased to five, synergy errors rose to $\approx 0.017$ bits, a more modest increase, but still indicating that methods which select or weight features via greedy, pairwise, or univariate criteria capture the emergent joint information less completely as more sources interact simultaneously.

\paragraph{Scaling with the number of sources.}
We next examined how approximation error changed with the number of sources $N$.
For ACIB, mean synergy error increased smoothly from $0.0046$ bits at $N=3$ to $0.0095$ bits at $N=5$, remaining below $0.01$ bits throughout. Because the UK Biobank-calibrated networks carry little synergy per edge, these small absolute errors correspond to a larger fraction of the (small) per-edge synergy: the synergy atom's median per-edge error, normalised by each edge's synergy, rose from about $21\%$ at $N=3$ to $76\%$ at $N=5$. We therefore emphasise the absolute error, which stays below $0.01$ bits; the relative figure is inflated by the near-zero synergy denominator in these low-synergy networks.
A sliced inverse regression (SIR) embedding with binning showed synergy error that increased with $N$ (from $\approx 0.008$ bits at $N=3$ to $\approx 0.016$ bits at $N=5$), comparable to the other manifold methods, whereas clustering methods had both larger baseline errors and steeper increases with $N$.
Thus, ACIB provides the best absolute fidelity, while SIR offers the most stable performance as systems become more complex. The extended multi-dataset benchmark (Appendix~\ref{subsec:results_multidataset}) confirms that ACIB's advantage persists at higher $N$, with nominally the shallowest error growth slope among all methods.

\paragraph{Embedding choice for symptom networks.}
Based on these benchmarks, we use ACIB as the primary embedding in all empirical symptom-network analyses.
The manifold/dimensionality family consistently preserved the structure of unique, redundant, and synergistic information with low error, whereas clustering and feature-selection methods either underestimated synergy or collapsed it toward zero.
The main text therefore focuses on ACIB-based two-source PIDs.

To confirm that embedding-based ePID generalises beyond the PHQ-9-calibrated synthetic networks, we ran an extended benchmark of all 13 embeddings across 83 real-world datasets (clinical, epidemiological, and psychometric; 5--56 variables) and all five PID measures at $N\in\{3,4,5\}$ sources---over 2.4 million atom-level comparisons. The best-performing embedding varied from dataset to dataset, but a single manifold-family method, ACIB, was the most consistent choice, ranking first on 75--84\% of datasets at $N=5$ for $I_{\min}$, $I_{\mathrm{mmi}}$, and $I_{\wedge}$; we adopt it as the default and report the full method-by-dataset comparison in Appendix~\ref{subsec:results_multidataset}. The one measure no embedding approximates well is the signed $I_{\pm}$, whose pointwise structure does not survive compression.

\subsection{PHQ-9 symptom networks}
\label{subsec:phq9_results}

Having validated the embedding pipeline on synthetic and real-world data, we apply it to the empirical PHQ-9 networks in UK Biobank and the Xinxiang student sample. We report two complementary views of the same data. Section~\ref{subsec:pairwise_results} characterises pairwise conditional dependence using partial correlations and CMI; Section~\ref{subsec:phq9_network_context} then decomposes those dependencies via ePID. Reporting both side by side allows the ePID atoms to be read against the conventional partial-correlation network used elsewhere in the symptom-network literature.

\subsubsection{Pairwise conditional dependence: partial correlations vs conditional mutual information}
\label{subsec:pairwise_results}

We found that PHQ-9 items in both cohorts have largely monotonic conditional dependence at the level detectable by Spearman partial correlation (SPC), while CMI provides a model-free scale in bits for the same conditional-dependence structure. Figure~\ref{fig:xinxiang_PC_PC_CMI} (Appendix~\ref{app:cmi_pid_comparison}) compares pairwise conditional-dependence estimates derived from partial correlations and conditional mutual information (CMI) in the Xinxiang student sample. Pearson (PPC) and Spearman (SPC) partial correlations were highly similar at the edge level (Fig.~\ref{fig:PPC_vs_PSP}), and both aligned closely with CMI. Permutation tests with false discovery rate control ($q=.05$) yielded statistically significant CMI for 56/56 directed source--target pairs in UK Biobank and 54/56 in the Xinxiang student sample (though many of the UK Biobank edges have nCMI well below 0.005, reflecting the high power afforded by $N \approx 154{,}000$; see Supplementary Table~\ref{tab:edge_overlap_pruning} for effect-size–pruned counts).

In the Xinxiang student sample the CMI and Spearman partial-correlation networks were comparably dense (54 significant directed CMI edges versus 54 for SPC at $q = .05$), indicating that monotonic partial correlations and CMI flag a similar set of conditional dependencies at the item level. Normalised CMI values ($\mathrm{nCMI} = I(X;Y \mid \mathbf{Z})/H(Y)$) ranged from 0.02 to 0.07 in the Xinxiang student sample and from 0.0003 to 0.06 in UK Biobank. For context, the largest nCMI in the Xinxiang student sample indicates that the source symptom carries conditional information about the target amounting to about 7\% of the target's marginal entropy, after accounting for the rest of the network; most values are far smaller. Conditional dependencies at the item level are therefore modest in absolute magnitude.

\subsubsection{Network-context decomposition via ePID}
\label{subsec:phq9_network_context}

We now decompose each conditional dependence into redundancy, synergy, and remainder-unique channels, asking whether this finer-grained view reveals structure that pairwise methods cannot express.

\paragraph{Aggregate atom distributions and cross-cohort stability.}
The composition of source--target information was similar in UK Biobank and the Xinxiang student sample: redundancy accounted for 51\% of total mutual information (TMI) in UK Biobank and 45\% in the Xinxiang student sample, synergy for 6\% and 8\%, and remainder-unique information for 42\% and 47\%, respectively. Under MMI with a target-optimised embedding the source-unique atom was zero or near-zero for all edges, so the decomposition effectively partitions dependence into three non-trivial channels: redundancy, synergy, and remainder-unique. We report atom fractions as shares of TMI, $I(X_i, E_{i\to k}; X_k)$, so that shares sum to one for each edge; target-entropy-normalised shares (Eq.~\ref{eq:pid_shares}) are reported alongside in bits when needed. Edge-level ranges were also comparable across datasets: redundancy 11--79\% (UK Biobank) and 19--68\% (Xinxiang student sample), synergy 0.4--28\% and 1--20\%, and remainder-unique 20--79\% and 13--80\%.

The cross-dataset similarity of atom shares was matched by edge-level concordance. Across the 54 directed edges that were significant in both datasets (FDR-corrected $q < .05$), PID atom fractions were positively correlated (Spearman $\rho$: redundancy $= 0.59$; synergy $= 0.78$; remainder-unique $= 0.68$), with mean absolute differences in atom share of $0.05$, $0.02$, and $0.06$ (i.e.\ $5$, $2$, and $6$ percentage points of TMI) respectively. The redundancy concordance ($\rho = 0.59$) is more modest than the synergy concordance. Together, these aggregate and edge-level convergences indicate that the relative importance of redundant, synergistic, and remainder-unique channels is a replicable feature of PHQ-9 symptom networks rather than a property of either dataset.

\begin{figure}[ht]
    \centering
    \includegraphics[width=\linewidth]{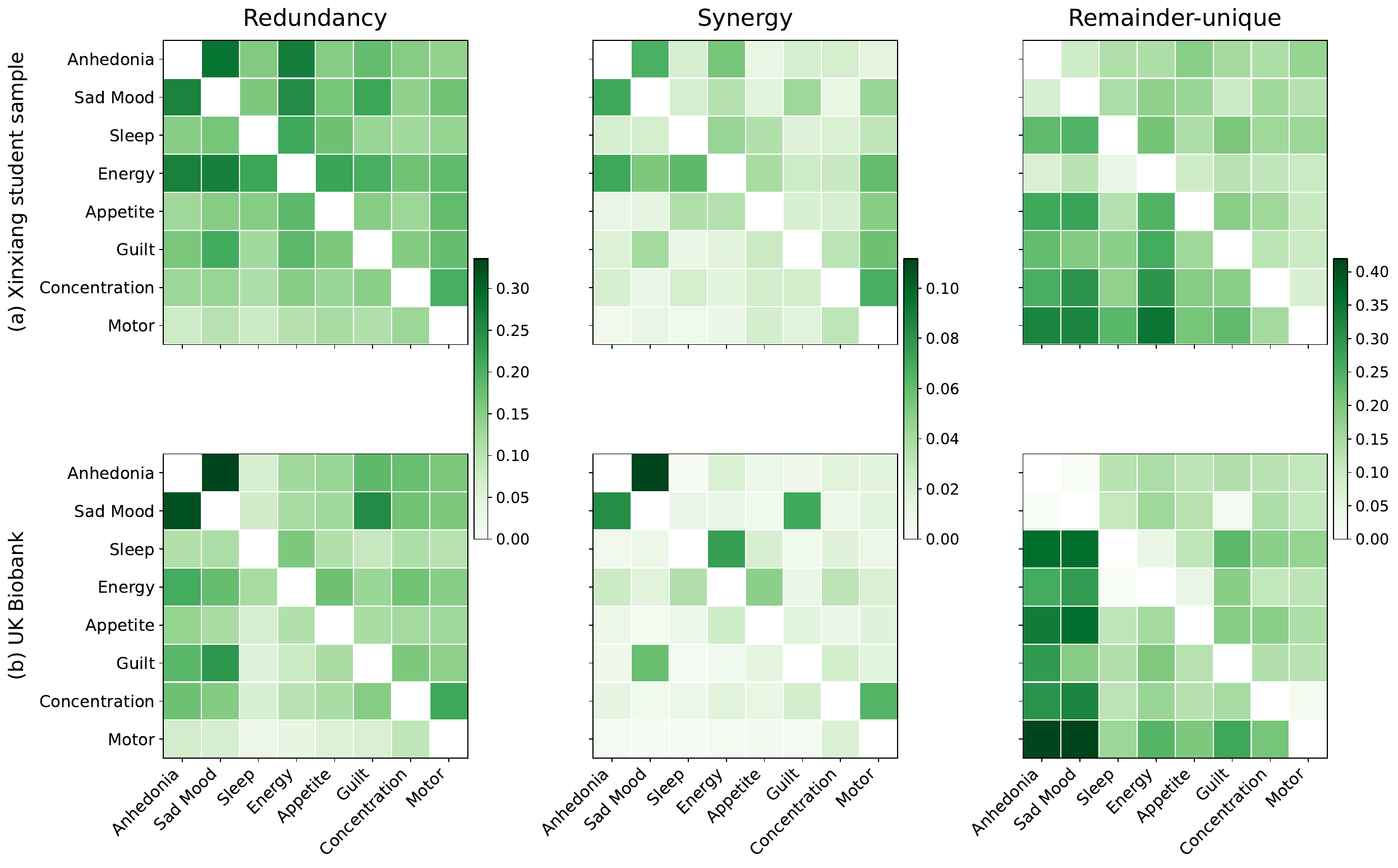}
    \caption{\apacap{Entropy-normalised PID atom shares across source--target pairs}{Each cell is the atom fraction $I_\mathrm{atom}(X\to Y)/H(Y)$ for a directed pair among the eight PHQ-9 symptoms (suicidality excluded); rows are sources, columns are targets, and diagonals are blanked. Top row (a): Xinxiang student sample; bottom row (b): UK Biobank. Columns show the three non-trivial PID channels under ACIB (redundancy, synergy, and remainder-unique) with shared colour scales within each atom to make cross-cohort comparisons visible. The source-unique atom is zero or near-zero across all edges under MMI with a target-optimised embedding (see text above) and is omitted; the conditional-mutual-information share at the edge level is visualised in Fig.~\ref{fig:pid_network_top3_per_target}.}}
    \label{fig:PID_distributions_acib}
\end{figure}

\paragraph{Network visualisation.}
To make the PID networks interpretable at a glance, we report two complementary visualisations. In the \emph{full} view (Fig.~\ref{fig:Xinxiang_Two_PIDs_withUnique}, Appendix~\ref{app:cmi_pid_comparison}), each directed edge $X\to Y$ is segmented into remainder-unique, redundancy, and synergy; edge thickness encodes the target-normalised total information $I(X, E_{X\to Y};Y)/H(Y)$. The \emph{no-remainder-unique} view (Fig.~\ref{fig:Xinxiang_Two_PIDs_noUnique}) removes the remainder-unique atom and rescales the remaining segments, so that colour differences directly compare redundant versus synergistic source involvement. Read together, these views distinguish targets that are predictable primarily from the broader symptom context from targets where the focal symptom participates through overlap (redundancy) or interaction-only structure (synergy).

\begin{figure}[!ht]
\centering
\begin{subfigure}[t]{0.49\linewidth}
\centering
\includegraphics[width=\linewidth]{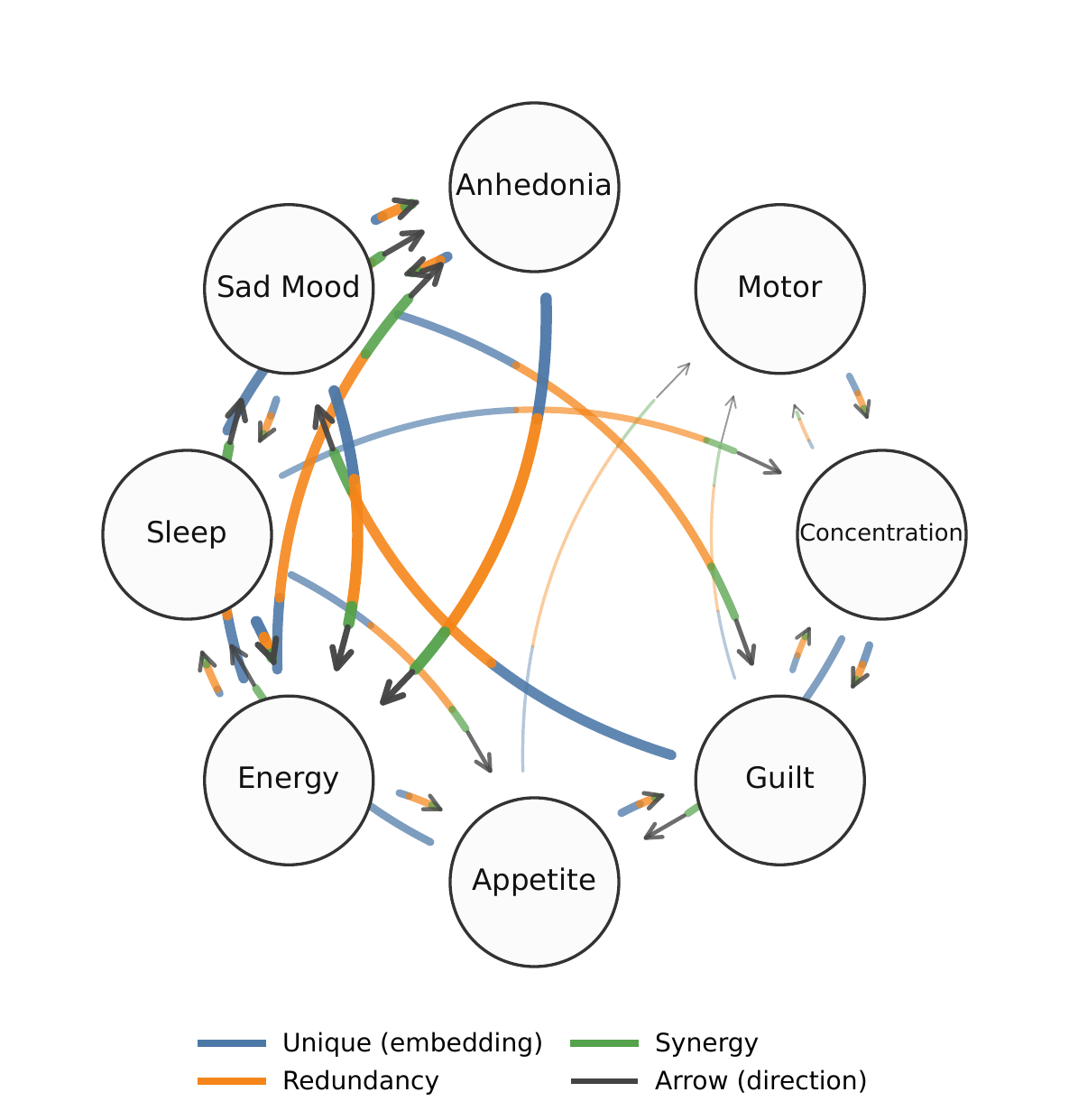}
\caption{Xinxiang student sample}
\end{subfigure}\hfill
\begin{subfigure}[t]{0.49\linewidth}
\centering
\includegraphics[width=\linewidth]{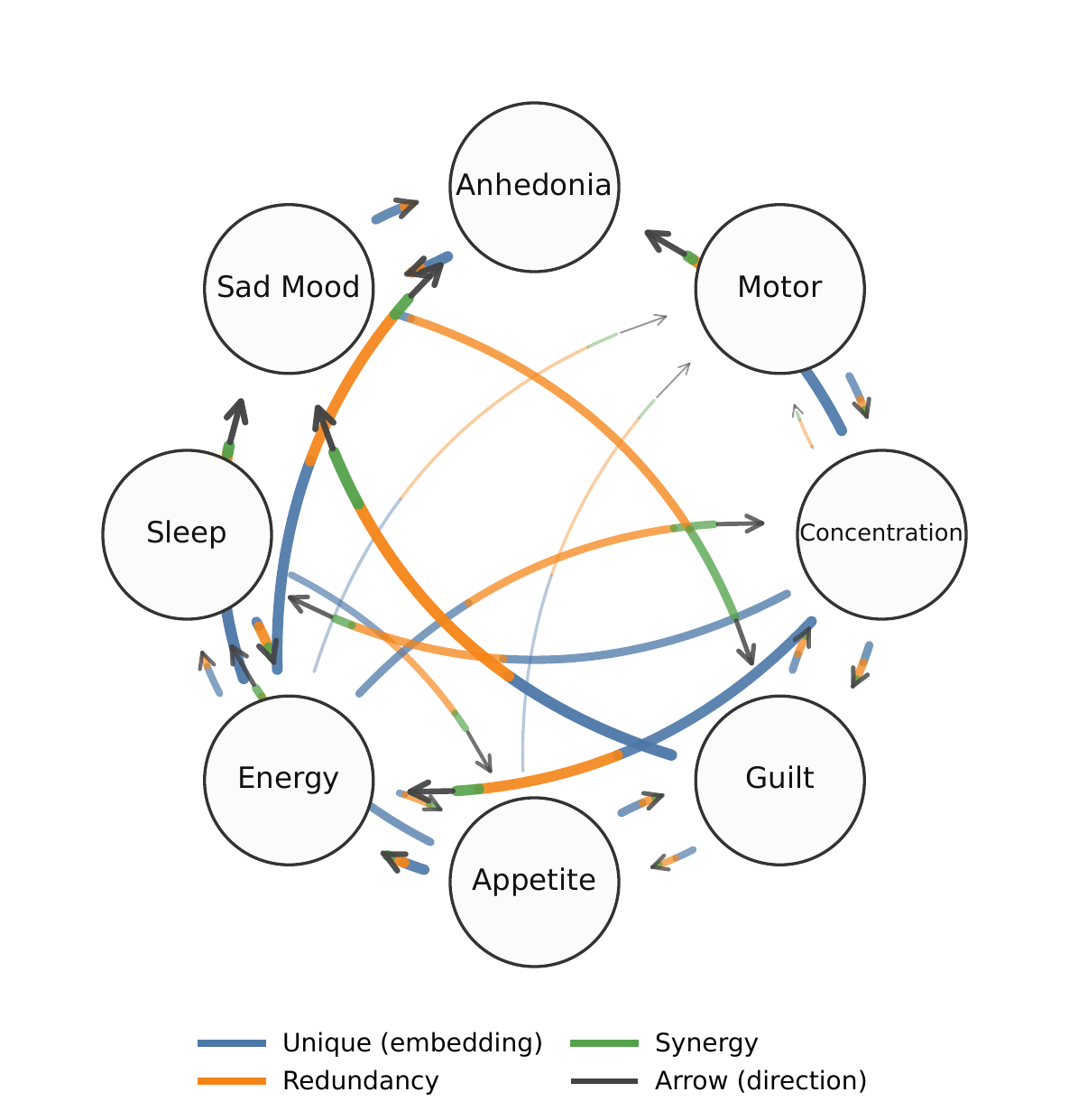}
\caption{UK Biobank}
\end{subfigure}
\caption{\apacap{Replication of ePID networks (top-3 incoming edges per target) in UK Biobank and the Xinxiang student sample}{Nodes are eight PHQ-9 symptoms (suicidality excluded).
Edges were first screened using permutation-tested conditional mutual information (CMI) with Benjamini--Hochberg FDR control ($q=.05$) and then pruned for visualisation by retaining, for each target symptom $Y$, the three incoming edges $X\rightarrow Y$ with the largest entropy-normalised conditional mutual information
$\mathrm{nCMI}_{X\rightarrow Y}=I(X;Y\mid\mathbf{Z})/H(Y)$.
For each retained directed edge, we compute an embedding-based two-source partial information decomposition (PID) on $(X, E_{X\rightarrow Y}; Y)$ using an ACIB embedding of the remaining symptoms.
Edge thickness is proportional to total mutual information $I(X, E_{X\rightarrow Y}; Y)$. Coloured segments indicate the fraction of this total carried uniquely by the embedded remainder (blue), redundantly by $X$ and the remainder (orange), or synergistically (green); a terminal black segment indicates direction. Edges were selected for display using CMI; decompositions for all 56 directed pairs are reported in Appendix~\ref{app:cmi_pid_comparison}. Because nCMI conditions on the full remaining symptom set, it captures the source's unique-plus-synergistic information and excludes redundancy; since the source-unique atom is near zero throughout these networks, the top-three-per-target display, if anything, emphasises synergistic edges rather than hiding them. The reported atom fractions (Fig.~\ref{fig:PID_distributions_acib}) are in any case computed from all 56 directed decompositions and all informative edges, not only the displayed subset.}}
\label{fig:pid_network_top3_per_target}
\end{figure}

Figure~\ref{fig:pid_network_top3_per_target} shows the resulting networks for UK Biobank and the Xinxiang student sample.
PID decomposes each retained source--target relation into overlapping versus interaction-only channels. Across retained edges, most dependence is carried by redundant and remainder-unique components, indicating that the apparent influence of a focal symptom on a target typically reflects information already present in the surrounding symptom context. Nonetheless, several edges display substantial synergistic components, consistent with the presence of interaction-dependent configurations in which a symptom becomes informative about a target primarily in conjunction with the remainder of the network.

\subsection{Depression-status contrast in PHQ-9 network composition}
\label{subsec:dep_vs_nondep}

\begin{figure}[htbp]
\centering
\begin{tikzpicture}
  \node[anchor=south west,inner sep=0] (img) at (0,0)
    {\includegraphics[width=\textwidth]{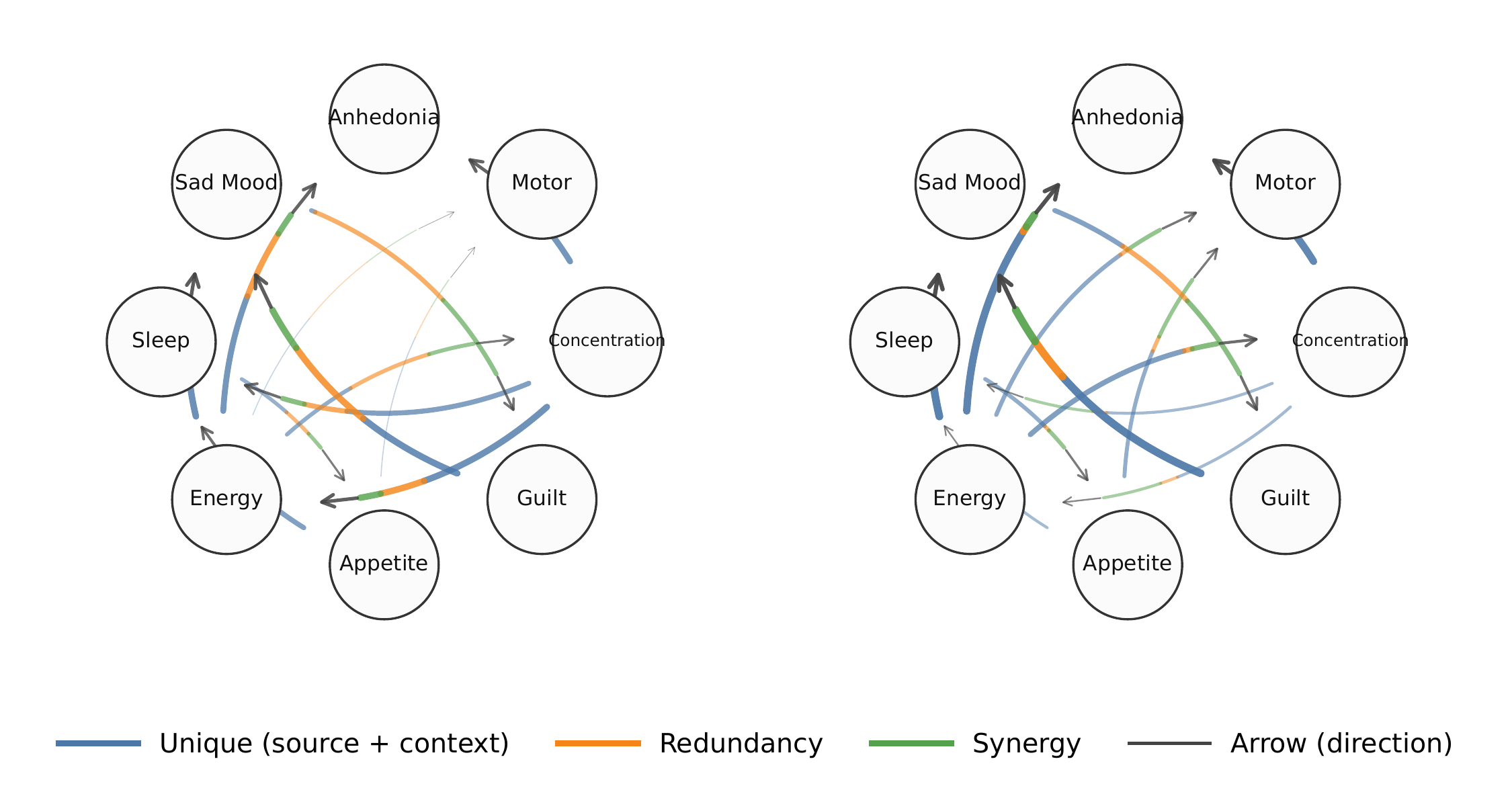}};
  \node[anchor=north west,font=\bfseries] at ([xshift=3pt,yshift=-3pt]img.north west) {(a)};
  \node[anchor=north west,font=\bfseries] at ([xshift=3pt,yshift=-3pt]img.north) {(b)};
\end{tikzpicture}
\caption{\apacap{ePID symptom networks in non-depressed versus depressed UK Biobank respondents}{Directed embedding-PID networks for the eight retained PHQ-9 symptoms (suicidality excluded), estimated separately by depression status. \textbf{(a)} non-depressed (PHQ-9 total $<10$); \textbf{(b)} depressed (PHQ-9 total $\geq 10$). The groups were equal-$N$ matched at $n=8{,}879$ per arm and each decomposition was averaged over five random subsamples of the larger group. Both panels display the same 24-edge skeleton (the top three incoming edges per target by entropy-normalised CMI, pooled across groups). For each directed edge, coloured segments partition the pair--target total mutual information into unique information carried by the source plus embedded remainder context (blue), redundancy (orange), and synergy (green); a terminal black segment indicates direction, and edge thickness is proportional to total mutual information on a scale shared across both panels. Relative to non-depressed respondents, the depressed network shifts from redundancy toward unique and synergistic information (mean shares: redundancy $0.34\rightarrow0.15$, unique $0.47\rightarrow0.59$, synergy $0.19\rightarrow0.26$); the clearest visual signal is the contraction of the orange (redundancy) segments. Because the absolute synergy share is sample-size dependent (full-cohort PHQ-9 synergy $\approx0.08$), the figure communicates the matched-$N$ \emph{contrast} between groups rather than an absolute synergy level. All 56 significant directed edges are shown in Fig.~\ref{fig:dep_vs_nondep_alledges}.}}
\label{fig:dep_vs_nondep}
\end{figure}

To test whether the compositional balance of the PHQ-9 network depends on symptom severity, we split the UK Biobank sample by the standard PHQ-9 cutoff (total score $\geq 10$ versus $<10$) and estimated the ePID network separately in each group. To remove sample size as a confound, the two arms were equal-$N$ matched at $n = 8{,}879$ per group and each decomposition was averaged over five random subsamples of the larger (non-depressed) group. Relative to non-depressed respondents, the depressed network shifts away from redundancy toward unique and synergistic information (mean shares across the displayed top-three-per-target edges: redundancy $0.34 \rightarrow 0.15$, unique $0.47 \rightarrow 0.59$, synergy $0.19 \rightarrow 0.26$; Fig.~\ref{fig:dep_vs_nondep}), the clearest signal being the contraction of redundancy. Because the absolute synergy share is sample-size dependent (the full-cohort PHQ-9 synergy share is $\approx 0.08$), these values should be read as a matched-$N$ \emph{contrast} between depression groups rather than as absolute synergy levels; the same qualitative shift holds over all 56 significant edges (Fig.~\ref{fig:dep_vs_nondep_alledges}).

\subsection{PID and ePID applied to the Interpersonal Reactivity Index}
\label{subsec:iri_results}

The PHQ-9 atom decomposition (Fig.~\ref{fig:PID_distributions_acib}) returned a modest synergy share: 6--9\% of pair--target information across the cohorts analysed, with no individual edge synergy-dominant. This raises an interpretive question. Does ePID systematically return low synergy, or do PHQ-9 items share substantial overlapping information about a common depressive-symptom construct by design? To distinguish these possibilities we applied the identical pipeline to the Interpersonal Reactivity Index (IRI; Section~\ref{subsec:data})~\citep{Davis1983_IRI}. Using protocol parameters matched to the multi-dataset benchmark ($n_{\mathrm{sources}} = 5$ ACIB embedding, $I_{\mathrm{mmi}}$ measure), the global 28-item IRI yielded a mean synergy fraction of 23.1\%, with 71\% (95/133) of informative directed pairs synergy-dominant.

The IRI analyses combine three protocols matched to system size. For within-subscale analyses (7 items per subscale, applied to each of the four subscales separately), we apply ACIB-based ePID at $n_{\mathrm{sources}} = 5$ and pool atom fractions across the four subscales for reporting. For cross-subscale triplet analyses (one item from each of three different subscales as $X_1$, $X_2$, target), we report exact two-source PID without embedding, since the triplet system has only two sources and the embedding step is unnecessary. For the cross-subscale triplet analysis, synergy-dominance percentages are computed on the subset of triplets with total mutual information $I(X_i, X_j; X_k) \geq 0.01$ bits, excluding those whose four atoms are individually too small to distinguish synergy and redundancy reliably. For the full 28-item analysis, we report two complementary runs: a primary protocol matched to PHQ-9 ($n_{\mathrm{sources}} = 5$ ACIB-embedded remainder, $I_{\mathrm{mmi}}$ measure) supplies the headline atom fractions reported below, and a full-context run (conditioning on all 26 remaining items) supplies the per-pair atom heatmaps and the source--target retention diagnostic in this section. The two full-instrument runs differ in conditioning-set saturation: at $|Z|=26$ many directed pairs have $I(X_i; X_k \mid \mathbf{Z}) \approx 0$ before any embedding, so the full-context run is reported chiefly as a structural diagnostic rather than as the source of the synergy-dominance claim. Atom fractions across these protocols are comparable in interpretation only when read together as a gradient (Table~\ref{tab:iri_gradient}).

The atoms reported here use the same two-stage pipeline as the PHQ-9 application: ACIB compresses each remainder context into an embedding with $n_{\mathrm{sources}} = 5$, after which a two-source PID is computed using the $I_{\mathrm{mmi}}$ redundancy measure. For the full-context run (conditioning on all 26 remaining items), we restrict the atom-fraction analysis to pairs whose ACIB embedding respected its 5\% conditional-mutual-information loss tolerance, retaining 133 of 756 directed source--target pairs.

\begin{table}[ht]
\centering
\small
\resizebox{\textwidth}{!}{%
\begin{tabular}{lrrrr}
\toprule
\textbf{Configuration} & $n$ & \textbf{Synergy (mean)} & \textbf{Redundancy (mean)} & \textbf{Synergy-dominant pairs} \\
\midrule
IRI within-subscale ($n_{\mathrm{sources}}=5$, pooled)  & 140     & 15.5\% & 29.2\% & 4\%  \\
IRI global 28-item                                      & 133     & 23.1\% & 18.0\% & \textbf{71\%} \\
IRI cross-subscale triplet (raw 2-source PID)           & 1{,}903 & 21.8\% & 22.6\% & 45\% \\
\bottomrule
\end{tabular}%
}
\caption{\apacap{Scale-dependent synergy-dominance gradient across IRI configurations}{Synergy fraction increases from within-subscale IRI through cross-subscale triplet IRI to the global 28-item IRI; redundancy decreases in mirror image. Atom fractions are means over informative directed pairs whose ACIB embedding respected its 5\% conditional-mutual-information loss tolerance (where applicable); synergy-dominant pairs are those for which the synergy share exceeds the redundancy share.}}
\label{tab:iri_gradient}
\end{table}

Atom fractions in Table~\ref{tab:iri_gradient} change systematically with analysis scale. Within a single 7-item subscale (e.g.\ Perspective-Taking analysed in isolation), items are designed to converge on a single empathy facet: redundancy is high (29\%) and synergy is modest (16\%). At the full 28-item scale, synergy reaches 23\% and dominates in 71\% of pairs, while redundancy contracts to 18\%. When the analysis widens to triplets that cross subscales, redundancy drops to 23\% and synergy rises to 22\%. This pattern is consistent with the multi-facet construct theory of empathy~\citep{Davis1983_IRI}: within-facet convergent validity manifests as information-theoretic redundancy, and cross-facet integration manifests as synergy.

Three patterns within the IRI atoms warrant a closer reading. First, among cross-subscale source pairs, the combination of Fantasy and Personal Distress targeting Perspective-Taking items is the most strongly synergy-dominant configuration: 69\% of such triplets cross the synergy--redundancy threshold. Mature cognitive empathy is informed synergistically by the affective extremes: neither Fantasy nor Personal Distress alone is strongly informative, but their joint state is. Second, the Fantasy subscale, rather than Davis's cognitive and affective division, organises redundancy. Every source pair containing Fantasy ranked above every pair without it (24.9\% versus 20.3\% of TMI; difference 4.6 percentage points, 95\% CI [2.1, 6.9], cluster bootstrap over target items), although individual pairs within each group were not separable. Perspective-Taking paired with Empathic Concern was the least redundant combination (17.2\%; lowest in 94\% of replicates). Fantasy items therefore duplicate information the rest of the scale already carries, consistent with long-standing doubts about the subscale's validity~\citep{lawrence2004measuring}; the cognitive and affective core of the instrument contributes the most distinct information. Third, when pooled across all source pairs targeting any Perspective-Taking item, 49\% of pairs are synergy-dominant, the highest subscale-level synergy-dominance rate. Mature cognitive empathy emerges as the IRI facet most synergistically informed by the others, consistent with developmental accounts in which cognitive empathy emerges later than its affective counterpart and comes to integrate it~\citep{decety2021emergence}.

The dominance regime is not an artefact of pipeline choices. The global 28-item mean synergy is stable across $n_{\mathrm{sources}} \in \{3, 4, 5\}$ (range 22.2--24.2\%) and across the twelve embedding methods compared in the multi-dataset benchmark (range 22--28\%), indicating that neither the ACIB cardinality nor the supervised-embedding family drives the result.
\section{Discussion}

Symptom networks have become a widely used framework for representing psychopathology, with edges encoding the (conditional) dependence between symptom pairs. Yet a scalar edge cannot express how several symptoms act together to inform a third, and partial information decomposition (PID), the natural tool for recovering that higher-order structure, becomes intractable and hard to interpret beyond a handful of sources. Embedding-based partial information decomposition (ePID) makes network-context partial information decomposition tractable for psychometric instruments: by compressing the multivariate remainder of a symptom network into a target-directed discrete proxy, it decomposes each directed pair's conditional dependence into redundant, synergistic, and unique channels at realistic item counts. In both PHQ-9 datasets, redundancy and remainder-unique channels jointly dominated pair--target information, with a smaller but consistently non-zero synergistic component, and this compositional profile replicated across two large, demographically distinct datasets (synergy fraction cross-dataset $\rho=0.78$; redundancy $\rho=0.59$), one a population cohort (UK Biobank) and the other an open online sample (the Xinxiang student sample); the redundancy concordance is more modest than the synergy concordance. Applied without parameter changes to the Interpersonal Reactivity Index, the same pipeline instead returned a synergy-dominated decomposition, indicating that ePID resolves structural differences between instruments rather than imposing them. That this added structure sits atop a monotonic pairwise backbone is an assumption check rather than a result in its own right: the close agreement of the Pearson and Spearman partial-correlation networks (Fig.~\ref{fig:PPC_vs_PSP}), together with the rarity of correlation-invisible non-linear dependencies across datasets (Appendix~\ref{app:dependency_comparison}), shows that ePID's contribution is the compositional layer, not a relaxation of pairwise linearity. After situating ePID against existing approaches, we organise the discussion around three questions: what PHQ-9 redundancy dominance implies for depression-instrument design, what the IRI contrast implies about construct heterogeneity, and what the principal methodological limitations and next steps are.

\subsection{What ePID adds, and how it relates to existing approaches}
\label{subsec:triplet_vs_epid}
PPC, SPC, and CMI quantify whether a symptom pair is conditionally associated, but not whether that association reflects information shared with the broader symptom profile or information expressed only when symptoms are considered jointly. PID converts the scalar edge weight into a compositional profile and so represents this distinction directly: two pairs with identical CMI may differ in whether their information is largely redundant with the surrounding context (suggesting substitutability) or partly synergistic (suggesting interaction-dependent structure). The cross-cohort stability of these fractions supports that the decomposition captures reproducible structure rather than cohort-specific noise.

Triplet-level PIDs provide a useful local lens but cannot attribute information to the remainder of the symptom system (all symptoms except the focal source and target), so information redundant with unmodelled symptoms can be misclassified as unique or synergistic within a restricted triplet. The embedding-based approach fills this gap by including a proxy for the remainder as a source, yielding decompositions aligned with the network question: how a focal symptom behaves \emph{in context} of the full symptom set.

Embedding-based PID provides an information-theoretic analogue of ``incremental contribution'' that is not restricted to linear models and does not require enumerating all higher-order interactions. The simulation results indicate that manifold and dimensionality-reduction methods, in particular target-directed ones such as ACIB, preserve synergy structure, whereas Hamming-based clustering substantially distorts it (Fig.~\ref{fig:embedding_overall_performance}).

The stylised symptom triplet (Appendix~\ref{app:stylised_example}) illustrates this principle in miniature: when synergy is high, nonlinear models dramatically outperform linear models, and partial correlations fail to detect the underlying interaction. Although the deterministic structure of the example exaggerates the effect relative to empirical data, the implication carries over: edges with high synergy fractions in the PHQ-9 networks may represent dependence patterns that additive clinical models cannot capture. Testing that possibility would require predictive validation against outcomes, which the present cross-sectional design cannot provide, and would be subject to the estimation caveats set out in Section~\ref{subsec:limitations}.

A central methodological contribution is the validation of ACIB as a practical embedding for network-context PID. The simulation benchmark demonstrates that this supervised design choice is consequential: target-directed embeddings preserve synergy with high fidelity, whereas similarity-based clustering collapses synergy toward zero. This supervised-embedding step therefore provides a general recipe for extending interaction-aware information decompositions to symptom systems of realistic size.

Recent work has argued that common dimensionality-reduction techniques, particularly variance-preserving methods such as PCA, can preferentially retain redundant structure while failing to preserve synergistic higher-order dependencies \citep{varley2025_topology_synergy}. Our embedding-based PID does not treat the symptom system as lying on a low-dimensional manifold and does not optimise a geometric or variance criterion. Instead, we use a target-directed discrete coarse-graining of the remaining symptoms to form a proxy conditioning variable and then quantify unique, redundant, and synergistic contributions in the embedded two-source system. This distinction matters for interpretation: our goal is not to recover a continuous latent geometry of symptoms, but to obtain an interpretable approximation to the distribution of information about a target across a focal symptom and the remainder of the network.

Several complementary frameworks quantify higher-order statistical structure in multivariate systems, and situating ePID relative to these clarifies its specific contribution. Unlike system-level measures such as the O-information (Section~\ref{sec:higher_order}), ePID provides per-edge resolution, decomposing each directed pair's dependence into redundant, synergistic, and unique channels. In neuroscience, PID has been applied to characterise information processing in neural circuits, revealing synergistic coding and redundant representations across neural populations \citep{timme2018synergy_neuroscience}. Symptom networks differ from neural data in important respects: fewer variables, ordinal rather than continuous measurement, and clinical constraints on data collection. ePID addresses the scalability challenge differently from neural applications, by embedding the multivariate remainder rather than restricting analysis to small subsets of variables, making it applicable to the moderate-dimensional, discrete systems typical of psychopathological research.

Among psychometric methods specifically, the closest existing approach to higher-order structure is the Moderated Network Model (MNM) framework \citep{haslbeck2021moderated}, which warrants direct contrast on three axes. First, MNM is parametric: it extends a mixed graphical model with multiplicative interaction terms and reads edge moderation off estimated coefficients, whereas ePID is model-free and information-theoretic, with synergy and redundancy obtained from joint distributions via PID atoms and no functional-form assumption on the interaction. The trade-off is between interpretable coefficients with standard parametric inference on the one hand, and freedom from any assumed functional form for the interaction on the other. Second, MNM requires the analyst to nominate one or a few candidate moderators and enter them explicitly, which suits confirmatory questions about specific moderators; ePID instead treats the entire remainder of the symptom profile simultaneously as conditioning context via the target-directed embedding, which suits exploratory characterisation of system-wide higher-order structure. Third, the two methods identify different objects: MNM tests \emph{which} edges are moderated by a chosen variable and by how much, yielding a moderated-edge map, whereas ePID quantifies \emph{how much} of each edge's information is interaction-dependent (synergy), overlapping with the remainder (redundancy), or carried uniquely: a per-edge compositional profile. The two outputs are not interchangeable: a moderated edge under MNM and a synergistic edge under ePID need not coincide. MNM and ePID therefore occupy complementary positions in the modelling-flexibility and pre-specification trade-off, and the choice between them depends on whether the analyst seeks confirmatory tests of specific moderators or exploratory decomposition of system-level higher-order information.

\subsection{What does PHQ-9 redundancy dominance imply for instrument design?}

In both cohorts the decomposition was dominated by redundancy and remainder-unique channels, and this composition is itself informative about how the instrument is built. It is the information-theoretic signature of a well-targeted, largely unidimensional severity scale whose items function as partial substitutes for one another, which aligns with longstanding critiques of depression sum scores assembled from overlapping, content-similar items~\citep{fried2016good,fried2017moving}. What that redundancy looks like in detail, and where synergy nonetheless appears, is the question we now address, beginning with the dominant channels and working towards the synergistic edges that pairwise measures cannot represent.

\paragraph{Redundancy and remainder-unique channels dominate.}
In both cohorts, redundancy and remainder-unique channels jointly accounted for the vast majority of pair--target information (the Xinxiang student sample: 45\% redundancy, 47\% remainder-unique; UK Biobank: 51\% redundancy, 42\% remainder-unique), with synergy contributing 6--9\%. This finding should be interpreted as a structural characterisation of the symptom network rather than a limitation of the method. High redundancy between a focal symptom and the remainder means that the information the focal symptom carries about the target is also available through other routes in the symptom network, regardless of the underlying generative model. This is a statement about the information topology of the system: many symptoms share overlapping information channels about a given target. The practical value of redundancy maps is that they identify which edges are ``substitutable'' (where the focal symptom's contribution to predicting the target is largely duplicated by other symptoms) and which carry unique or interaction-dependent information that would be lost if the focal symptom were removed from the model. This distinction is not available from PC or CMI alone.

\paragraph{Specific synergistic symptom pairs.}
Synergistic contributions were not uniformly distributed across targets, but the distribution proved reproducible without being attributable to particular symptoms. The edge-level ranking of synergy agreed closely between the two cohorts (Spearman $\rho = 0.78$), and the agreement was insensitive to normalisation ($\rho = 0.76$ with synergy expressed in bits), so the relative distribution of synergy is a reproducible property of the PHQ-9 network rather than of the chosen scale. Per-symptom rankings, by contrast, were not stable. The target ranked highest on synergy fraction differed between cohorts (psychomotor disturbance in the Xinxiang student sample, energy in UK Biobank), and both rankings reordered substantially when synergy was expressed in bits rather than as a fraction of pair--target information (Fig.~\ref{fig:PID_distributions_acib}). We therefore report the reproducibility of the synergy profile and refrain from a symptom-specific interpretation, which the present sample sizes and the sensitivity to normalisation do not support. We record this as a negative result, since naming individual synergistic symptom pairs is the use to which a decomposition of this kind would most naturally be put: ePID recovers the composition of dependence reliably at the level of the whole network, but these data do not identify which PHQ-9 pairs carry the synergy. Localisation of that kind will require either uncertainty intervals on individual atoms, which demand participant-level resampling, or instruments whose synergy is concentrated rather than diffuse, as the Interpersonal Reactivity Index proved to be. In contrast, for many pairs the embedded remainder alone captured most of the pairwise information, indicating that these targets are largely determined by the broader symptom context rather than by any single additional source.

\paragraph{Clinical reading.}
The introduction states that ``if two source symptoms are highly redundant, changing the state of only one source would change little to the target's state.'' The empirical results allow us to revisit this observation concretely. Redundancy-dominated edges suggest that the focal symptom's predictive contribution to the target is largely duplicated by other symptoms; removing it from a predictive model would change little, because the same information is available through other routes. Synergistic edges suggest a different pattern: the information about the target would be lost if either the focal symptom or the remainder configuration were unobserved, because it arises only from their joint state. These are hypothesis-generating insights rather than treatment recommendations, given the cross-sectional design and the observational nature of the data. Nonetheless, they illustrate how the redundancy--synergy distinction can generate qualitatively different clinical hypotheses from the same conditional-dependence backbone.

Whether such hypotheses generalise depends on whether the compositional profile itself is reproducible across samples, an empirical question that connects the within-cohort findings above to the broader symptom-network replicability literature.

The cross-cohort stability of PID atom fractions speaks to the broader replicability debate in symptom network research. Previous work has raised concerns about the replicability of network edge weights and centrality indices across samples \citep{forbes2017mdd_replicability,fried2017moving}. The \emph{composition} of dependence, and not only its magnitude, replicates across two large and demographically distinct datasets. It does not, however, replicate better than the pairwise quantities it decomposes. Over the same 28 symptom pairs, cross-cohort rank agreement was $\rho = 0.82$ for CMI and $\rho = 0.74$ for both partial-correlation variants, against $\rho = 0.76$, $0.59$ and $0.59$ for the synergy, redundancy and remainder-unique fractions respectively. The compositional layer should therefore be read as reproducible to a degree comparable with standard edge weights, rather than as a more stable representation of network structure.

\subsection{What does the IRI contrast imply about construct heterogeneity?}

\paragraph{Synergy dominance and the construct-structure hypothesis.}
The cross-instrument application to the Interpersonal Reactivity Index (Section~\ref{subsec:iri_results}) demonstrates that the redundancy dominance observed in PHQ-9 is not an artefact of the embedding pipeline. The same ACIB-based ePID, applied without parameter changes, returned a synergy-dominated decomposition on the IRI: 71\% of informative directed pairs had higher synergy than redundancy, against 0\% for PHQ-9. The monotonic gradient across analysis scales (within-subscale 16\% synergy; cross-subscale triplets 22\%; full instrument 23\%) indicates that the dominance regime is shaped by the construct structure of the instrument rather than by the embedding choice. The same ePID pipeline that returned a redundancy-dominated PHQ-9 recovered a synergy-dominated IRI without parameter changes, indicating that the dominance regime is a property of the instrument rather than of the method. Together with the PHQ-9 anchor, these findings indicate that ePID discriminates between redundancy-dominated and synergy-dominated psychometric instruments using identical machinery: the decomposition captures structural differences in how information is distributed across items, rather than artefacts of the instrument or the embedding choice.

\paragraph{Specific synergy patterns.}
Three patterns within the IRI atoms warrant a closer reading. First, in the cross-subscale raw 2-source PID (no remainder embedding), the combination of Fantasy and Personal Distress targeting Perspective-Taking items is the most strongly synergy-dominant configuration: 69\% of such triplets cross the synergy--redundancy threshold. This suggests that mature cognitive empathy is informed synergistically by affective extremes: neither Fantasy nor Personal Distress alone is strongly informative, but their joint state is. Second, also within the cross-subscale 2-source PID, redundancy is organised by the Fantasy subscale rather than by Davis's cognitive and affective division: pairs containing Fantasy carried reliably more redundant information than pairs without it, while pairs within each group were not separable, and Perspective-Taking with Empathic Concern was the least redundant combination. Fantasy items therefore duplicate information the rest of the scale already carries, consistent with long-standing doubts about the subscale's validity~\citep{lawrence2004measuring}. Third, in the full 28-item ePID ($n_{\mathrm{sources}}=5$ ACIB embedding, $I_{\mathrm{mmi}}$), 49\% of source pairs targeting any Perspective-Taking item are synergy-dominant, the highest subscale-level synergy-dominance rate, consistent with developmental accounts in which cognitive empathy emerges later than its affective counterpart and comes to integrate it~\citep{decety2021emergence}.

\paragraph{Theoretical implications and limitations.}
The redundancy--synergy distinction therefore captures structural differences between psychometric instruments, not just between symptoms within an instrument. Instruments designed around a single underlying construct (PHQ-9 depressive symptoms) yield high redundancy because items are partial substitutes for one another, whereas multi-facet instruments (IRI's four empathy facets) yield high synergy because diagnostic information about a target item often resides in the joint configuration of items from different facets. This suggests an information-theoretic axis for psychometric construct validation that complements conventional reliability and factor-analytic criteria. Because self-report items are imperfectly reliable, and measurement noise attenuates higher-order structure more than linear structure, by analogy with the reliability attenuation of degree-$k$ interactions~\citep{schulz2026measurement} the reported synergy fractions are best read as a plausible lower bound. Two caveats qualify the IRI-specific findings. The IRI sample is drawn from the Open-Source Psychometrics Project (a self-selected online sample whose composition differs from clinical or community samples), and, in the full-context analysis (conditioning on all 26 remaining items), the atom-fraction analysis is restricted, by design, to the 133 of 756 directed pairs (17.6\%) for which the ACIB embedding preserved at least 95\% of the source--target conditional mutual information. The 5\% loss tolerance is applied because the atoms of a PID are unstable summaries when the embedding has thrown away an appreciable share of the quantity being decomposed; the remaining pairs therefore represent the IRI source--target combinations on which the decomposition can be read with confidence. The IRI patterns should therefore be read as a within-instrument illustration of discriminative validity rather than a population-level inference about empathy structure.

\subsection{Methodological limitations and next steps}
\label{subsec:limitations}

The choice of PID measure is itself a non-trivial modelling choice, since PID has no single canonical measure; we therefore report $I_{\mathrm{mmi}}$ as primary with $I_{\min}$ as a robustness check, and the qualitative patterns (edge rankings by synergy fraction, cross-cohort stability) held across both. The multi-dataset benchmark across 83 real-world datasets and five PID measures provides a comprehensive assessment of embedding fidelity for this class of methods. Three findings deserve emphasis.

First, ACIB's dominance is robust across four of five measures ($I_{\mathrm{mmi}}$, $I_{\min}$, $I_{\mathrm{rr}}$, $I_{\wedge}$) and across diverse data structures, from small clinical samples to large population surveys. At $N=5$, ACIB won 84.4\% of per-dataset contests for $I_{\wedge}$ [95\% CI: 78.4, 91.1], 75--78\% for $I_{\min}$/$I_{\mathrm{mmi}}$, and 64.9\% for $I_{\mathrm{rr}}$ [56.2, 73.3] (its most modest margin), with the lowest error growth slope among all methods (0.041 [0.035, 0.046] bits per additional source). This consistency suggests that target-directed coarse-graining is a generally effective strategy for preserving the information structure that PID quantifies, not an artefact of the synthetic benchmark's calibration to PHQ-9.

Second, the striking failure of ACIB on the $I_{\pm}$ (pointwise partial information) measure, where ACIB won only 2.6\% of contests, has a specific mechanistic origin. Unlike non-negative measures, $I_{\pm}$ produces signed atoms: 77.7\% of ground-truth unique-source values were negative at $N=5$ (compared to 0.7\% for $I_{\min}$). The embedding compression disrupts the structured cancellation between positive and negative contributions, inflating absolute error beyond the total signal (RE $> 100$\%). This finding has practical implications: researchers who wish to use pointwise PID measures should consider alternative embeddings such as ReliefF or SVD+$k$-means, though errors remain substantial.

Third, at $N=3$ all methods were statistically indistinguishable (Appendix~\ref{subsec:results_multidataset}), consistent with the minimal compression required when only two sources are embedded into one variable. Method choice becomes consequential only at $N \geq 4$, where ACIB's information-bottleneck objective provides a clear advantage.

More broadly, the finding that no single embedding method dominates all PID measures cautions against treating any embedding as a universal approximation. The choice of embedding should be aligned with the choice of PID measure, and sensitivity analyses across both dimensions are advisable.

\paragraph{Limitations.}
Three limitations qualify the interpretation of these findings. First, embedding fidelity has been validated for $N=3$ to $N=5$ sources, whereas the empirical setting embeds $N=6$ remainder symptoms. Computing a ground-truth PID at $N=6$ is infeasible (the redundancy lattice contains approximately $7.8 \times 10^{6}$ antichains). The sub-linear error growth from $N=3$ to $N=5$ and the bounded atom magnitudes ($\leq H(X_k)$) support but do not guarantee extrapolation to $N=6$; accordingly, the empirical atom fractions should be interpreted as approximate compositional profiles rather than exact quantities.

Second, sampling uncertainty in the atom estimates themselves is not formally quantified. We report point estimates from plug-in (frequency-based) estimators, for which the entropy term is negatively biased in finite samples while the mutual-information functionals built from it are typically biased upward~\citep{paninski2003entropy}; participant-level bootstrap confidence intervals for PID atoms would be informative but are computationally demanding because each resample requires re-fitting the ACIB embedding for all source--target pairs. The confidence intervals we do report, for contrasts between groups of source--target pairs, are obtained by resampling analysis units (target items or directed edges) rather than participants, and therefore describe variability across pairs within a cohort, not estimation error in any individual atom. They should not be read as sampling intervals on the atom values. Bias is partially mitigated by the large sample sizes ($N > 24{,}000$ for the Xinxiang student sample; $N > 154{,}000$ for UK Biobank) and the low cardinality (three levels per item), but no formal bias correction was applied. Small differences in atom fractions across edges or cohorts should therefore be interpreted cautiously.

Third, three design choices in the empirical pipeline systematically affect atom estimates. We merged the two highest PHQ-9 response categories to reduce sparsity, which may underestimate synergy by collapsing fine-grained joint configurations that distinguish interaction-dependent patterns from additive ones. Separately, the ACIB embedding is optimised to preserve information about the target, which by design ensures the remainder embedding is maximally informative about $X_k$ and could inflate the remainder-unique component relative to a target-agnostic embedding. The decomposition should therefore be interpreted as conditional on these design choices: estimated synergy fractions are likely conservative, and the atoms describe how information about $X_k$ is distributed between the focal source and a \emph{target-optimised} summary of the remainder. A third design choice concerns the 5\% relative-CMI tolerance used by ACIB, which serves both as an internal stopping rule and as a downstream retention filter in the full-context IRI analysis. The 17.6\% retention rate (133 of 756 pairs) reflects principally that conditioning on 26 of the 28 IRI items saturates the conditioning set: most directed pairs have $I(X_i; X_k \mid \mathbf{Z})$ at or below numerical zero before any embedding step, and the filter therefore drops mathematically vacuous pairs rather than badly compressed ones. The PHQ-9 application restricts the conditioning set to the five highest-mutual-information sources and so does not encounter this saturation regime (45/56 directed pairs retained); the divergent retention rates between PHQ-9 and IRI reflect the protocol's interaction with instrument size, not a property of the embedding family. The 5\% threshold is supported by the sensitivity analysis in Appendix~\ref{app:acib_sensitivity}, where the atom estimates are insensitive to the tolerance and to the smoothing prior across the tested grid.

A further interpretive limitation concerns sign. Because the measures used in the main analyses ($I_{\mathrm{mmi}}$, $I_{\min}$, $I_{\mathrm{rr}}$, $I_{\wedge}$) are non-negative, they carry no sign: unlike partial correlation, ePID has no analogue of a negative or inhibitory association and cannot, on its own, distinguish facilitative from suppressive relations. The signed $I_{\pm}$ measure is the only exception, and for the reasons above it is not used in the main analyses.

\paragraph{Directedness does not imply causality.}
Directed edges in the ePID framework are a computational convention required by PID's source--target formulation, not causal claims. The decomposition is not symmetric: $\mathrm{PID}(X_i \to X_k)$ and $\mathrm{PID}(X_k \to X_i)$ generally yield different atom profiles because the remainder embedding is constructed separately for each target. This asymmetry reflects the information structure of the system (how predictive each symptom is of a given target in the context of the remaining symptoms) rather than causal direction. Cross-sectional observational data cannot distinguish causal influence from confounding, reverse causation, or shared latent causes. The clinical hypotheses described in the introduction (e.g., that insomnia can induce fatigue) motivate the choice of symptom instrument and provide domain context, but they are not tested or validated by the present analyses.

\paragraph{Next steps.}
\label{subsec:future}
Two extensions would most directly strengthen the ePID framework.
First, bootstrap or permutation-based confidence intervals for PID atoms would enable formal statistical inference on atom differences across edges, targets, and cohorts, addressing the estimation uncertainty noted in Section~\ref{subsec:limitations}.
Second, a systematic sensitivity analysis of category merging (varying the number of ordinal levels from 2 to 4) would quantify how discretisation affects atom estimates, particularly synergy.

\subsection*{Conclusion}
ePID makes network-context partial information decomposition tractable for psychometric instruments by compressing the multivariate remainder of a symptom network into a target-directed discrete proxy, validated up to $N=5$ remainder sources. A comprehensive benchmark, spanning calibrated synthetic data and 83 real-world datasets and five PID measures, establishes that ACIB preserves the information structure PID quantifies while identifying measure-specific failure modes, notably for the signed $I_{\pm}$. Applied to two PHQ-9 datasets and to the Interpersonal Reactivity Index, the method shows that decomposing conditional dependence into redundant and synergistic channels adds interpretive value beyond scalar association measures: the composition of dependence replicates across datasets within an instrument, the same pipeline discriminates redundancy-dominated from synergy-dominated instruments, and the redundancy/synergy distinction generates qualitatively different clinical hypotheses about symptom substitutability and interaction dependence. Together, these results support embedding-based PID as a practical complement to standard symptom-network methodology.

\section*{Ethics}
This research has been conducted using the UK Biobank Resource under Application Number 68746; UK Biobank holds generic ethical approval from the North West Multi-centre Research Ethics Committee, and its participants provided written informed consent. All other datasets, namely the Xinxiang student sample (openly available; Su et al. 2024), the Interpersonal Reactivity Index from the Open-Source Psychometrics Project, and the 83 datasets in the multi-dataset benchmark (Supplemental Table~\ref{tab:benchmark_datasets}), are publicly available and de-identified.

\section*{Data and Code Availability}
UK Biobank data are available through the UK Biobank Access Management System (\url{https://www.ukbiobank.ac.uk}). The Xinxiang student sample is openly available on Zenodo (\url{https://zenodo.org/records/10423537}; DOI 10.5281/zenodo.10423537). The Interpersonal Reactivity Index is openly available from the Open-Source Psychometrics Project (\url{https://openpsychometrics.org}). Sources for the 83 multi-dataset benchmark datasets are listed in Supplemental Table~\ref{tab:benchmark_datasets}. Code implementing the ePID pipeline will be made available upon publication at \url{https://github.com/CillianHourican/ePID}.

\newpage
\FloatBarrier
\appendix
\renewcommand{\thefigure}{S\arabic{figure}}
\renewcommand{\thetable}{S\arabic{table}}
\setcounter{figure}{0}
\setcounter{table}{0}

This appendix comprises the following sections:
Appendix~\ref{app:multi_dataset} describes the multi-dataset benchmark design and participating datasets.
Appendix~\ref{app:embedding_methods} provides detailed descriptions and hyperparameter settings for all 13 embedding methods.
Appendix~\ref{app:embedding_scaling} discusses the information-bottleneck interpretation of the ACIB embedding and computational scaling properties.
Appendix~\ref{app:bn_calibration} presents calibration statistics for the synthetic Bayesian Network ensemble.
Appendix~\ref{app:benchmark_stats} reports omnibus and post-hoc statistical tests for the embedding benchmark.
Appendix~\ref{app:additional_diagnostics} reports additional embedding-fidelity diagnostics and symptom-level PID examples.
Appendix~\ref{app:dependency_comparison} examines the prevalence of non-linear dependencies in clinical datasets.
Appendix~\ref{app:cmi_pid_comparison} provides cross-cohort comparisons of information-theoretic PHQ-9 networks.
Appendix~\ref{app:stylised_example} presents an illustrative three-variable example contrasting how partial correlation, conditional mutual information, and PID reveal progressively richer dependence structure.
Appendix~\ref{app:acib_sensitivity} demonstrates that the reported decompositions are robust to the ACIB embedding hyperparameters and justifies embedding only a bounded remainder context.

\section{Multi-Dataset Benchmark}
\label{app:multi_dataset}

This section describes the assembly and preprocessing of the 83 real-world datasets used to evaluate embedding fidelity across diverse data structures and PID measures (Section~\ref{subsec:multi_dataset_benchmark}).


\subsection{Benchmark dataset assembly and preprocessing}
\label{app:datasets}

We assembled 83 datasets from publicly available repositories spanning clinical, epidemiological, and psychometric domains, including NHANES~\citep{CDC_NHANES_2017}, SHARE~\citep{BorschSupan2022_SHARE_W7}, ELSA~\citep{Banks2023_ELSA}, HRS~\citep{HRS_Harmonized}, PROMIS~\citep{Cella2010_PROMIS}, OpenPsychometrics, the UCI Machine Learning Repository, OSF, BRFSS~\citep{CDC_BRFSS_2022}, MEPS~\citep{AHRQ_MEPS_2022}, the Household Pulse Survey~\citep{Census_HPS_2024}, ICPSR, Zenodo, Figshare, and Mendeley Data.
Datasets span depression, anxiety, PTSD, stress, personality, eating disorders, physical functioning, chronic conditions, sleep, and other clinical domains, with 5--56 variables per dataset and sample sizes ranging from 32 to 445{,}000 observations.
For each dataset, we selected only substantive item-level or symptom-level variables, excluding identifiers, demographic covariates, composite scores, and diagnostic labels.
Source-specific missing-value codes (e.g., 7/77/777 and 9/99/999 in NHANES; negative codes in SHARE) were recoded to missing.
For NHANES SAS transport files, the XPT format encodes zero as a tiny floating-point number (${\approx}5.4 \times 10^{-79}$); these values were rounded to zero before further processing.

Columns with more than 30\% missing values were dropped, followed by rows with more than 50\% missing values across the remaining columns.
Remaining missing values were imputed using $k$-nearest neighbours ($k = 5$) via scikit-learn's \texttt{KNNImputer}.
For four datasets where this filtering left fewer than three columns or thirty rows (\texttt{nhanes\_phys\_func}, \texttt{covidistress\_28}, \texttt{ncsr\_depression\_18}, \texttt{nhanes\_alcohol}), listwise deletion (complete-case analysis) was used instead, retaining all variables and only rows with no missing values.
After imputation, any genuinely constant columns (zero variance) were removed.

For each dataset, two association matrices were estimated.
Partial correlations were computed from an $\ell_1$-regularised precision matrix obtained via \texttt{GraphicalLassoCV} with 5-fold cross-validation, which estimates partial correlations as scaled off-diagonal entries of the regularised precision matrix $\Sigma^{-1}$, yielding a sparse estimate of conditional dependencies.
For mutual information estimation, all imputed values were first rounded to the nearest integer to recover ordinal response levels.
Variables with five or fewer unique integer values were retained at their original levels; variables with more than five unique values were discretised into eight equal-frequency bins.
For partial information decomposition, all variables were further reduced to at most $k$ discrete states by iteratively merging adjacent categories with the smallest combined frequency, where $k$ was the largest integer satisfying $k^{(N_{\mathrm{sources}}+1)} \leq N/5$, ensuring adequate coverage of the joint state space.
Pairwise mutual information was estimated using the discrete estimator from the \texttt{npeet} library, with diagonal entries set to the Shannon entropy of each variable.

Table~\ref{tab:benchmark_datasets} lists all 83 datasets, their domain, number of selected variables, sample size, source repository, and citation.

{%
\footnotesize
\setlength{\LTcapwidth}{\textwidth}
\begin{longtable}{lllrll}
\caption{\apacap{Benchmark datasets}{The 83 datasets used in the multi-dataset embedding benchmark. \emph{Items} indicates the number of substantive variables selected before preprocessing; \emph{N} is the raw sample size before row-level missingness filtering. easySHARE waves (6a--6g) are drawn from the same longitudinal study but are treated as separate networks because each wave has an independent cross-section of respondents.}}
\label{tab:benchmark_datasets} \\
\toprule
Dataset & Domain & Items & \multicolumn{1}{c}{$N$} & Source & Citation \\
\midrule
\endfirsthead
\toprule
Dataset & Domain & Items & \multicolumn{1}{c}{$N$} & Source & Citation \\
\midrule
\endhead
\midrule
\multicolumn{6}{r}{\emph{Continued on next page}} \\
\bottomrule
\endfoot
\bottomrule
\endlastfoot
CSWS Items             & Self-worth         & 35  & 680       & OSF        & \citep{Briganti2019_CSWS} \\
CSWS Subscales         & Self-worth         & 7   & 680       & OSF        & \citep{Briganti2019_CSWS} \\
SHARE W7               & Depression         & 17  & 77{,}000  & SHARE      & \citep{BorschSupan2022_SHARE_W7} \\
ELSA W6                & Depression          & 17  & 10{,}000  & UK Data    & \citep{Banks2023_ELSA} \\
HRS W13                & Sleep/pain/health  & 18  & 21{,}000  & HRS        & \citep{HRS_Harmonized} \\
easySHARE 2004 (W1)    & Summary health     & 8   & 30{,}000  & SHARE      & \citep{BorschSupan2022_easySHARE, Gruber2014_easySHARE} \\
easySHARE 2007 (W2)    & Summary health     & 9   & 37{,}000  & SHARE      & \citep{BorschSupan2022_easySHARE, Gruber2014_easySHARE} \\
easySHARE 2011 (W4)    & Summary health     & 8   & 58{,}000  & SHARE      & \citep{BorschSupan2022_easySHARE, Gruber2014_easySHARE} \\
easySHARE 2013 (W5)    & Summary health     & 8   & 66{,}000  & SHARE      & \citep{BorschSupan2022_easySHARE, Gruber2014_easySHARE} \\
easySHARE 2015 (W6)    & Summary health     & 9   & 68{,}000  & SHARE      & \citep{BorschSupan2022_easySHARE, Gruber2014_easySHARE} \\
easySHARE 2017 (W7)    & Summary health     & 6   & 77{,}000  & SHARE      & \citep{BorschSupan2022_easySHARE, Gruber2014_easySHARE} \\
easySHARE 2020 (W8)    & Summary health     & 9   & 47{,}000  & SHARE      & \citep{BorschSupan2022_easySHARE, Gruber2014_easySHARE} \\
NHANES PHQ-9           & Depression         & 9   & 5{,}500   & CDC        & \citep{CDC_NHANES_2017, kroenke2001phq9} \\
NHANES Sleep           & Sleep              & 10  & 6{,}200   & CDC        & \citep{CDC_NHANES_2017} \\
NHANES Phys.\ Func.    & Disability         & 24  & 8{,}400   & CDC        & \citep{CDC_NHANES_2017} \\
NHANES Health Status   & General health     & 8   & 8{,}400   & CDC        & \citep{CDC_NHANES_2017} \\
NHANES Alcohol         & Alcohol            & 9   & 5{,}500   & CDC        & \citep{CDC_NHANES_2017} \\
NHANES Blood Pressure  & Cardiovascular     & 10  & 6{,}200   & CDC        & \citep{CDC_NHANES_2017} \\
NHANES Med.\ Cond.     & Chronic conditions & 17  & 8{,}900   & CDC        & \citep{CDC_NHANES_2017} \\
NHANES Smoking         & Smoking            & 10  & 6{,}700   & CDC        & \citep{CDC_NHANES_2017} \\
NHANES Phys.\ Activity & Activity           & 16  & 5{,}900   & CDC        & \citep{CDC_NHANES_2017} \\
NHANES Hospital Util.  & Healthcare         & 9   & 9{,}300   & CDC        & \citep{CDC_NHANES_2017} \\
DASS-42                & Dep/Anx/Stress     & 42  & 39{,}800  & OpenPsych  & \citep{Lovibond1995_DASS} \\
TMAS                   & Trait anxiety      & 50  & 5{,}400   & OpenPsych  & \citep{Taylor1953_TMAS} \\
Big Five (IPIP)        & Personality        & 50  & 19{,}700  & OpenPsych  & \citep{Goldberg1999_BigFive} \\
RSE                    & Self-esteem        & 10  & 48{,}000  & OpenPsych  & \citep{Rosenberg1965_RSE} \\
ECR                    & Attachment         & 36  & 51{,}500  & OpenPsych  & \citep{Brennan1998_ECR} \\
HSNS+DD                & Narcissism/dep     & 10  & 54{,}000  & OpenPsych  & \citep{Hendin1997_HSNS, Jonason2010_DD} \\
KIMS                   & Mindfulness        & 39  & 601       & OpenPsych  & \citep{Baer2004_KIMS} \\
SCS                    & Sexual compulsiv.  & 12  & 3{,}400   & OpenPsych  & \citep{Kalichman1995_SCS} \\
Dermatology            & Skin disease       & 34  & 366       & UCI        & \citep{UCI_Dermatology, Guvenir1998_Dermatology} \\
Primary Tumor          & Cancer features    & 17  & 339       & UCI        & \citep{UCI_PrimaryTumor} \\
Early-Stage Diabetes   & Diabetes symptoms  & 16  & 520       & UCI        & \citep{UCI_EarlyDiabetes, Islam2019_EarlyDiabetes} \\
Thyroid Cancer Recur.  & Cancer clinical    & 16  & 383       & UCI        & \citep{UCI_ThyroidCancer} \\
Autism Screen.\ Adult  & ASD screening      & 21  & 704       & UCI        & \citep{UCI_AutismAdult, Thabtah2018_AQ10} \\
Thoracic Surgery       & Surgical symptoms  & 16  & 470       & UCI        & \citep{UCI_ThoracicSurgery, Zieba2014_ThoracicSurgery} \\
Mammographic Mass      & Breast imaging     & 5   & 961       & UCI        & \citep{UCI_MammographicMass, Elter2007_MammographicMass} \\
Chronic Kidney Disease & Kidney signs/labs  & 24  & 400       & UCI        & \citep{UCI_CKD} \\
MI Complications       & Cardiac symptoms   & 111 & 1{,}700   & UCI        & \citep{UCI_MI_Complications, Golovenkin2020_MI} \\
Glioma Grading         & Cancer mutations   & 25  & 839       & UCI        & \citep{UCI_GliomaGrading} \\
Cervical Cancer Risk   & STD/risk factors   & 36  & 858       & UCI        & \citep{UCI_CervicalCancerRisk} \\
CDC Diabetes Indic.    & Health indicators  & 21  & 253{,}700 & UCI/CDC    & \citep{UCI_CDC_Diabetes} \\
AIDS Clinical Trials   & HIV clinical       & 25  & 2{,}139   & UCI        & \citep{UCI_ACTG175, Hammer1996_ACTG175} \\
Mesothelioma           & Cancer symptoms    & 34  & 324       & UCI        & \citep{UCI_Mesothelioma} \\
NPHA Healthy Aging     & Health self-report & 15  & 714       & UCI        & \citep{UCI_NPHA} \\
HCV Egyptian Patients  & Liver symptoms     & 28  & 1{,}385   & UCI        & \citep{UCI_HCV_Egypt} \\
Xinxiang PHQ-9         & Depression         & 9   & 24{,}292  & Zenodo     & \citep{Su2024_TemporalDynamics, Su2024_TemporalDynamics_data} \\
Xinxiang GAD-7         & Anxiety            & 7   & 24{,}292  & Zenodo     & \citep{Su2024_TemporalDynamics, Su2024_TemporalDynamics_data} \\
Xinxiang ISI           & Insomnia           & 7   & 24{,}292  & Zenodo     & \citep{Su2024_TemporalDynamics, Su2024_TemporalDynamics_data} \\
Xinxiang PSS           & Stress             & 10  & 24{,}292  & Zenodo     & \citep{Su2024_TemporalDynamics, Su2024_TemporalDynamics_data} \\
PHQ+GAD+ESS Mexico     & Dep/Anx/Sleep      & 33  & 783       & Figshare   & \citep{Figshare_PHQ_GAD_ESS_Mexico} \\
MHP Students           & Anx/Stress/Dep     & 26  & 2{,}000   & Figshare   & \citep{Syeed2024_MHP_data} \\
Bangladesh MH          & Multi-scale dep    & 45  & 502       & Mendeley   & \citep{Mendeley_BangladeshMH} \\
PHQ-9 Students         & Depression         & 9   & 682       & Mendeley   & \citep{Mendeley_PHQ9_Students} \\
Colombia Well-Being    & Stress/Anx/Dep     & 26  & 3{,}000   & Mendeley   & \citep{Martinez2024_ColombiaWellbeing, Martinez2024_ColombiaWellbeing_data} \\
McNally 2014           & PTSD               & 17  & 344       & OSF        & \citep{McNally2015_PTSD} \\
ED Network             & Eating disorders   & 22  & 245       & OSF        & \citep{Vervaet2021_ED} \\
IRI Empathy            & Empathy            & 28  & 1{,}973   & OSF        & \citep{Davis1983_IRI} \\
CPS Chinese            & Purpose in life    & 12  & 598       & OSF        & \citep{Wu2024_CPS_Chinese, Wu2024_CPS_Chinese_data} \\
COVIDiSTRESS           & Stress/Loneliness  & 28  & 173{,}000 & OSF        & \citep{Yamada2021_COVIDiSTRESS} \\
PROMIS Anxiety         & Anxiety            & 56  & 817       & Harvard DV & \citep{Cella2010_PROMIS} \\
PROMIS Depression      & Depression         & 56  & 811       & Harvard DV & \citep{Cella2010_PROMIS} \\
PROMIS Anger           & Anger              & 56  & 918       & Harvard DV & \citep{Cella2010_PROMIS} \\
PROMIS Fatigue (Exp)   & Fatigue            & 56  & 821       & Harvard DV & \citep{Cella2010_PROMIS} \\
PROMIS Fatigue (Imp)   & Fatigue            & 56  & 820       & Harvard DV & \citep{Cella2010_PROMIS} \\
PROMIS Pain Interf.    & Pain               & 56  & 866       & Harvard DV & \citep{Cella2010_PROMIS} \\
PROMIS Pain Quality    & Pain               & 56  & 862       & Harvard DV & \citep{Cella2010_PROMIS} \\
PROMIS Pain Behavior   & Pain               & 56  & 859       & Harvard DV & \citep{Cella2010_PROMIS} \\
PROMIS PhysFun A       & Physical function  & 56  & 812       & Harvard DV & \citep{Cella2010_PROMIS} \\
PROMIS PhysFun B       & Physical function  & 56  & 814       & Harvard DV & \citep{Cella2010_PROMIS} \\
PROMIS Social Perf     & Social function    & 56  & 864       & Harvard DV & \citep{Cella2010_PROMIS} \\
PROMIS Social Sat      & Social function    & 56  & 851       & Harvard DV & \citep{Cella2010_PROMIS} \\
PROMIS Alcohol         & Alcohol            & 56  & 903       & Harvard DV & \citep{Cella2010_PROMIS} \\
PROMIS Profile 29      & Multi-domain       & 29  & 4{,}500   & Harvard DV & \citep{Hays2018_PROMIS29} \\
BRFSS 2022             & Health indicators  & 25  & 445{,}000 & CDC        & \citep{CDC_BRFSS_2022} \\
PHQ+GAD+ISI+PSS comb.  & All 4 scales       & 33  & 24{,}292  & Zenodo     & \citep{Su2024_TemporalDynamics, Su2024_TemporalDynamics_data} \\
NCS-R Depression       & MDE symptoms       & 18  & 2{,}300   & ICPSR      & \citep{Kessler2004_NCSR, ICPSR_CPES} \\
MIDUS 2 Phys.\ Symp.   & Somatic symptoms   & 10  & 4{,}000   & ICPSR      & \citep{ICPSR_MIDUS2} \\
MIDUS 2 Chronic Cond.  & Multimorbidity     & 20  & 4{,}000   & ICPSR      & \citep{ICPSR_MIDUS2} \\
MIDUS 2 Health Limits  & Functional limits  & 10  & 4{,}000   & ICPSR      & \citep{ICPSR_MIDUS2} \\
MEPS 2022 SF-12        & Health status      & 17  & 11{,}300  & AHRQ       & \citep{AHRQ_MEPS_2022} \\
MEPS 2022 Conditions   & Chronic conditions & 17  & 22{,}300  & AHRQ       & \citep{AHRQ_MEPS_2022} \\
HPS Dec 2024           & PHQ-2+GAD-2+Disab. & 14  & 9{,}400   & Census     & \citep{Census_HPS_2024} \\
\end{longtable}
}%

Per-dataset variable selections, exclusion criteria, and recoding rules are available in the project's data documentation repository.

\section{Embedding Methods: Descriptions and Hyperparameters}
\label{app:embedding_methods}

To compress the multivariate remainder $\mathbf{Z}_{i\to k} = \mathbf{X} \setminus \{X_i, X_k\}$ into a single discrete embedding variable $E_{i\to k}$ for two-source PID computation, we benchmarked 13 embedding methods spanning five families.
Table~\ref{tab:embedding_hyperparams} summarises the method families, software implementations, and key hyperparameters used throughout the benchmark.
All methods were applied with a global random seed of 42 for reproducibility.
Below, we describe each method and its hyperparameter settings.

\begin{table}[htbp]
\centering
\caption{Summary of the 13 embedding methods, grouped by family. All methods share a global random seed of 42. ``$K$'' denotes the number of clusters or output cardinality; ``$k$'' denotes the number of selected features; ``$B$'' denotes the number of quantile bins per dimension.}
\label{tab:embedding_hyperparams}
\begin{tabular}{llll}
\toprule
Method & Family & Package/Library & Key Hyperparameters \\
\midrule
ACIB & Info-theoretic & Custom (npeet) & $K_{\max}=12$, $K_{\min}=2$, $\alpha=0.5$, \\
     &                &                & loss tol.\ $=5\%$, max\_iter $=40$ \\
Greedy CMI & Info-theoretic & Custom (npeet) & max\_vars $=4$, loss tol.\ $=5\%$ \\
\midrule
JMI & MI-based selection & scikit-feature & $k=3$ \\
CMIM & MI-based selection & scikit-feature & $k=3$ \\
ReliefF & MI-based selection & skrebate & $k=3$, $n\_\text{neighbours}=100$ \\
\midrule
$k$-modes & Clustering & kmodes & $K=6$, init $=$ Huang, $n\_\text{init}=5$ \\
Spectral Hamming & Clustering & scikit-learn & $K=6$, $\gamma=5.0$, labels $=$ $k$-means \\
Agglom.\ Hamming & Clustering & scikit-learn & $K=6$, linkage $=$ average \\
\midrule
MCA $+$ $k$-means & Dim.\ red.\ $+$ clust. & prince, scikit-learn & $n\_\text{comp}=2$, $K=6$, $n\_\text{init}=10$ \\
SVD $+$ $k$-means & Dim.\ red.\ $+$ clust. & scikit-learn & $n\_\text{comp}=2$, $K=6$, $n\_\text{init}=10$ \\
NMF $+$ $k$-means & Dim.\ red.\ $+$ clust. & scikit-learn & $n\_\text{comp}=2$, $K=6$, init $=$ nndsvda, \\
                   &                        &               & $n\_\text{init}=10$ \\
\midrule
SIR $+$ binning & Supervised SDR & sliced & $n\_\text{dir}=2$, $B=4$ \\
PLS $+$ binning & Supervised SDR & scikit-learn & $n\_\text{comp}=2$, $B=4$ \\
\bottomrule
\end{tabular}
\end{table}

\subsection{Information-theoretic methods}
\label{app:embed_infotheo}

\paragraph{Agglomerative Conditional Information Bottleneck (ACIB).}
ACIB is a two-phase, deterministic, target-directed compression algorithm that maps the joint states of the multivariate remainder $\mathbf{Z}_{i\to k}$ into a low-cardinality discrete variable $E_{i\to k}$ while preserving as much information as possible about the target $X_k$.
In the first phase, a $k$-CIB (Conditional Information Bottleneck) initialisation is performed.
The unique joint states of $\mathbf{Z}_{i\to k}$ are assigned to $K_{\max} = 12$ initial clusters using a farthest-first seeding procedure based on the Jensen--Shannon (JS) divergence of the conditional target distributions $P(X_k \mid \mathbf{Z}_{i\to k} = z, X_i = s)$ across source states $s$.
Cluster assignments are then refined by iteratively reassigning each joint state to the cluster whose prototype minimises the Kullback--Leibler divergence, with conditional distributions smoothed via Laplace smoothing ($\alpha = 0.5$) to regularise sparse cells.
This alternating assignment--update loop runs for up to 40 iterations or until assignments stabilise.

In the second phase, a greedy agglomerative merging procedure reduces the number of clusters below $K_{\max}$.
At each step, the pair of clusters whose merger incurs the smallest JS divergence loss is identified and merged.
After each merge, the relative information loss is evaluated as $|I(X_i; X_k \mid \mathbf{Z}) - I(X_i; X_k \mid E)| \,/\, I(X_i; X_k \mid \mathbf{Z})$, where $I(\cdot;\cdot\mid\cdot)$ denotes conditional mutual information estimated via plug-in discrete estimators \citep[npeet;][]{ver2014npeet}.
Merging continues as long as the relative loss remains below a tolerance of $5\%$, and stops at a minimum of $K_{\min} = 2$ clusters.
The result is a deterministic mapping $f_{i\to k}$ from joint remainder states to a compact discrete code whose cardinality adapts to the complexity of each edge.

\paragraph{Greedy CMI subset selection.}
Greedy CMI is a forward feature-selection wrapper that selects a small subset of the remainder variables whose joint state preserves the conditional mutual information $I(X_i; X_k \mid \mathbf{Z})$.
Starting from an empty set, it greedily adds the variable that minimises the relative information loss at each step, stopping when the loss falls below $5\%$ or when a maximum of 4 variables have been selected.
The embedding is the joint categorical code of the selected variables.
This method directly targets information preservation without intermediate dimensionality reduction.

\subsection{MI-based feature selection methods}
\label{app:embed_featsel}

\paragraph{Joint Mutual Information (JMI).}
JMI \citep{yang1999jmi} is a filter-based feature selection criterion that scores each candidate feature by its joint mutual information with the target, accounting for redundancy among already-selected features.
We use the implementation from the scikit-feature library \citep{li2018scikitfeature} with $k = 3$ features selected.
The embedding is the joint categorical code of the three selected remainder variables.

\paragraph{Conditional Mutual Information Maximisation (CMIM).}
CMIM \citep{fleuret2004cmim} selects features by maximising the minimum conditional mutual information between each candidate feature and the target, given every previously selected feature.
This criterion is more conservative than JMI, as it penalises redundancy more aggressively.
We use the scikit-feature implementation with $k = 3$ selected features, and the embedding is again the joint categorical code.

\paragraph{ReliefF.}
ReliefF \citep{kononenko1994relieff} is an instance-based feature weighting algorithm that evaluates features by their ability to distinguish between instances from different classes and instances from the same class, using nearest-neighbour distances.
We use the skrebate implementation \citep{urbanowicz2018skrebate} with 100 neighbours and $k = 3$ top-ranked features.
The embedding is the joint categorical code of the selected variables.

\subsection{Clustering methods}
\label{app:embed_clustering}

\paragraph{$k$-modes.}
The $k$-modes algorithm \citep{huang1998kmodes} is a categorical analogue of $k$-means that uses the Hamming distance and a frequency-based mode update rule.
We use the kmodes Python package with Huang initialisation, $n\_\text{init} = 5$ random restarts, and $K = 6$ clusters.
Cluster labels serve directly as the discrete embedding.

\paragraph{Spectral clustering on Hamming affinity.}
Spectral clustering \citep{ng2001spectral} is applied to a Hamming-based affinity matrix.
The pairwise Hamming distance matrix is converted to an affinity matrix via a Gaussian kernel $A_{ij} = \exp(-\gamma \cdot d_H(z_i, z_j))$ with bandwidth $\gamma = 5.0$.
Spectral clustering with $K = 6$ clusters and $k$-means label assignment is then performed on the Laplacian eigenvectors using scikit-learn \citep{pedregosa2011sklearn}.
To manage memory, rows are subsampled to 5{,}000 when the dataset is larger; remaining rows are assigned to the cluster of their nearest subsampled neighbour by Hamming distance.

\paragraph{Agglomerative clustering on Hamming distance.}
Agglomerative (hierarchical) clustering with average linkage is applied to the pairwise Hamming distance matrix using scikit-learn, with $K = 6$ clusters.
As with spectral clustering, datasets exceeding 5{,}000 rows are subsampled and out-of-sample rows are assigned via nearest-neighbour lookup.

\subsection{Dimensionality reduction followed by clustering}
\label{app:embed_dimred}

\paragraph{MCA $+$ $k$-means.}
Multiple Correspondence Analysis (MCA) is a dimensionality reduction technique for categorical data that generalises PCA to indicator matrices \citep{greenacre2017mca}.
We extract the first two MCA components using the prince Python library \citep{halford2023prince}, then apply $k$-means clustering ($K = 6$, $n\_\text{init} = 10$) to the two-dimensional scores.
Cluster labels form the discrete embedding.

\paragraph{Truncated SVD $+$ $k$-means.}
The remainder variables are one-hot encoded and projected onto two components via truncated singular value decomposition (SVD) using scikit-learn.
The two-dimensional projections are then clustered with $k$-means ($K = 6$, $n\_\text{init} = 10$).
This approach treats categorical data as sparse binary features before linear dimensionality reduction.

\paragraph{NMF $+$ $k$-means.}
Non-negative Matrix Factorisation (NMF) decomposes the one-hot encoded remainder matrix into two non-negative low-rank factors \citep{lee1999nmf}.
We extract two components using the \texttt{nndsvda} initialisation in scikit-learn, then apply $k$-means ($K = 6$, $n\_\text{init} = 10$) to the coefficient matrix.
The non-negativity constraint can produce parts-based representations that may capture interpretable response patterns.

\subsection{Supervised sufficient dimension reduction followed by binning}
\label{app:embed_sdr}

\paragraph{Sliced Inverse Regression (SIR) $+$ binning.}
SIR \citep{li1991sir} estimates the central dimension-reduction subspace by exploiting the inverse regression of predictors on the response.
Using the sliced Python package \citep{koepke2018sliced}, we project the one-hot encoded remainder onto two SIR directions with respect to the target $X_k$.
The two-dimensional projections are discretised via a $B \times B = 4 \times 4$ quantile grid, yielding up to 16 discrete categories.
Because SIR uses the target for projection, this method produces a target-directed embedding.

\paragraph{Partial Least Squares (PLS) $+$ binning.}
PLS regression \citep{wold1984pls} finds latent components that maximise the covariance between the (one-hot encoded) remainder and the target variable.
We extract two PLS components using scikit-learn's \texttt{PLSRegression} and discretise the scores with a $4 \times 4$ quantile grid, producing up to 16 categories.
Like SIR, PLS is target-directed, making the embedding tailored to information about $X_k$.

\subsection{Computational considerations}
\label{app:embed_computational}

Two practical decisions affect the computational footprint of the embedding benchmark.
First, for the two distance-based methods---spectral clustering on Hamming affinity and agglomerative clustering on Hamming distance---the pairwise distance matrix requires $O(n^2)$ memory.
For datasets exceeding 5{,}000 rows, we subsample to 5{,}000 observations (drawn without replacement using the global random seed) and assign out-of-sample observations to the cluster of their nearest subsampled neighbour.
This cap ensures that the $n \times n$ distance matrix fits in RAM while retaining the full dataset size for all other methods.

Second, all methods that involve random initialisation (e.g., $k$-means, $k$-modes, spectral label assignment) use a fixed random seed of 42 to ensure reproducibility across runs and computing environments.
The ACIB and Greedy CMI methods are deterministic given the data, so the seed affects only the subsampling step for distance-based methods and the initialisation of downstream clustering routines.

\paragraph{Amalgamation of the multivariate PID lattice.}
This appendix gives the full antichain classification rule summarised in the main text (Section~\ref{subsec:embedding_benchmark}). Each multivariate atom corresponds to an antichain: a collection of source subsets, none of which contains another. We first label each subset according to whether it contains only $X_i$ (label ``A''), only non-focal sources (label ``B''), or both $X_i$ and at least one other source (label ``AB''). We then apply a conservative classification rule: if any A or B label is present in the antichain, all AB labels are discarded, and the atom is assigned based on the remaining labels:
\begin{itemize}
  \item source-unique if only A labels remain;
  \item remainder-unique if only B labels remain;
  \item redundancy if both A and B labels remain;
  \item synergy only if the antichain contained exclusively AB labels (i.e., no pure A or B subset existed).
\end{itemize}
This rule is conservative with respect to synergy: an atom is classified as synergistic only when every subset in its antichain necessarily spans both the focal source and the remainder group, with no subset attributable to either alone. When a pure-A or pure-B subset coexists with an AB subset, the atom is attributed to the group that can account for it without invoking cross-group interaction. The grouping preserves the two-source consistency relations: the amalgamated source-unique plus redundancy equals $I(X_i; X_t)$, and the amalgamated remainder-unique plus redundancy equals $I(\{X_j : j \neq i\}; X_t)$.

\paragraph{Software.}
\label{sec:software}
PID was computed using the \texttt{dit} Python library~\citep{james2018dit}, with the $I_\text{mmi}$ (minimum mutual information) redundancy measure as the primary PID specification. As robustness checks we additionally computed decompositions with the $I_\text{min}$ (Williams--Beer)~\citep{williams2010nonnegative}, $I_{\pm}$ (Finn--Lizier)~\citep{finn2018pointwise}, and $I_{\wedge}$ (G\'acs--K\"orner)~\citep{gacs1973common} measures. For $N \geq 5$ sources, $I_{\pm}$ and $I_{\wedge}$ were computed via the fast M\"obius transform using precomputed lattice data from the \texttt{algebraicPID} package~\citep{jansma2025fastmobius}. Mutual information was estimated using plug-in frequency-based estimators via \texttt{npeet}~\citep{ver2014npeet}; conditional mutual information for the PHQ-9 analyses was computed using \texttt{pyitlib}. Rank-based partial correlations for the PHQ-9 analyses were computed using the \texttt{pingouin} package~\citep{vallat2018pingouin}.

\subsection{Normalisation and effect-size scaling}
\label{app:normalisation}
Because information-theoretic measures are expressed in bits and are bounded above by the uncertainty of the target, we report normalised effect sizes to facilitate comparisons across targets and across cohorts.

For conditional mutual information (CMI), we report the target-normalised quantity
\begin{equation}
\mathrm{nCMI}_{X\rightarrow Y}
\;=\;
\frac{I(X;Y\mid \mathbf{Z})}{H(Y)} \in [0,1],
\label{eq:ncmi}
\end{equation}
where $\mathbf{Z}$ denotes the remaining symptoms.

For embedding-based PID atoms about a given target $Y$, we report shares of the target entropy,
\begin{equation}
\pi^{(\cdot)}_{X\rightarrow Y} \;=\; \frac{\mathrm{Atom}^{(\cdot)}(X, E_{X\rightarrow Y}; Y)}{H(Y)} \in [0,1],
\label{eq:pid_shares}
\end{equation}
where $\mathrm{Atom}^{(\cdot)}$ denotes source-unique, remainder-unique, redundant, or synergistic information from the two-source decomposition on $(X, E_{X\rightarrow Y}; Y)$, and $E_{X\rightarrow Y}$ is the embedding of the remaining symptoms. Both quantities lie in $[0,1]$ and read as the fraction of the target's marginal uncertainty explained via a given information channel.

PHQ-9 items are ordinal and empirically skewed toward low-severity response categories, which reduces their effective entropy and constrains the attainable magnitude of mutual information and related quantities; this motivates reporting normalised effect sizes rather than raw bits. The $\mathrm{nCMI}$ normalisation (Eq.~\ref{eq:ncmi}) is conservative because $H(Y)\geq H(Y\mid \mathbf{Z})$, and it allows $\mathrm{nCMI}$ to be interpreted as the fraction of the target's marginal uncertainty associated with the source after conditioning on the rest of the symptom set. The PID shares (Eq.~\ref{eq:pid_shares}) read as ``the fraction of $H(Y)$ explained via a given information channel'' and are reported alongside unnormalised totals in bits when needed.

\paragraph{Visualisation pruning.}
\label{sec:network_visualisation}
Statistical significance alone can yield dense networks in large samples (especially in UK Biobank). To obtain readable network figures and to enable direct cross-cohort comparison, we apply a two-stage rule: (i) edge screening by CMI significance, followed by (ii) rank-based pruning for visualisation by retaining, for each target, the incoming edges with the largest normalised effect size $\mathrm{nCMI}_{X\rightarrow Y}$ (Eq.~\ref{eq:ncmi}). The number of retained edges per target is specified in the figure captions.

\FloatBarrier
\section{Embedding Background and Computational Scaling}
\label{app:embedding_scaling}

\paragraph{Relation to information bottleneck and redundancy bottleneck.}
In the main analyses with real PHQ-9 data, we use the Agglomerative Conditional Information Bottleneck (ACIB) embedding, which is target-directed:
it iteratively merges states of the multivariate remainder to minimise information loss about $X_k$ under the fixed-cardinality constraint.
ACIB is estimated separately for each $(X_i,X_k)$ pair using $\mathbf{X} \setminus \{X_i,X_k\}$ as inputs and $X_k$ as the supervised target. ACIB can be viewed as an instance of the information bottleneck principle~\citep{tishby1999information}: it constructs a compressed discrete representation of the multivariate remainder $E_{i\to k}=f_{i\to k}(\mathbf{X}\setminus\{X_i,X_k\})$ that preserves information about the target $X_k$.
More precisely, ACIB solves a discrete variant of the information bottleneck: it seeks to minimise $I(Z_{i\to k}; E_{i\to k})$ (compression) subject to the constraint $I(E_{i\to k}; X_k) \geq (1-\delta)\cdot I(Z_{i\to k}; X_k)$, where $\delta$ is a loss tolerance set to 5\% in our analyses.
The greedy merging phase traverses the IB rate--distortion curve from high cardinality (many clusters) to low cardinality, stopping when further compression would violate the information-preservation constraint.
The 5\% loss tolerance corresponds to operating near the ``knee'' of the IB curve, the regime where additional compression incurs disproportionately large information loss about the target.
This connection provides principled guidance for the tolerance hyperparameter: it should be set small enough to preserve target-relevant structure while allowing sufficient compression for tractable PID computation.
Recent theoretical work has further shown that a principled notion of PID redundancy can itself be formulated as a bottleneck trade-off via the \emph{redundancy bottleneck}, yielding redundancy--compression curves and efficient optimisation procedures \citep{kolchinsky2024redundancy_bottleneck}.
Whereas redundancy-bottleneck formulations provide a decision-theoretic grounding for redundancy, we use bottleneck-style compression as a practical approximation step that enables scalable two-source PID (including synergy) in high-dimensional symptom systems.
Because $E_{i\to k}$ is learned without access to the focal source $X_i$, a theoretical failure mode arises when the multivariate remainder is informative about $X_k$ primarily through its interaction with $X_i$ (i.e., interaction-only motifs); in such settings, any lossy compression of the remainder can attenuate synergy estimates.
In the present application, symptoms exhibit broad marginal dependence and do not resemble near-deterministic XOR-like constructions in which marginal associations vanish, reducing the plausibility of this edge case for PHQ-9 networks.

\paragraph{Computational scaling.}
A full $n$-source PID requires bookkeeping over a lattice whose number of atoms grows according to the Dedekind numbers, making exact multivariate decompositions infeasible beyond small $n$. By contrast, our approximation replaces the multivariate remainder by a low-cardinality discrete embedding and then computes a two-source PID on $(X_i,E_{i\to k};X_k)$. For fixed item cardinality $K$ and embedding size $|E_{i\to k}|=K$, the PID step depends only on a $K^3$ contingency table, while the dominant cost lies in constructing $E_{i\to k}$ for each ordered pair $(i,k)$. This yields a scalable $O(p^2)$ pipeline in the number of symptom pairs, with the per-pair embedding cost determined by the number of observed joint states of the remainder.

\subsection{Toy SCM: methods comparison under BROJA}
\label{app:toy_methods_robustness}

\begin{figure*}[!t]
  \centering
  \includegraphics[width=\textwidth]{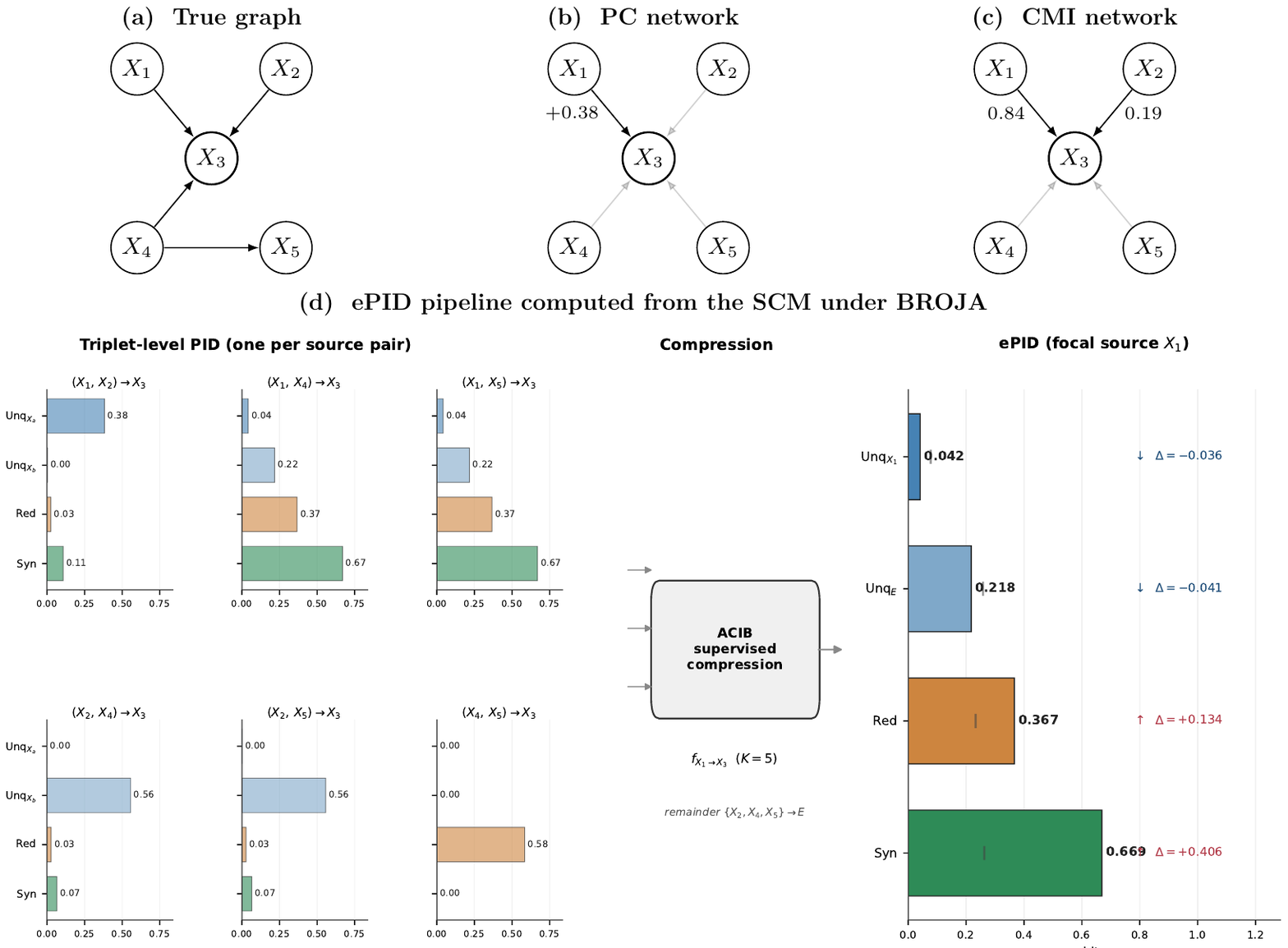}
  \caption{\apacap{Methods comparison under BROJA}{Companion to main-text Figure~\ref{fig:method_comparison}, identical structural model and pipeline, with the bottom-row two-source PIDs computed under the BROJA redundancy measure~\citep{bertschinger2014broja} instead of $I_{\min}$~\citep{williams2010nonnegative}. The top row (panels \textbf{a}--\textbf{c}: true graph, PC, CMI) is measure-independent. On the homogeneous source pair $(X_1, X_2)$, both source variables live in the $h$-channel of $X_3$ and the two measures agree to several decimals. On heterogeneous source pairs $(X_a, X_b)$ where one source lives in the $h$-channel and the other in the $W$-channel of $X_3$, $I_{\min}$ allocates information across redundancy and synergy while BROJA assigns the same information to pure uniques: a known divergence between the two measures that motivates BROJA's operationalist definition. BROJA $=$ Bertschinger--Rauh--Olbrich--Jost--Ay; PC $=$ partial correlation; CMI $=$ conditional mutual information.}}
  \label{fig:method_comparison_broja}
\end{figure*}

Figure~\ref{fig:method_comparison_broja} recomputes the bottom-row ePID pipeline of main-text Figure~\ref{fig:method_comparison} under the BROJA redundancy measure~\citep{bertschinger2014broja} on the identical SCM. The top-row panels (true graph, PC, CMI) are measure-independent and therefore reproduced unchanged from the main-text figure. The redundancy lattice underlying the type-1 amalgamation, alongside the truth-versus-ePID recovery for the focal source $X_1$, is shown in main-text Figure~\ref{fig:method_3panel}.

\FloatBarrier
\section{Bayesian Network Calibration Statistics}
\label{app:bn_calibration}

The following figures report calibration diagnostics for the ensemble of 50 synthetic Bayesian networks used in the embedding benchmark (Section~\ref{subsec:embedding_benchmark}). We compare the mutual information distributions of the synthetic networks against the empirical UK Biobank data to verify that the synthetic data preserve the statistical properties relevant to PID computation.

\begin{figure}[htbp]
    \centering
    \begin{subfigure}[b]{0.44\textwidth}
        \centering
        \includegraphics[width=\textwidth]{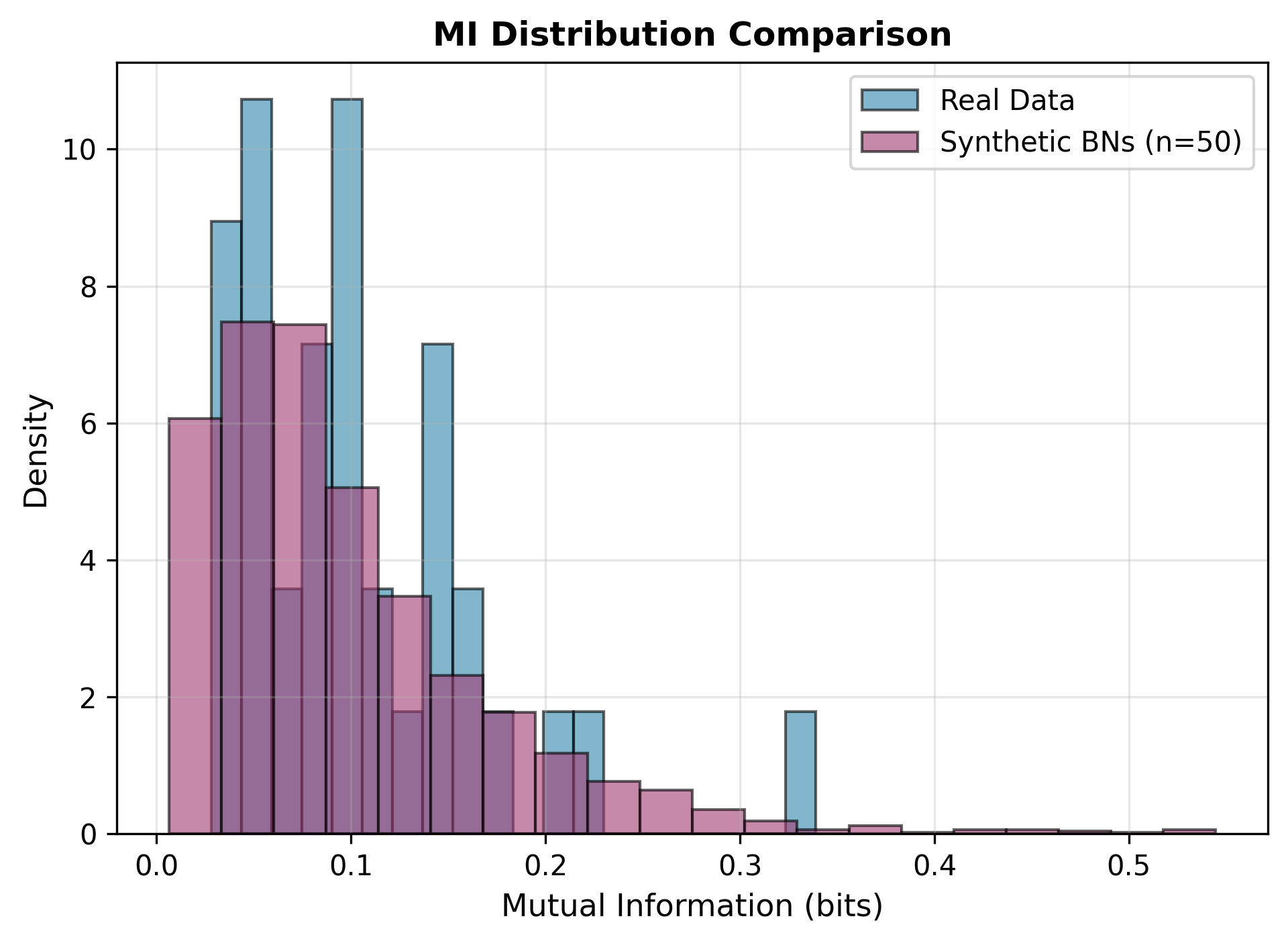}
        \caption{MI Distribution Comparison}
        \label{fig:mi_histogram}
    \end{subfigure}
    \hfill
    \begin{subfigure}[b]{0.48\textwidth}
        \centering
        \includegraphics[width=\textwidth]{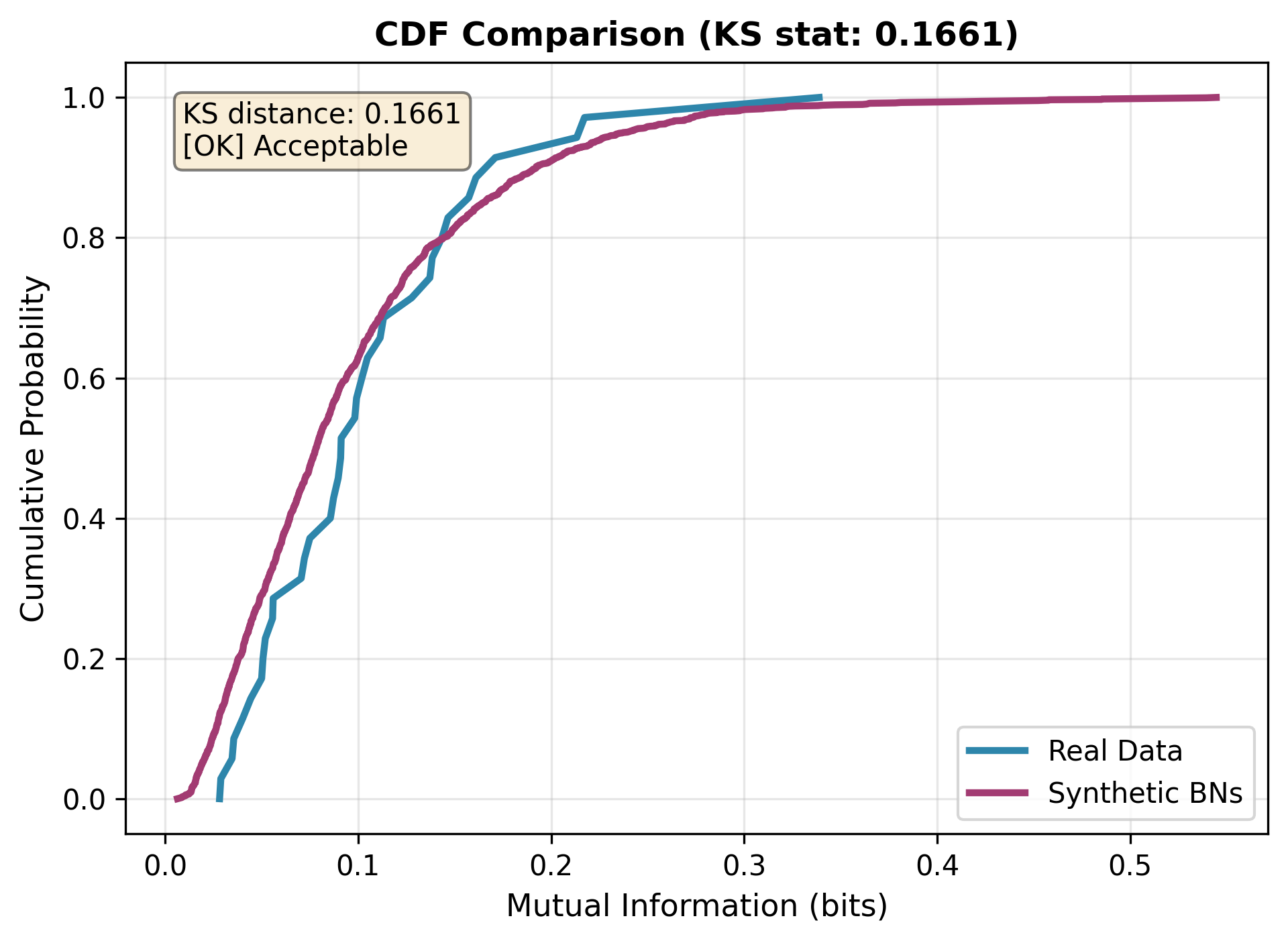}
        \caption{CDF Comparison (KS stat: 0.166)}
        \label{fig:mi_cdf}
    \end{subfigure}
    \caption{\apacap{Mutual information distribution comparison between real data and synthetic Bayesian network ensemble}{(a) Histogram overlay showing density distributions of MI values. (b) Cumulative distribution functions (CDFs) with Kolmogorov--Smirnov (KS) statistic quantifying the distributional match. The ensemble of 50 synthetic BNs closely matches the real-data distribution, validating the calibration procedure.}}
    \label{fig:mi_distribution_comparison}
\end{figure}

\begin{figure}[htbp]
    \centering
    \begin{subfigure}[b]{0.44\textwidth}
        \centering
        \includegraphics[width=\textwidth, height=6cm]{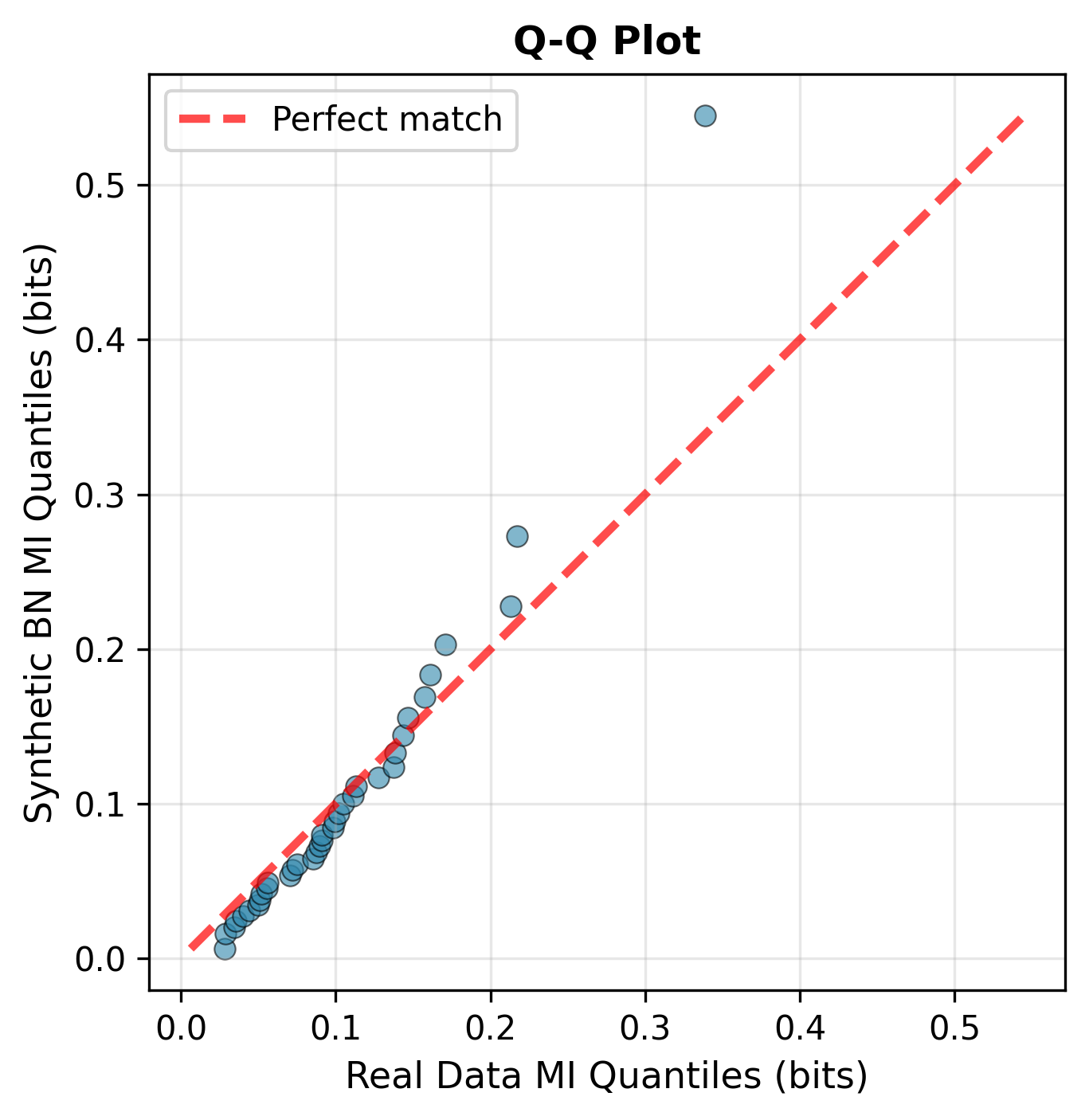}
        \caption{Q-Q Plot}
        \label{fig:qqplot}
    \end{subfigure}
    \hfill
    \begin{subfigure}[b]{0.48\textwidth}
        \centering
        \includegraphics[width=\textwidth]{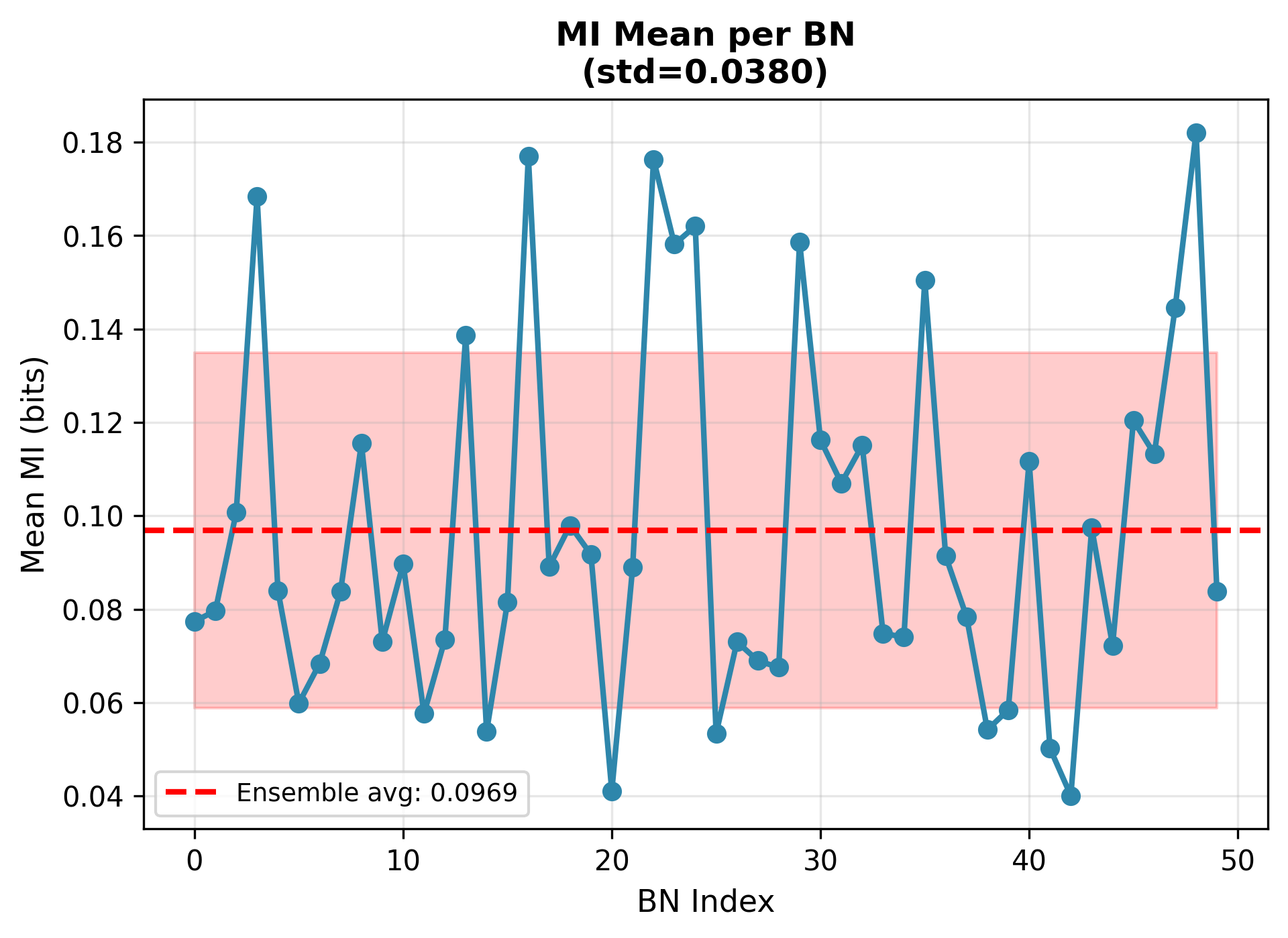}
        \caption{Ensemble Variability}
        \label{fig:ensemble_variability}
    \end{subfigure}
    \caption{\apacap{Detailed validation of Bayesian network ensemble calibration}{(a) Quantile--quantile (Q--Q) plot comparing MI quantiles between real data and synthetic BNs. Points close to the diagonal indicate good distributional match. (b) Mean MI per network across the ensemble, showing consistent calibration with low variability (std $=$ 0.0382). The red dashed line indicates the ensemble average, with shaded region showing $\pm 1$ standard deviation.}}
    \label{fig:detailed_validation}
\end{figure}

\FloatBarrier
\section{Multi-Dataset Benchmark: Full Results and Statistical Tests}
\label{app:benchmark_stats}

This appendix reports the full multi-dataset embedding benchmark whose headline is summarised in the main text (Section~\ref{subsec:results_embedding_benchmark}): per-measure fidelity and method rankings, the $I_{\pm}$ exception, and the omnibus and post-hoc statistical tests.

\subsection{Multi-dataset embedding benchmark}
\label{subsec:results_multidataset}

To assess embedding fidelity beyond the synthetic benchmark, we evaluated 13 embedding methods across 83 real-world datasets, 5 PID measures, and $N \in \{3,4,5\}$ sources (Section~\ref{subsec:multi_dataset_benchmark}). Figure~\ref{fig:benchmark_all_methods} summarises the results across all methods; Table~\ref{tab:benchmark_summary} reports key statistics for the recommended method (ACIB) at $N=5$.

\paragraph{Embedding fidelity is measure-dependent.}
At $N=5$, mean relative error (RE) for ACIB ranged from 19.4\% ($I_{\wedge}$) to 115.1\% ($I_{\pm}$) (Table~\ref{tab:benchmark_summary}), with corresponding mean absolute errors (MAE) of 0.057--0.247 bits. To put these in scale, typical ground-truth atom magnitudes are 0.05--0.20 bits, so an RE of 45\% reflects MAE on the same order as a single atom rather than near-total signal loss; for comparison, a trivial ``predict-zero'' baseline yields RE > 130\% for $I_{\min}$. $I_{\wedge}$ (G\'{a}cs--K\"{o}rner) was best preserved because its decomposition is sparsest, with most atoms exactly zero and the few non-zero atoms recovered with <13\% per-atom RE. $I_{\min}$ and $I_{\mathrm{mmi}}$ showed moderate distortion (RE $\approx$ 45\%; AE $\approx$ 0.10 bits), driven almost entirely by the remainder-unique and synergy atoms (Section~\ref{subsec:results_ipm_failure}); the redundancy and source-unique atoms were recovered with <3\% RE in absolute terms. $I_{\mathrm{rr}}$ showed substantial distortion (RE $=$ 70.9\%), and $I_{\pm}$ errors exceeded the total signal (RE $>$ 100\%) due to a signed-atom cancellation failure analysed in Section~\ref{subsec:results_ipm_failure}, which persists even when the dominant remainder-unique error component is removed.

\begin{figure}[!t]
  \centering
  \includegraphics[width=0.95\linewidth]{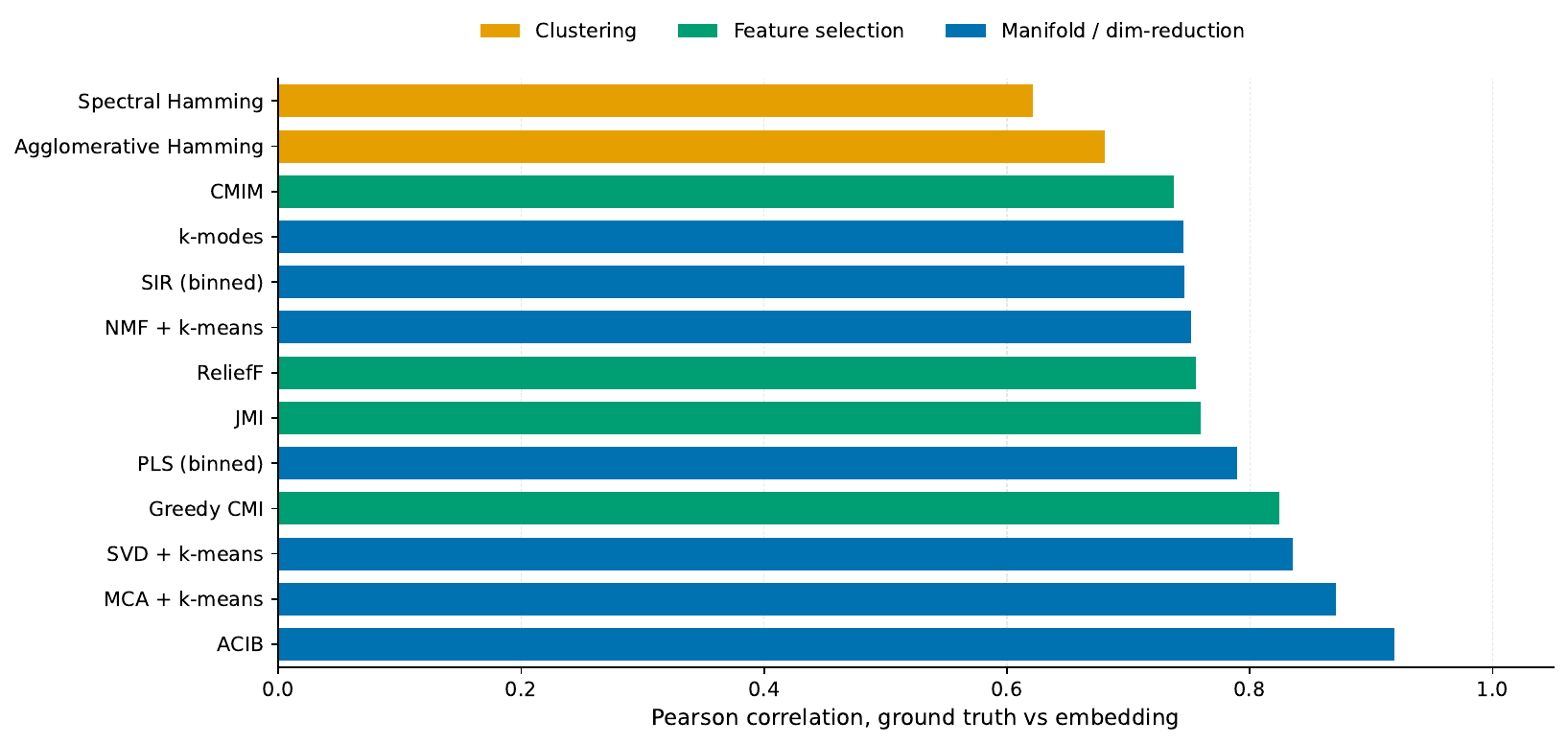}
  \caption{\apacap{Ranking of 13 embedding methods by ground-truth synergy correlation}{Pearson correlation $r$ between embedding-approximated and ground-truth synergy across the synthetic Bayesian network benchmark (50 networks $\times$ 9 targets $\times$ 3 source counts $N \in \{3, 4, 5\}$, $n = 5{,}377$ source--target configurations; type-1 amalgamation throughout). Bars are coloured by method family: \emph{manifold / dim-reduction} (blue), \emph{feature selection} (green), \emph{clustering} (orange). ACIB ranks first by this criterion; manifold-family methods occupy the top six positions. ACIB $=$ Agglomerative Conditional Information Bottleneck; JMI $=$ Joint Mutual Information; CMIM $=$ Conditional Mutual Information Maximisation; SIR $=$ Sliced Inverse Regression; MCA $=$ Multiple Correspondence Analysis; PLS $=$ Partial Least Squares; SVD $=$ Singular Value Decomposition; NMF $=$ Non-negative Matrix Factorisation.}}
  \label{fig:method_ranking}
\end{figure}

\paragraph{ACIB is the recommended default for non-$I_{\pm}$ measures.}
ACIB delivered the lowest approximation error on the largest share of datasets across non-$I_{\pm}$ measures, and its advantage grew with the number of sources. To make this claim concrete we ranked the 13 embedding methods on each dataset by their per-atom absolute error and counted, for every measure$\times$$N$ combination, how often each method ranked first. Across all 15 measure$\times$$N$ combinations ACIB ranked first in 8 (all $N \geq 4$ except $I_{\pm}$); pooling per-dataset comparisons, ACIB ranked first in 44.3\% of 1{,}141 cases overall, and in 55.7\% once $I_{\pm}$ is excluded. At $N=5$, ACIB ranked first on 84.4\% of datasets for $I_{\wedge}$ (95\% CI: [78.4, 91.1]\%), 77.9\% for $I_{\min}$ [71.1, 85.0], and 75.3\% for $I_{\mathrm{mmi}}$ [68.6, 82.9]; the next-best methods (CMIM: 10.5\%, ReliefF: 7.4\%) trailed substantially. ACIB's advantage is not that it is dramatically better at any single $N$, but that it degrades most slowly as $N$ grows: its error growth slope was 0.041 [95\% CI: 0.035, 0.046], versus 0.044 [0.039, 0.049] for SIR and PLS, and 0.046 [0.040, 0.053] for ReliefF. At $N=3$, where the embedding compresses only two sources, methods were statistically indistinguishable (Friedman $\chi^2 = 12.0$, $p = 0.45$).
Figure~\ref{fig:method_ranking} summarises method ordering on the synthetic benchmark by per-pair Pearson correlation between approximated and ground-truth synergy; the family-level separation is visible at a glance, and the win-rate ordering reported above is preserved.

\paragraph{Number of sources matters more than method choice.}
Approximation error roughly tripled from $N=3$ to $N=5$ regardless of method. At $N=3$, methods were indistinguishable (as noted above; Nemenyi post-hoc: 0 significant pairs for all measures). At $N=4$ and $N=5$, method differences became highly significant (Friedman $p < 10^{-22}$). The Nemenyi post-hoc test, which compares mean ranks across all $\binom{13}{2}=78$ method pairs while controlling the family-wise error rate, identified 14--37 significant pairs at $N=4$~\citep{demsar2006}. This pattern---visible as convergent lines at $N=3$ separating sharply at higher $N$ in Figure~\ref{fig:benchmark_all_methods}---indicates that the fundamental compression challenge of higher-dimensional remainder sets dominates the choice of compression algorithm. Full statistical test results are provided in the following subsections.

\begin{table}[ht]
\centering
\caption{\apacap{ACIB embedding fidelity by PID measure at $N=5$ sources}{Mean absolute error (MAE, bits) is the per-atom absolute error averaged across the four PID atoms (source-unique, remainder-unique, redundancy, synergy) and across datasets. Mean relative error (RE) is the same quantity divided by the magnitude of the corresponding ground-truth atom. Win rate is the proportion of per-dataset contests (across 77--83 datasets) in which ACIB achieved the lowest error among the 13 candidate embeddings, with bootstrap 95\% confidence intervals. For $I_{\mathrm{rr}}$, 12 of the 13 embeddings were available (greedy CMI lacked an $I_{\mathrm{rr}}$ ground truth).}}
\label{tab:benchmark_summary}
\small
\begin{tabular}{lcccc}
\toprule
Measure & MAE (bits) & RE (\%) & Win rate (\%) & 95\% CI \\
\midrule
$I_{\wedge}$ & 0.057 & 19.4 & 84.4 & [78.4, 91.1] \\
$I_{\min}$ & 0.105 & 44.7 & 77.9 & [71.1, 85.0] \\
$I_{\mathrm{mmi}}$ & 0.106 & 45.2 & 75.3 & [68.6, 82.9] \\
$I_{\mathrm{rr}}$ & 0.146 & 70.9 & 64.9 & [56.2, 73.3] \\
$I_{\pm}$ & 0.247 & 115.1 & 2.6 & [0.0, 4.4] \\
\bottomrule
\end{tabular}
\end{table}

\subsubsection{The $I_{\pm}$ exception}
\label{subsec:results_ipm_failure}

The high relative-error figures for $I_{\pm}$ reflect a structural mismatch between $I_{\pm}$'s information-flow semantics and the compression that any ePID pipeline performs. $I_{\pm}$ decomposes mutual information \emph{pointwise}, atom by atom over individual outcomes, and the resulting signed atoms encode local information flow that cancels in structured ways across the joint distribution~\citep{finn2018pointwise}. At $N=5$, 77.7\% of ground-truth $\mathrm{Unq}_{S_1}$ values were negative (compared to 0.7\% for $I_{\min}$). Embedding methods are designed to compress the joint distribution of the remainder into a single discrete summary; this compression is well-suited to atoms that average over the joint distribution, but it aggregates over precisely the local outcome-level structure that $I_{\pm}$ relies on, so cancellations no longer line up after compression. The mismatch is therefore conceptual, not a tunable embedding choice.

\paragraph{Practical recommendation.}
Based on the benchmark, we recommend ACIB for analyses using $I_{\min}$, $I_{\mathrm{mmi}}$, $I_{\mathrm{rr}}$, or $I_{\wedge}$ (best at $N \geq 4$); for $I_{\pm}$, ReliefF ($N \leq 4$) or SVD$+k$-means ($N=5$) achieved the lowest absolute errors in our runs, but the structural mismatch means embedding-based estimates of $I_{\pm}$ should be interpreted as approximate compositional profiles rather than exact quantities. At $N=3$, method choice is immaterial. The empirical PHQ-9 analyses in this paper use $I_{\mathrm{mmi}}$, for which ACIB is the best-validated embedding.

\subsection{Friedman omnibus test}

Table~\ref{tab:friedman} reports the Friedman test statistic and $p$-value for each combination of PID measure and number of sources. The Friedman test assesses whether the ranking of embedding methods differs significantly across datasets.

\begin{table}[H]
\centering
\caption{\apacap{Friedman test results by PID measure and number of sources}{$\chi^2$ statistic and $p$-value testing whether method rankings differ across datasets. At $N=3$, no significant differences were detected for any measure.}}
\label{tab:friedman}
\small
\begin{tabular}{l cc cc cc}
\toprule
& \multicolumn{2}{c}{$N=3$} & \multicolumn{2}{c}{$N=4$} & \multicolumn{2}{c}{$N=5$} \\
\cmidrule(lr){2-3} \cmidrule(lr){4-5} \cmidrule(lr){6-7}
Measure & $\chi^2$ & $p$ & $\chi^2$ & $p$ & $\chi^2$ & $p$ \\
\midrule
$I_{\min}$          & 12.0 & 0.45    & 131.4 & $9 \times 10^{-23}$ & 145.9 & $1 \times 10^{-25}$ \\
$I_{\mathrm{mmi}}$  & 12.0 & 0.45    & 128.6 & $3 \times 10^{-22}$ & 131.1 & $1 \times 10^{-22}$ \\
$I_{\mathrm{rr}}$   & 12.0 & 0.45    & 134.9 & $2 \times 10^{-23}$ &  37.6 & $9 \times 10^{-5}$  \\
$I_{\pm}$            & 12.0 & 0.45    & 252.7 & $8 \times 10^{-48}$ & 237.7 & $1 \times 10^{-44}$ \\
$I_{\wedge}$         & 12.0 & 0.45    & 181.8 & $4 \times 10^{-33}$ & 238.3 & $8 \times 10^{-45}$ \\
\bottomrule
\end{tabular}
\end{table}

\subsection{Nemenyi post-hoc tests}

Table~\ref{tab:nemenyi_pairs} reports the number of significantly different method pairs (out of 78 possible) identified by the Nemenyi post-hoc test following significant Friedman results.

\begin{table}[H]
\centering
\caption{\apacap{Number of significantly different method pairs (Nemenyi post-hoc)}{Out of 78 possible pairwise comparisons between 13 methods. Zero pairs at $N=3$ confirms that all methods perform equivalently when compressing only two sources.}}
\label{tab:nemenyi_pairs}
\small
\begin{tabular}{lccc}
\toprule
Measure & $N=3$ & $N=4$ & $N=5$ \\
\midrule
$I_{\min}$          & 0 & 15 & 22 \\
$I_{\mathrm{mmi}}$  & 0 & 14 & 19 \\
$I_{\mathrm{rr}}$   & 0 & 15 &  6 \\
$I_{\pm}$            & 0 & 37 & 36 \\
$I_{\wedge}$         & 0 & 23 & 35 \\
\bottomrule
\end{tabular}
\end{table}

\subsection{Supplementary benchmark figures}

\begin{figure}[H]
\centering
\includegraphics[width=\textwidth]{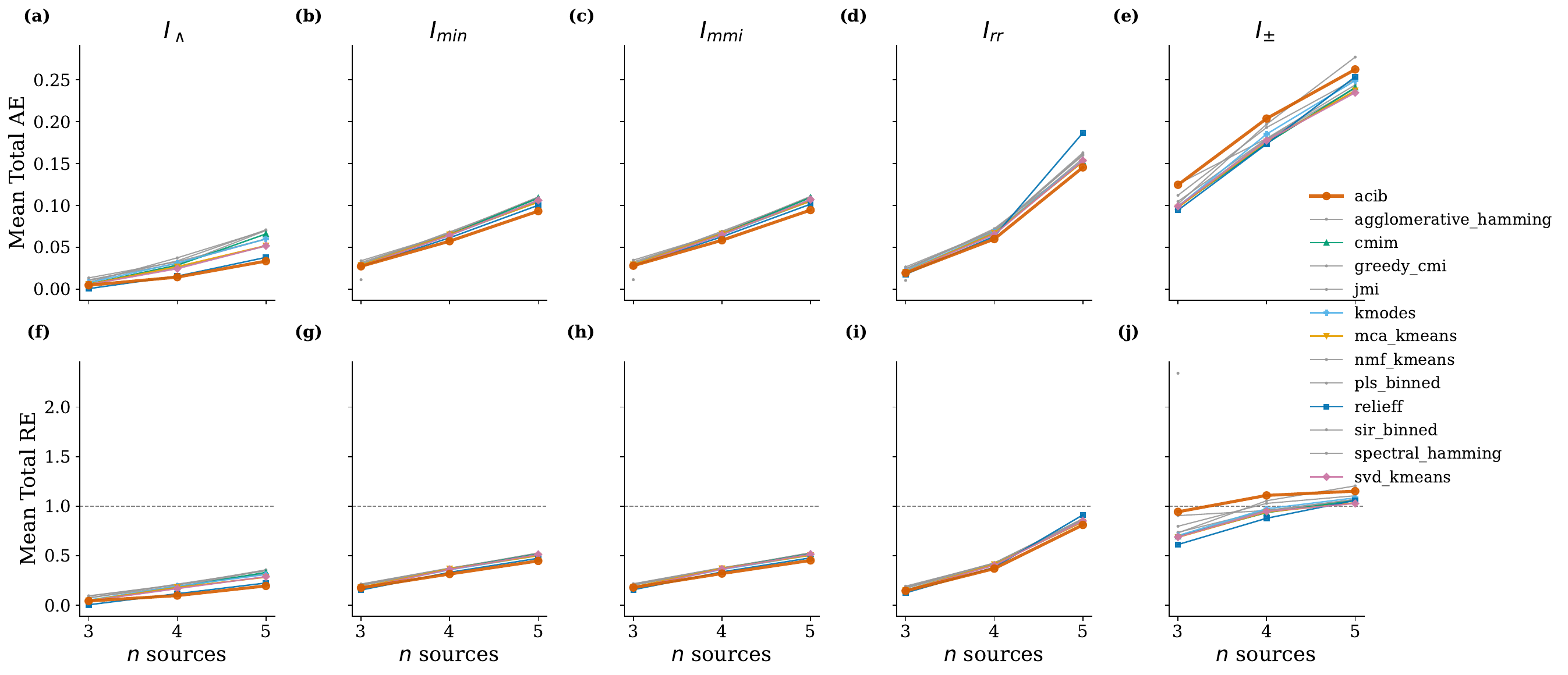}
\caption{\apacap{Embedding approximation error across PID measures and number of sources (all 13 methods)}{Top row: mean absolute error (bits); bottom row: mean relative error (\%). Each column corresponds to one PID measure, ordered by increasing difficulty. ACIB (bold orange) is the lowest-error method at $N \geq 4$ for all measures except $I_{\pm}$. The dashed line at RE~$= 100$\% marks the threshold where approximation error equals the total signal. $I_{\mathrm{rr}}$ at $N=5$ is based on 23 datasets; all others on 77--83 datasets.}}
\label{fig:benchmark_all_methods}
\end{figure}

Figure~\ref{fig:benchmark_all_methods} reports embedding approximation error across all 13 methods, 5 PID measures, and $N \in \{3,4,5\}$ sources.

\begin{figure}[H]
  \centering
  \includegraphics[width=\textwidth]{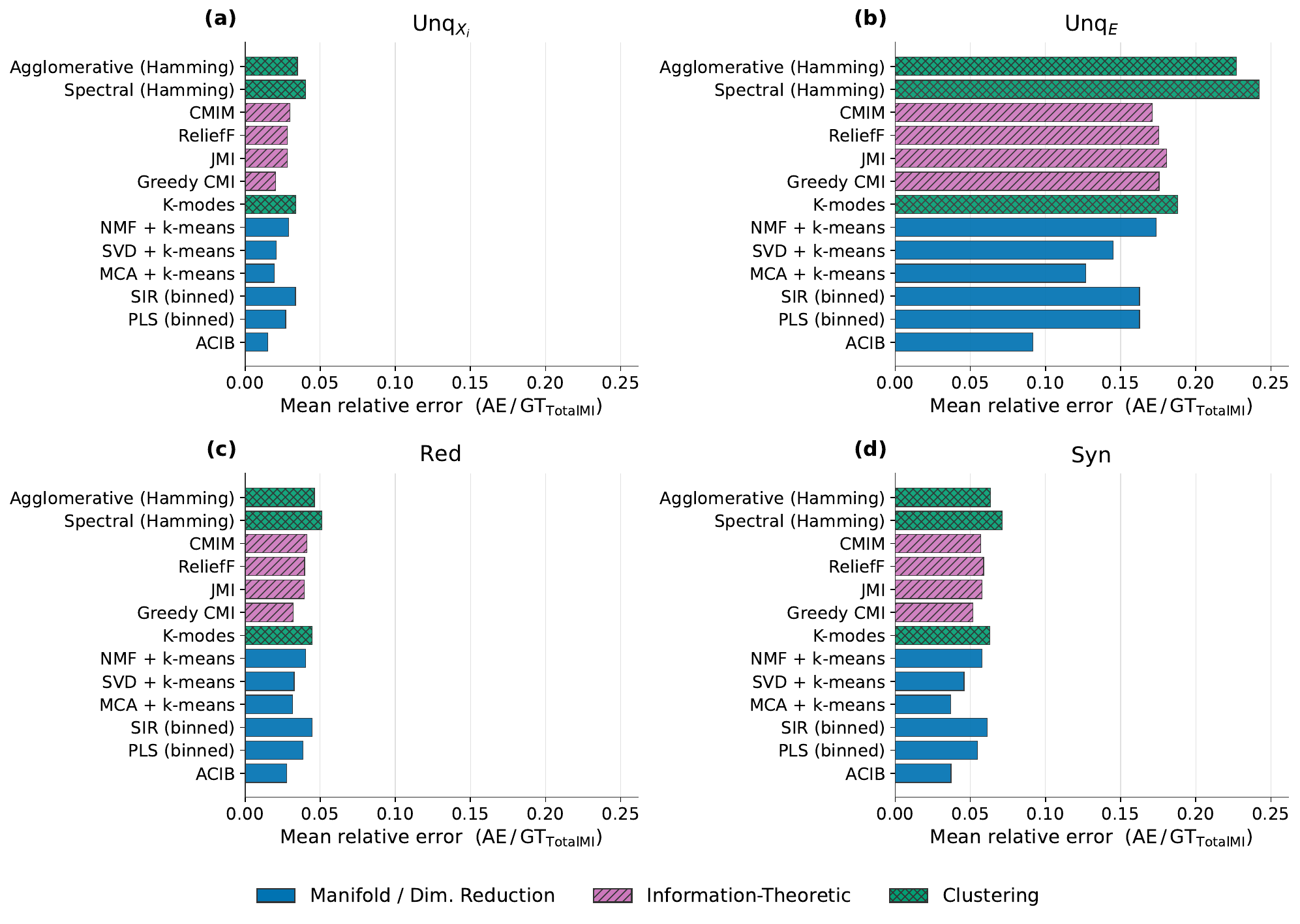}
  \caption{\apacap{Mean relative error by embedding method and PID atom}{Per-atom error normalised by the magnitude of the corresponding ground-truth atom, complementing the absolute-error view in Figure~\ref{fig:embedding_overall_performance} of the main text. Panels, methods, and family colour coding are identical to the main-text figure. Relative-error normalisation makes small absolute errors visible when ground-truth atoms are small in magnitude; the ranking of methods within each panel is preserved against the absolute-error view. ACIB $=$ Agglomerative Conditional Information Bottleneck.}}
  \label{fig:component_errors_re}
\end{figure}

Figure~\ref{fig:component_errors_re} reports the relative-error analogue of the main-text Figure~\ref{fig:embedding_overall_performance}. Normalising each absolute error by the magnitude of the corresponding ground-truth atom is informative because absolute errors are bounded above by atom magnitude: methods that appear comparable in bits can diverge sharply once normalised. The relative-error ranking is consistent with the absolute-error ranking, with ACIB dominant across all four atoms; the gap between manifold and clustering families widens slightly under relative-error scaling.

\subsection{Atom-level error profile across PID measures}
\label{app:component_profiles}
\begin{figure}[H]
\centering
\includegraphics[width=0.8\textwidth]{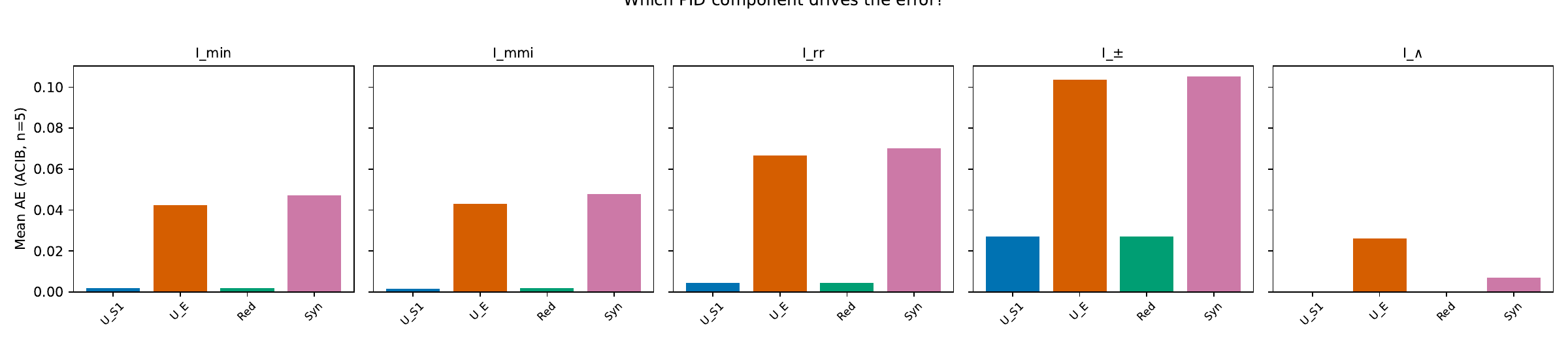}
\caption{\apacap{Atom-level error profiles by PID measure (ACIB, $N=5$)}{For non-$I_{\pm}$ measures, error is concentrated in the remainder-dependent atoms ($\mathrm{Unq}_E$, $\mathrm{Syn}$). For $I_{\pm}$, all atoms show high error due to disruption of signed cancellation structure.}}
\label{fig:appx_component_profiles}
\end{figure}

For non-$I_{\pm}$ measures, embedding error in the multi-dataset benchmark is concentrated in the atoms that involve the compressed remainder: $\mathrm{Unq}_E$ (remainder-unique) and $\mathrm{Syn}$ (synergy) carry most of the approximation error, while $\mathrm{Unq}_{S_1}$ (focal-source unique) and $\mathrm{Red}$ (redundancy) are well-preserved (Fig.~\ref{fig:appx_component_profiles}). For $I_{\pm}$, all four atoms exhibit high error, consistent with the disruption of signed cancellation across the full decomposition discussed in the main text.

\FloatBarrier
\section{Additional Fidelity Diagnostics and Symptom-Level Examples}
\label{app:additional_diagnostics}

This appendix collects three diagnostic items referenced from the main text: per-pair synergy fidelity for representative methods (Section~\ref{app:per_pair_scatter}), the scaling of synergy approximation error with the number of embedded sources (Section~\ref{app:synergy_scaling}), and target-centred PID examples for two representative PHQ-9 symptoms in the Xinxiang student sample (Section~\ref{app:symptom_pid_examples}).

\subsection{Per-pair synergy fidelity for representative methods}
\label{app:per_pair_scatter}
\begin{figure}[!t]
  \centering
  \begin{subfigure}[t]{0.32\linewidth}
    \centering
    \includegraphics[width=\linewidth]{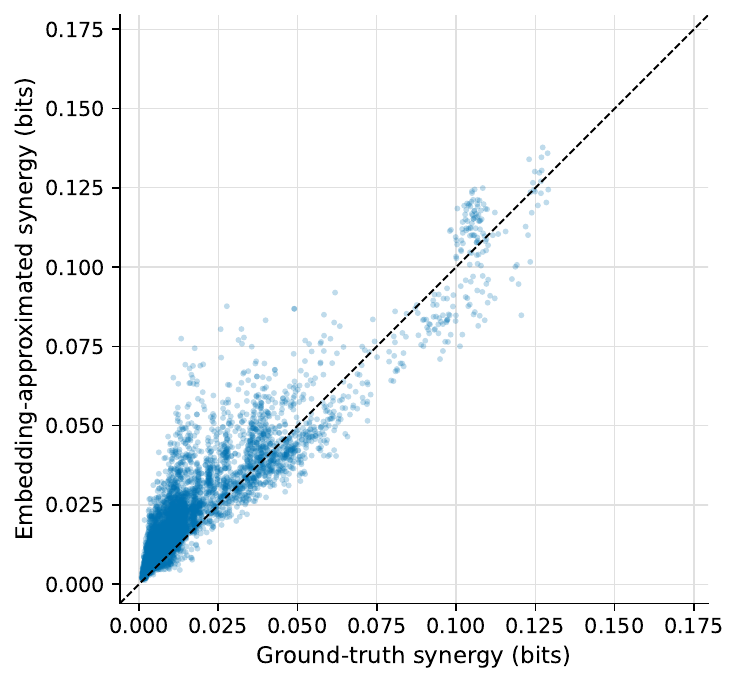}
    \caption{ACIB ($r=0.98$)}
    \label{fig:embedding_scatter_acib}
  \end{subfigure}\hfill
  \begin{subfigure}[t]{0.32\linewidth}
    \centering
    \includegraphics[width=\linewidth]{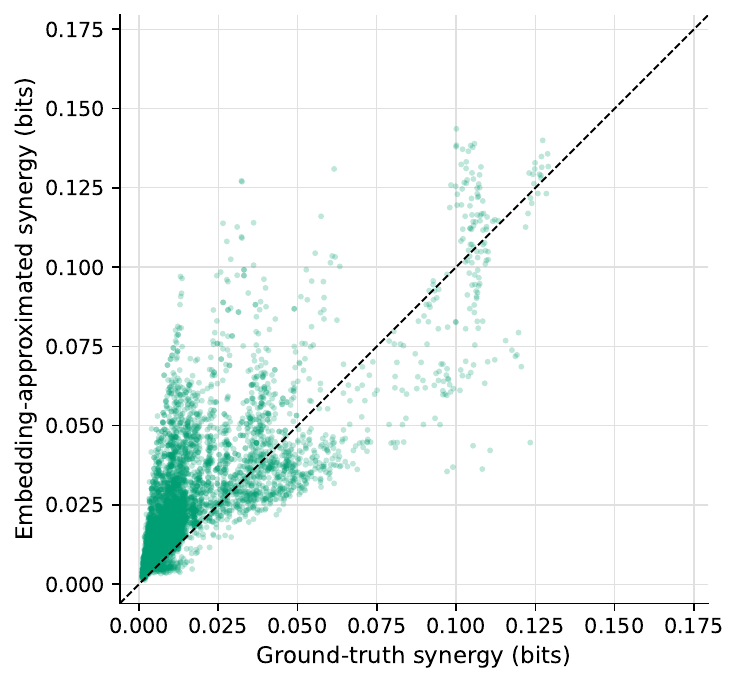}
    \caption{JMI ($r=0.41$)}
    \label{fig:embedding_scatter_jmi}
  \end{subfigure}\hfill
  \begin{subfigure}[t]{0.32\linewidth}
    \centering
    \includegraphics[width=\linewidth]{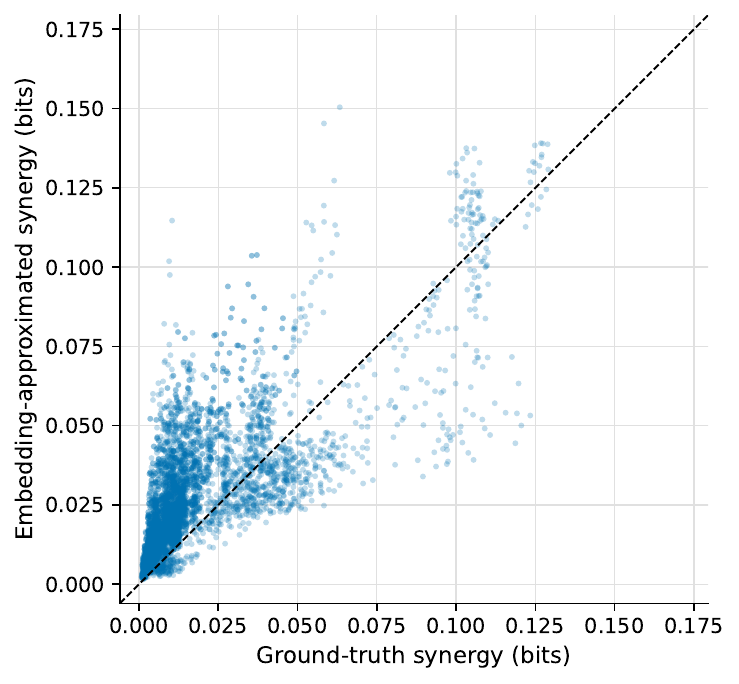}
    \caption{$k$-modes ($r=0.90$)}
    \label{fig:embedding_scatter_kmodes}
  \end{subfigure}
  \caption{\apacap{Approximated versus ground-truth synergy for one representative method per family tier}{Per-pair scatter of embedding-approximated synergy (y-axis) against ground-truth synergy (x-axis), in bits, across 5{,}377 ($G, X_t, N$) configurations from the synthetic Bayesian network benchmark. Each panel shows one representative method per family. \textbf{(a)} ACIB, manifold/dim-reduction family, Pearson $r=0.98$. \textbf{(b)} JMI, feature-selection family, $r=0.41$. \textbf{(c)} $k$-modes, clustering family, $r=0.90$. Dashed diagonal is the identity line. The visible density clustering near ground-truth synergy $\approx 0.10$ reflects multiple synthetic networks producing similar synergy values. Methods within each family vary in fidelity; main-text Figure~\ref{fig:method_ranking} ranks all 13 methods. The corresponding scaling of synergy approximation error with the number of embedded sources is reported in Section~\ref{app:synergy_scaling}. ACIB $=$ Agglomerative Conditional Information Bottleneck; JMI $=$ Joint Mutual Information.}}
  \label{fig:embedding_scatter}
\end{figure}

Figure~\ref{fig:embedding_scatter} reports per-pair scatter for one representative method from each family tier, illustrating the spread visible in the aggregate atom errors of main-text Figure~\ref{fig:embedding_overall_performance} and in the 13-method ranking of main-text Figure~\ref{fig:method_ranking}. ACIB (manifold/dim-reduction) aligns closely with the identity line ($r = 0.98$); JMI (feature selection) shows moderate scatter ($r = 0.41$); $k$-modes (clustering) aligns reasonably well ($r = 0.90$), illustrating the upper end of clustering performance. Hamming-based clustering methods (collapse to $r \approx 0$; see main-text Figure~\ref{fig:method_ranking}) are not shown in this trio.

\subsection{Synergy approximation error scaling with the number of sources}
\label{app:synergy_scaling}
\begin{figure}[H]
\centering
\includegraphics[width=0.55\textwidth]{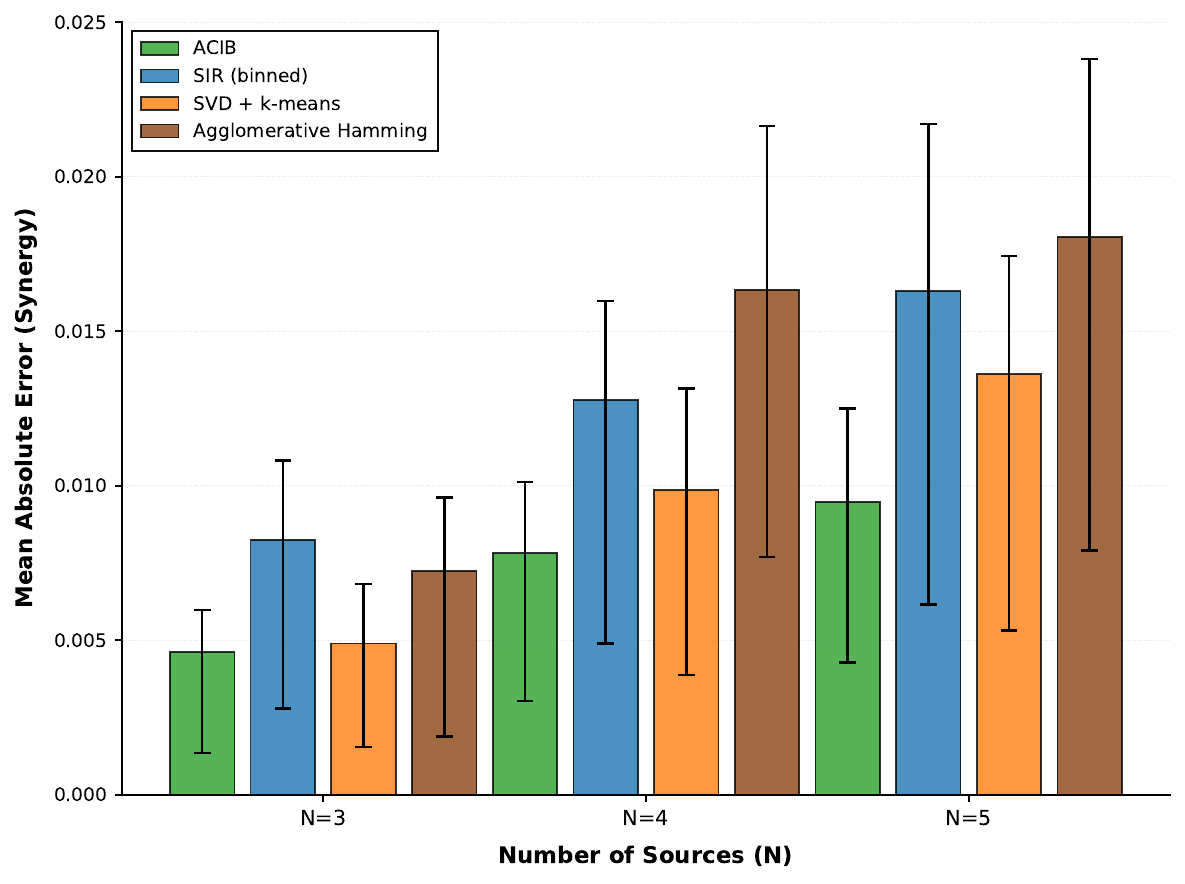}
\caption{\apacap{Synergy approximation error as a function of the number of embedded sources ($N=3,4,5$)}{ACIB error increases sub-linearly from $0.0008$ bits at $N=3$ to $0.0056$ bits at $N=5$.}}
\label{fig:embedding_synergy_scaling}
\end{figure}

For ACIB on the synthetic Bayesian network ensemble, mean absolute error in the synergy atom increased smoothly from $0.0008$ bits at $N=3$ to $0.0056$ bits at $N=5$, remaining below $0.006$ bits throughout (Fig.~\ref{fig:embedding_synergy_scaling}). A sliced inverse regression (SIR) embedding with binning showed slightly higher but remarkably stable synergy error across $N$ ($\approx 0.0045$--$0.0048$ bits), whereas linear SVD$+k$-means and clustering methods had both larger baseline errors and steeper increases with $N$.

\subsection{Symptom-level PID examples in the Xinxiang student sample}
\label{app:symptom_pid_examples}
\begin{figure}[H]
    \centering
    \begin{subfigure}[t]{0.48\textwidth}
        \centering
        \includegraphics[width=\linewidth]{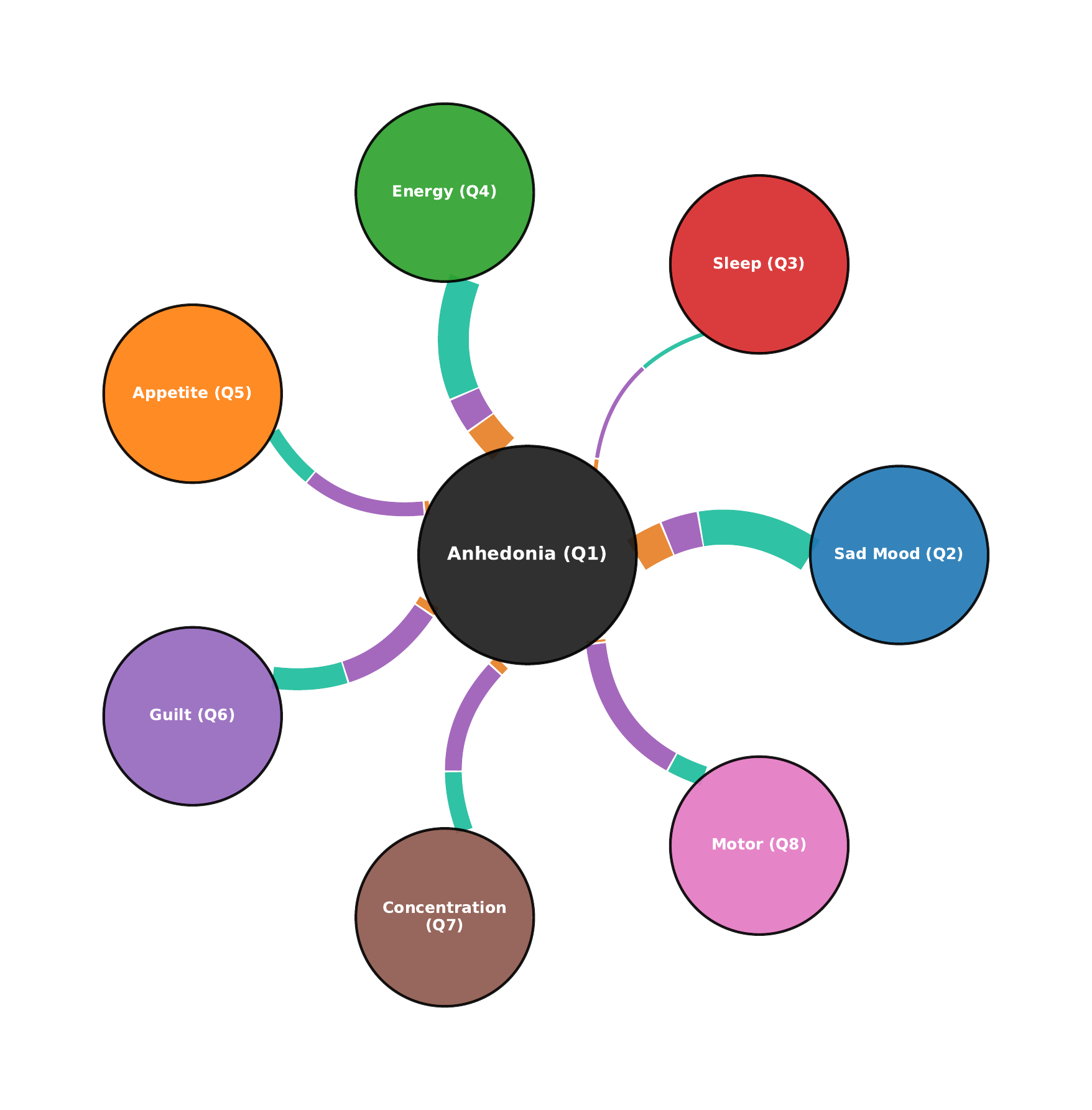}
        \caption{Anhedonia as target}
    \end{subfigure}
    \hfill
    \begin{subfigure}[t]{0.48\textwidth}
        \centering
        \includegraphics[width=\linewidth]{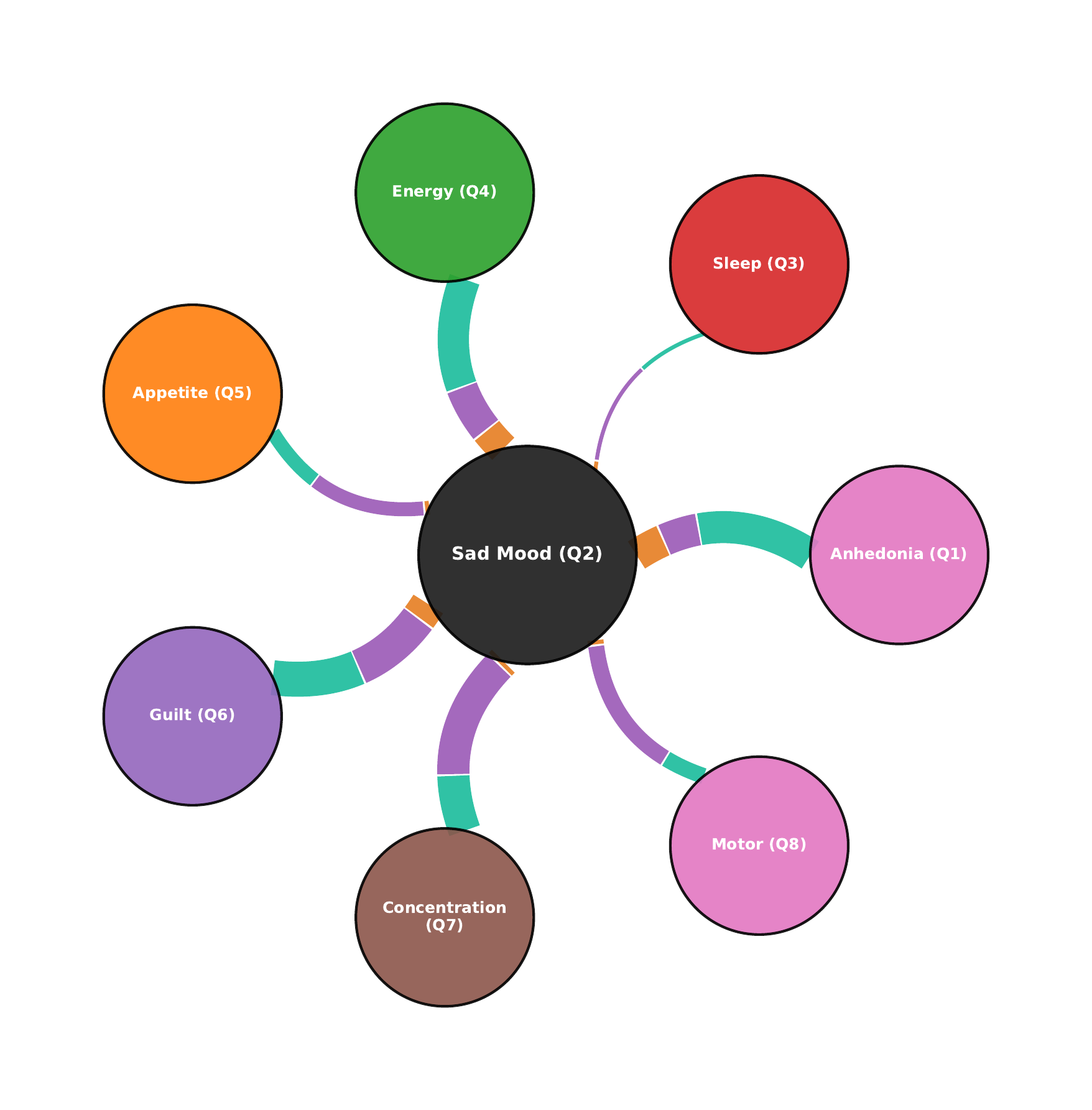}
        \caption{Sad mood as target}
    \end{subfigure}
    \caption{\apacap{Target-centred PID radial graphs for two representative targets in the Xinxiang student sample}{Each edge is segmented into four PID atoms (teal: redundancy; blue: unique source; purple: unique embedding; orange: synergy); edge thickness scales with total mutual information.}}
    \label{fig:appx_pid_decomp_pair}
\end{figure}

Figure~\ref{fig:appx_pid_decomp_pair} visualises the embedding-based PID for two representative targets in the Xinxiang student sample (anhedonia, sad mood) as target-centred radial graphs. The central dark grey node represents the target; surrounding coloured nodes represent source variables. Each edge is segmented into four PID components (teal: redundancy; blue: unique source; purple: unique embedding; orange: synergy), with edge thickness scaled to total mutual information. These examples illustrate how strongly connected affective symptoms (e.g., anhedonia and sad mood) combine substantial unique contributions with non-negligible synergy, whereas somatic symptoms such as sleep and motor disturbance more often contribute via synergy or redundancy rather than purely unique channels.

\section{Dependency Structure in Clinical Datasets}
\label{app:dependency_comparison}

To assess whether non-linear conditional dependencies are prevalent in the types of datasets used for symptom network analysis, we compared three unconditional bivariate dependency measures across 80 clinical and psychometric datasets (32{,}889 variable pairs in total).

For each dataset, we computed (i) absolute Pearson correlation on the imputed continuous data, (ii) absolute Spearman correlation on rank-transformed data, and (iii) mutual information on discretised data (8 equal-frequency bins). All three measures are unconditional and bivariate, ensuring direct comparability without confounding by conditioning structure. Statistical significance was assessed using permutation tests (1{,}000 permutations per dataset, columns shuffled independently) with Benjamini--Hochberg FDR correction at $\alpha = 0.05$, applied uniformly across all three measures.

\begin{figure}[H]
\centering
\includegraphics[width=\textwidth]{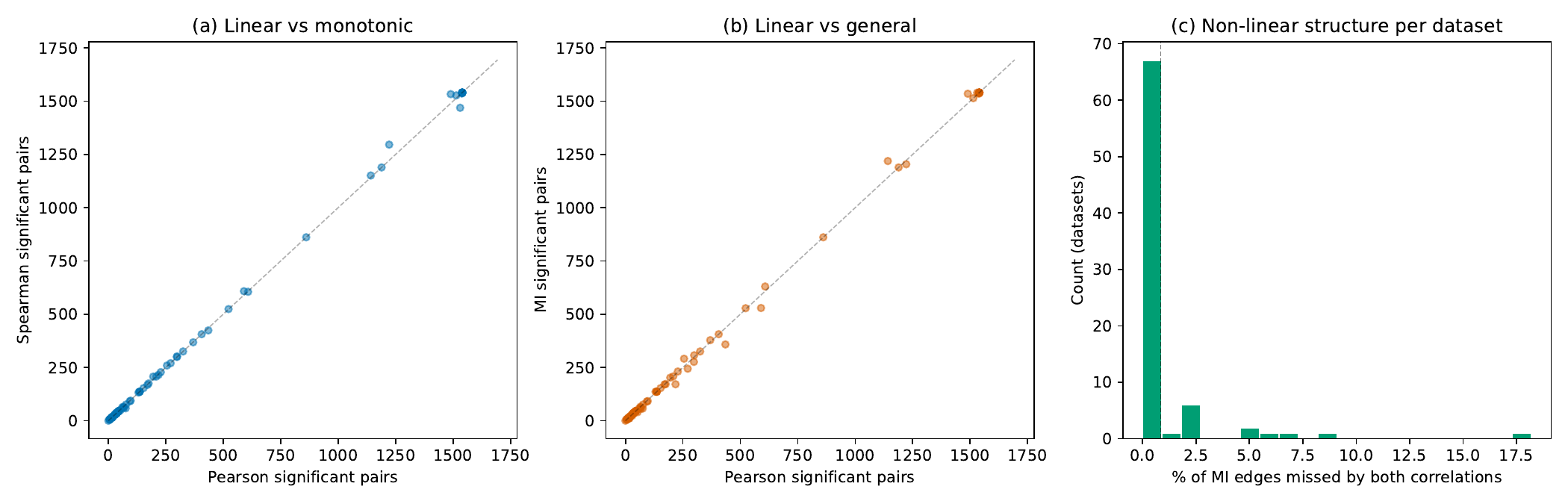}
\caption{\apacap{Comparison of bivariate dependency measures across 80 clinical and psychometric datasets}{(a) Number of significant pairs detected by Pearson vs.\ Spearman correlation per dataset; near-perfect agreement indicates negligible non-linear monotonic structure. (b) Number of significant pairs detected by Pearson correlation vs.\ mutual information; MI generally detects fewer pairs, consistent with discretisation-induced information loss. (c) Distribution of the percentage of MI-significant pairs not detected by either correlation measure; the vast majority of datasets show fewer than 2.5\% non-linear-only dependencies.}}
\label{fig:dependency_comparison}
\end{figure}

Figure~\ref{fig:dependency_comparison} summarises the results. At the level of statistical significance, Pearson and Spearman correlation agreed almost perfectly (panel a): fewer than 1\% of significant pairs were detected by one but not the other, confirming that monotonic-but-non-linear dependencies are negligible in these data. Mutual information detected fewer significant pairs than either correlation measure (panel b), consistent with discretisation-induced information loss rather than with MI capturing additional non-linear structure. Across all 80 datasets, only 193 of 30{,}467 MI-significant pairs (0.6\%) were detected by MI but not by either correlation measure (panel c), indicating that non-linear dependencies invisible to correlation are rare in clinical and psychometric datasets with ordinal response scales.

These results suggest that the linearity assumption underlying partial correlation networks is largely met in practice for ordinal symptom items. The primary motivation for information-theoretic methods in this context is therefore not the detection of non-linear pairwise associations, but the decomposition of multivariate dependence into redundant, unique, and synergistic components---a distinction that correlation-based methods cannot provide regardless of functional form.

\section{Cross-cohort Comparison of Information-Theoretic PHQ-9 Networks}
\label{app:cmi_pid_comparison}

\begin{figure}
    \centering
    \begin{subfigure}[t]{0.48\textwidth}
        \centering
        \includegraphics[width=\linewidth]{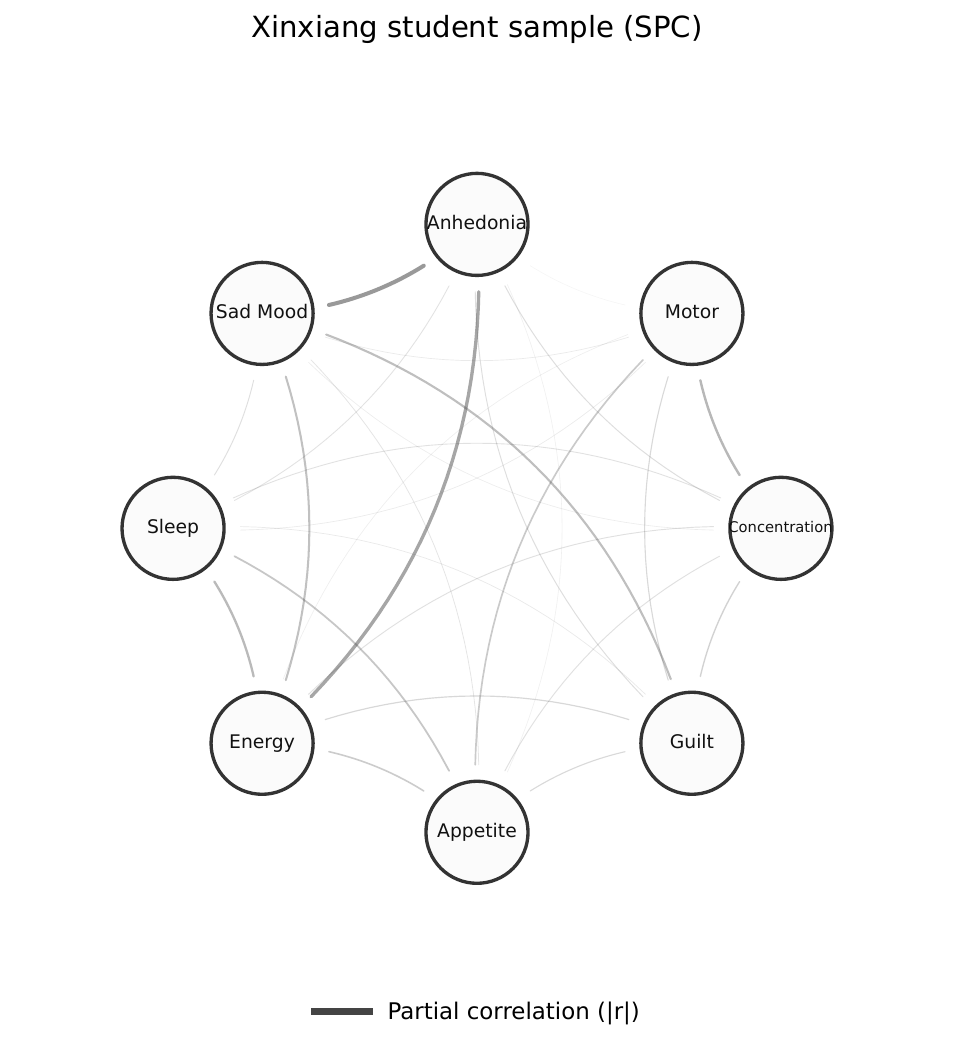}
        \caption{Pearson Partial Correlation Network}
        \label{fig:xinxiang_pearson_pc}
    \end{subfigure}
    \hfill
    \begin{subfigure}[t]{0.48\textwidth}
    \centering
    \includegraphics[width=\linewidth]{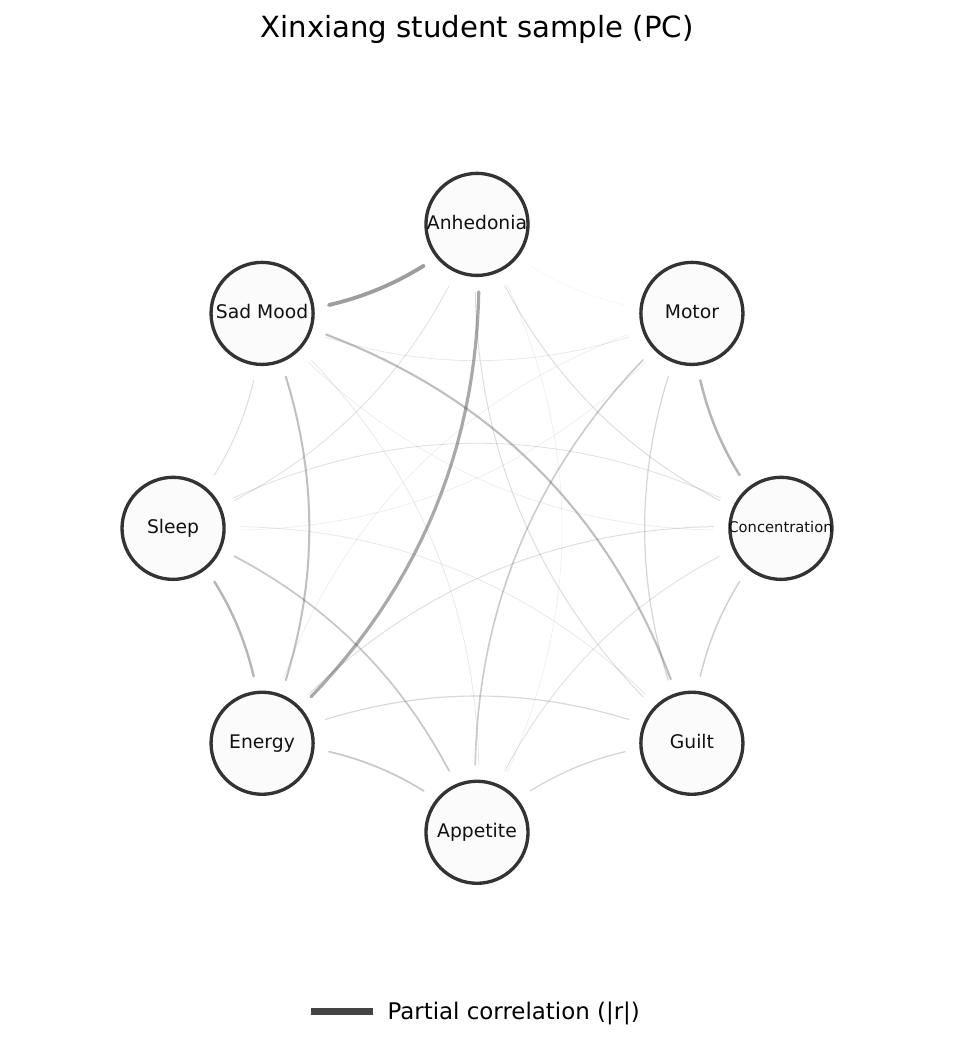}
    \caption{Spearman Partial Correlation Network}
    \label{fig:xinxiang_spearman_pc}
\end{subfigure}
        \begin{subfigure}[t]{0.48\textwidth}
        \centering
        \includegraphics[width=\linewidth]{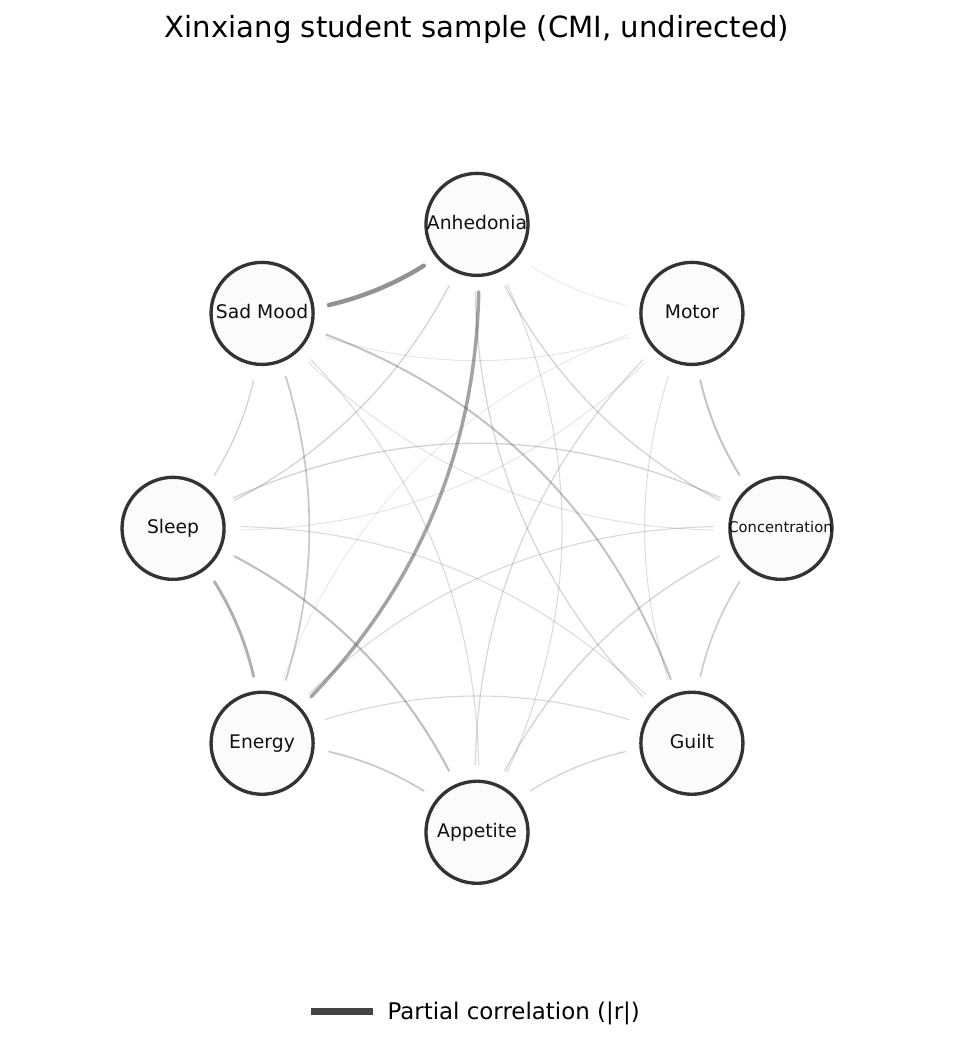}
        \caption{Conditional Mutual Information Network}
        \label{fig:xinxiang_cmi_network}
    \end{subfigure}
    \caption{\apacap{Xinxiang student sample PHQ-9 networks under three association measures}{Three measures that progressively relax modelling assumptions of linearity and monotonicity, each leading to a similar network structure: (a) Pearson partial correlation, (b) Spearman partial correlation, (c) conditional mutual information.}}
    \label{fig:xinxiang_PC_PC_CMI}
\end{figure}

\begin{figure}
    \centering
    \begin{subfigure}[t]{0.48\textwidth}
        \centering
        \includegraphics[width=\linewidth]{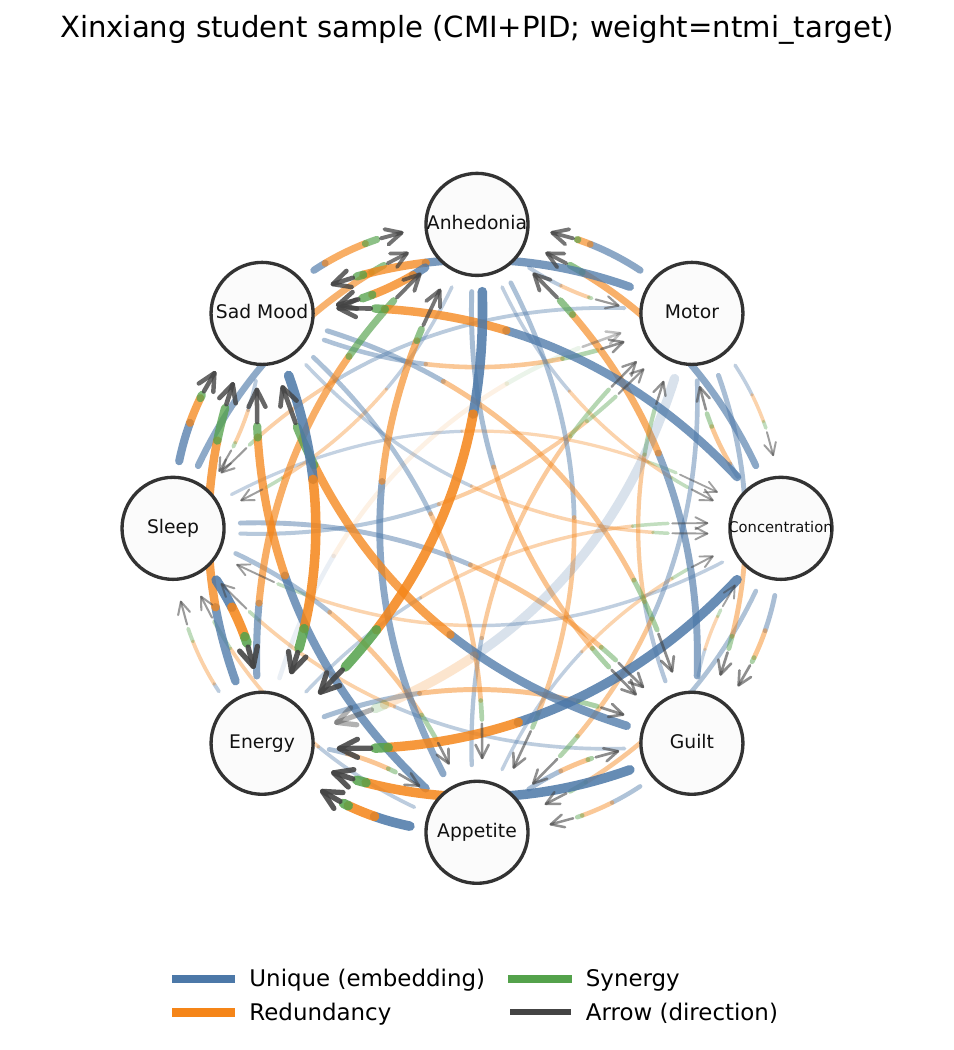}
        \caption{Full ePID view (all four atoms)}
        \label{fig:Xinxiang_Two_PIDs_withUnique}
    \end{subfigure}
    \hfill
    \begin{subfigure}[t]{0.48\textwidth}
        \centering
        \includegraphics[width=\linewidth]{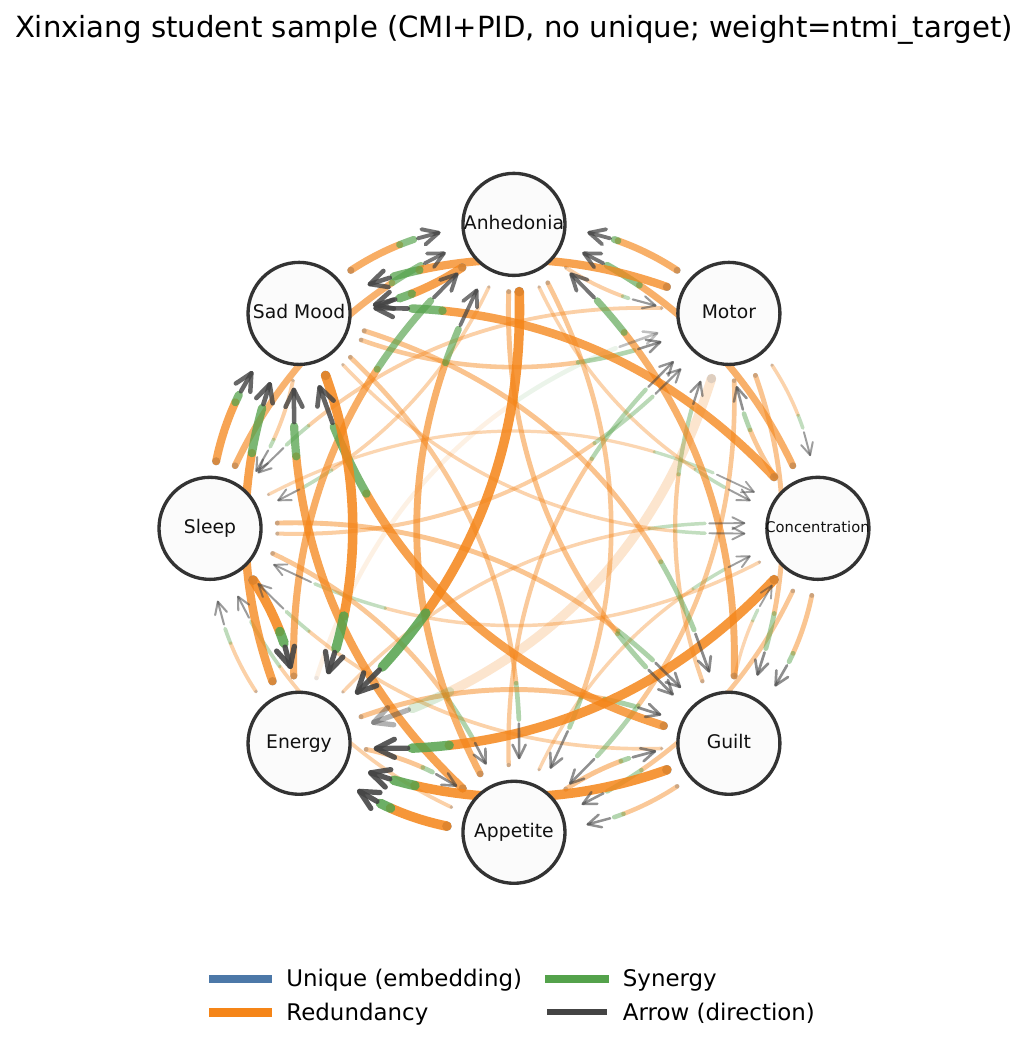}
        \caption{No-remainder-unique view (redundancy and synergy only)}
        \label{fig:Xinxiang_Two_PIDs_noUnique}
    \end{subfigure}
\caption{\apacap{Two ePID network views in the Xinxiang student sample}{Directed edges represent ordered source$\rightarrow$target relations.
\textbf{(a) Full ePID view:} each edge is partitioned into remainder-unique (blue), redundancy (orange), and synergy (green) atoms from the two-source PID of $(X_i, E_{i\to k}; X_k)$; the terminal black segment indicates direction.
\textbf{(b) No-remainder-unique view:} the remainder-unique atom is removed and the remaining redundancy and synergy atoms are rescaled, highlighting dependence that involves the focal source beyond remainder-only predictability.
Edge thickness encodes target-normalised total information $I(X_i, E_{i\to k}; X_k)/H(X_k)$.}}
    \label{fig:Xinxiang_Two_PIDs}
\end{figure}

\begin{figure}
    \centering
    \begin{subfigure}[t]{0.48\textwidth}
        \centering
        \includegraphics[height=6cm]{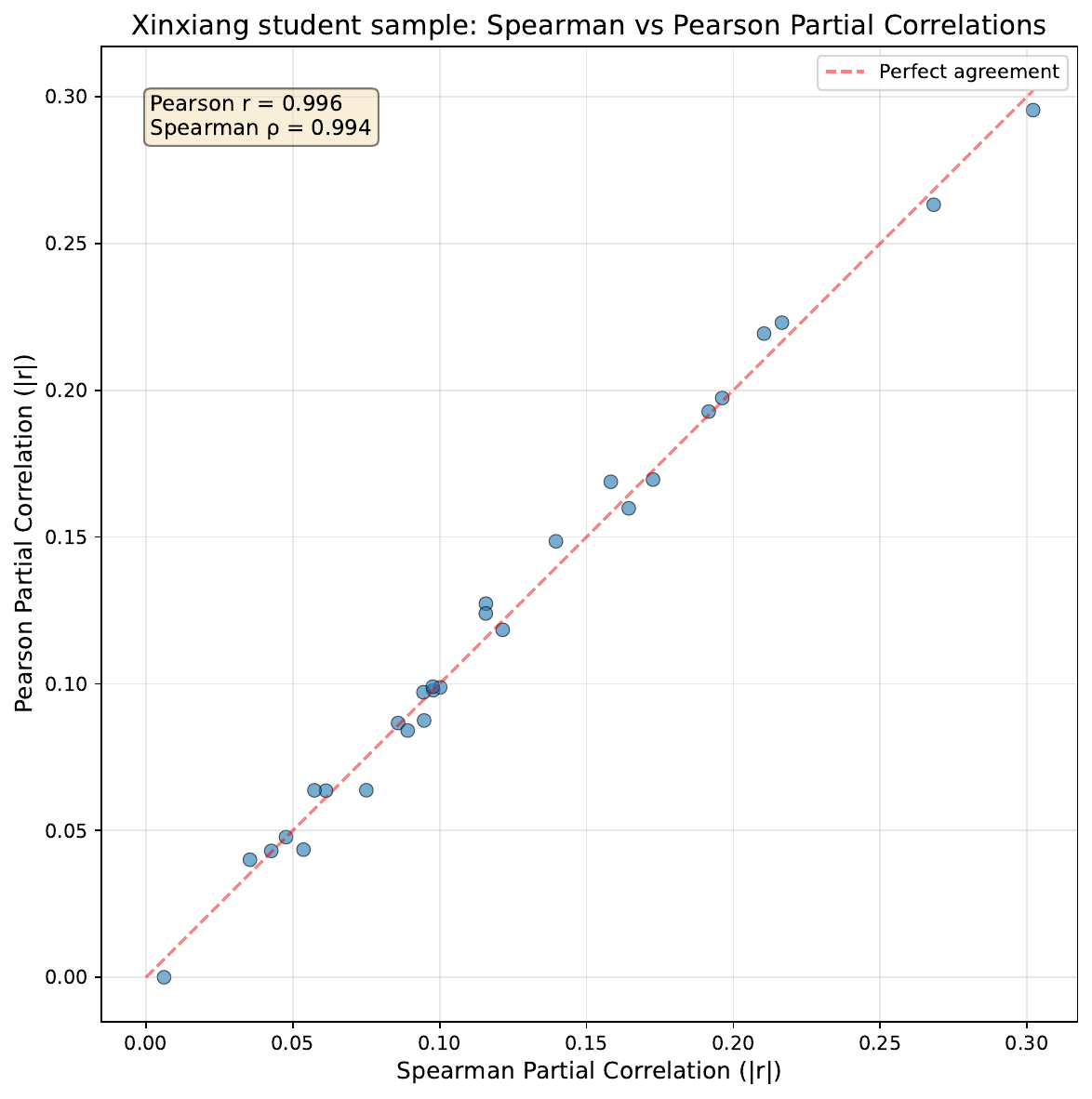}
        \caption{Edge-wise scatter comparison}
        \label{fig:xinxiang_pc_scatter}
    \end{subfigure}
    \hfill
    \begin{subfigure}[t]{0.48\textwidth}
        \centering
        \includegraphics[height=6cm]{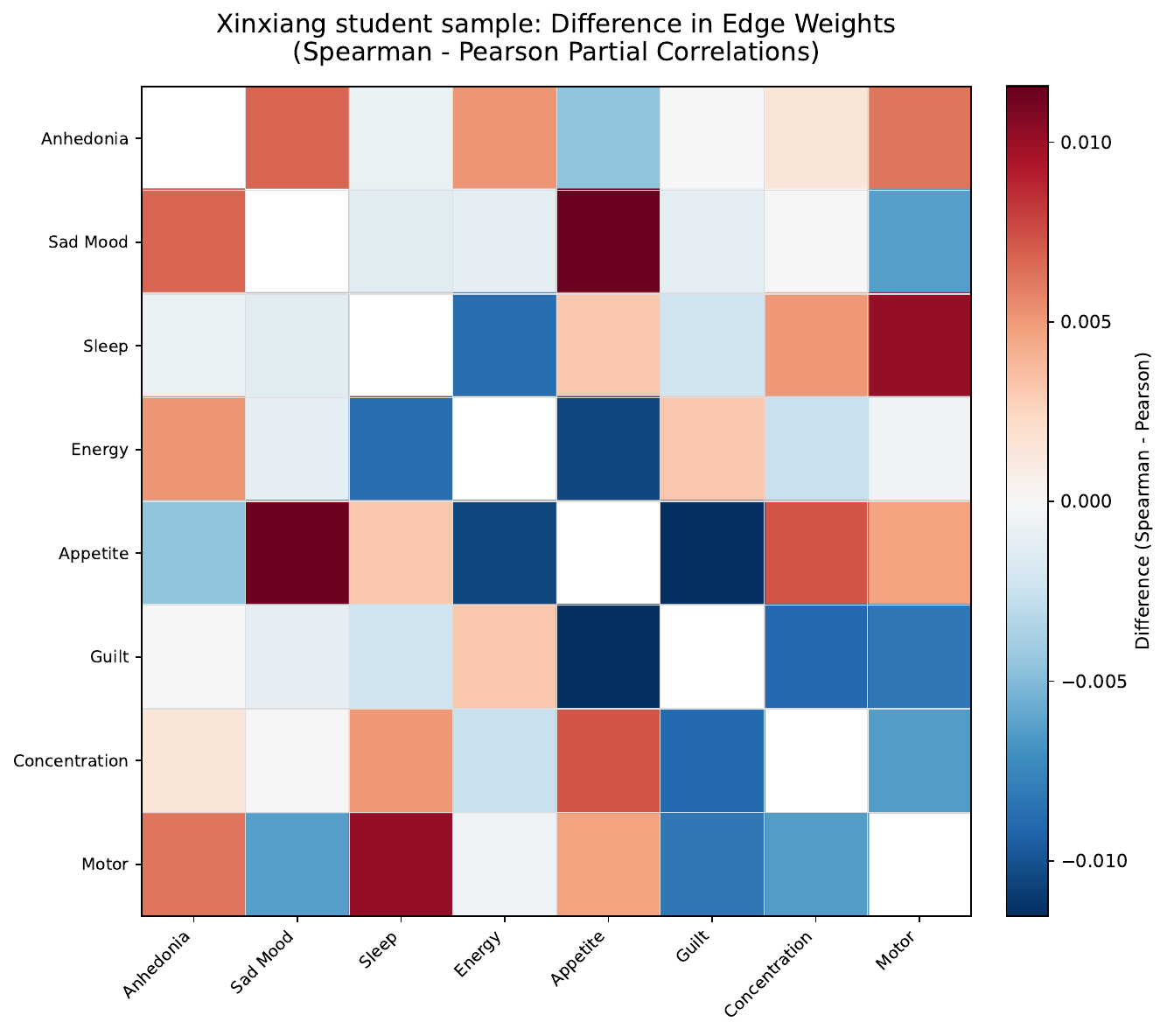}
        \caption{Edge-wise difference heatmap}
        \label{fig:xinxiang_pc_diff_heatmap}
    \end{subfigure}
\caption{\apacap{Agreement between partial Pearson and partial Spearman correlation networks in the Xinxiang student sample}{Partial Pearson correlations (PPC) and Spearman partial correlations (SPC) were computed for each symptom pair while conditioning on all remaining PHQ-9 symptoms (suicidality excluded).
\textbf{(a)} Edge-wise comparison of absolute partial correlations ($|r|$) shows near-identical weighting across pairs (Pearson $r=0.997$; Spearman $\rho=0.995$).
\textbf{(b)} Edge-wise differences (SPC$-$PPC) are small across the matrix, indicating that rank-based partial correlation and linear partial correlation yield highly similar conditional-association structures in this setting.}}
    \label{fig:PPC_vs_PSP}
\end{figure}

\begin{figure}[ht]
    \centering
    \includegraphics[width=0.72\linewidth]{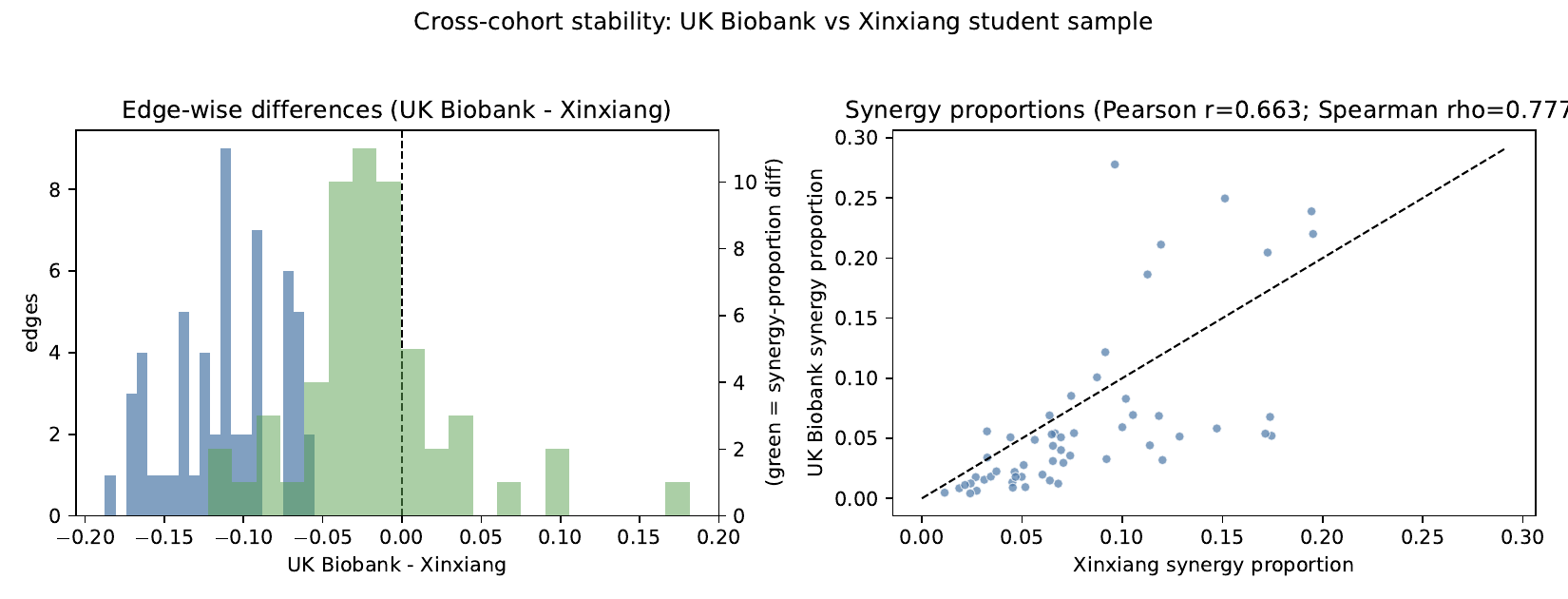}
\caption{\apacap{Cross-dataset comparison of information-theoretic dependence strength and decomposition stability (UK Biobank vs the Xinxiang student sample)}{For each directed source$\to$target symptom pair (eight PHQ-9 symptoms; suicidality excluded), we compare total mutual information (TMI) and the \emph{synergy proportion} (synergy/TMI) derived from the embedding-based PID framework. Edge-wise differences (UK Biobank $-$ Xinxiang) show systematically larger dependence strengths in UK Biobank, whereas synergy-proportion differences are concentrated near zero. Synergy proportions show substantial cross-dataset agreement (Pearson $r=0.663$; Spearman $\rho=0.777$), indicating that the \emph{composition} of dependence is more stable across datasets than the overall dependence magnitude.}}
\label{fig:cross_cohort_stability}
\end{figure}

Figure~\ref{fig:cross_cohort_stability} quantifies cross-dataset agreement in dependence strength and PID composition. Edge-wise differences in TMI are systematically positive (reflecting larger absolute dependence in UK Biobank), whereas synergy-proportion differences are concentrated near zero, and the synergy proportions show substantial cross-dataset agreement (Pearson $r = 0.663$; Spearman $\rho = 0.777$).

\subsection{Edge definition and comparability across cohorts}
We compared information-theoretic symptom networks estimated separately in two datasets using the same eight PHQ-9 symptom variables. For each symptom pair $(X,Y)$, we estimated the conditional mutual information (CMI) $I(X;Y\mid \mathbf{Z})$, where $\mathbf{Z}$ denotes the remaining six symptoms. The ``significant edge'' sets reported below correspond to the cohort-specific CMI screening procedure exported in the significant-edge files. Although CMI is symmetric in $X$ and $Y$ given the same conditioning set, we represent edges as directed ($X\to Y$) for two reasons: (i) to align with the directional partial information decomposition (PID) that is defined for a designated target variable, and (ii) to permit visualisation of asymmetries in the PID atoms between the two orientations. Accordingly, edge direction in these plots should not be interpreted as causal direction.

Because the cohorts differed in marginal response distributions, their marginal entropies $H(X)$ also differed across items. Since CMI in bits is upper-bounded by target uncertainty (e.g., $I(X;Y\mid \mathbf{Z})\leq H(Y)$), absolute thresholds expressed in bits are not directly comparable across cohorts when $H(Y)$ differs. To improve cross-cohort comparability, we therefore implemented an entropy-normalised effect size,
\begin{equation}
\mathrm{nCMI}_{X\rightarrow Y} \;=\; \frac{I(X;Y\mid \mathbf{Z})}{H(Y)},
\end{equation}
which can be interpreted as the fraction of the target's marginal uncertainty associated with the source after conditioning on the other symptoms. This normalisation is conservative because $H(Y)$ upper-bounds $H(Y\mid \mathbf{Z})$.

\subsection{Do cohorts yield the same significant CMI edges?}
At the level of statistical significance alone (i.e., without additional pruning), the UK Biobank cohort exhibited statistically significant CMI for all $56$ directed symptom pairs among eight nodes, whereas the Xinxiang student sample exhibited $54$ significant directed edges. Thus, the Xinxiang student sample's significant edge set was a strict subset of UK Biobank's. Table~\ref{tab:biobank_only_edges} lists the $2$ directed edges that were significant in UK Biobank but not in the Xinxiang student sample; equivalently, the Xinxiang student sample contained all remaining directed edges.

\begin{table}[!ht]
\centering
\caption{Directed CMI edges that were statistically significant in UK Biobank but not in the Xinxiang student sample (2 edges). UK Biobank contained all 56 directed symptom pairs; the Xinxiang student sample contained the complementary set of 54 edges.}
\label{tab:biobank_only_edges}
\begin{tabular}{ll}
\toprule
Source & Target \\
\midrule
Energy & Motor \\
Motor & Energy \\
\bottomrule
\end{tabular}
\end{table}

\subsection{Pruning rules for visualisation and ``clinical relevance''}
Statistical significance alone can yield dense graphs (notably in UK Biobank). To obtain readable figures while maintaining transparency, we evaluated several simple pruning rules applied \emph{after} significance filtering. Table~\ref{tab:edge_overlap_pruning} summarizes the number of retained edges in each cohort and the overlap between cohorts under: (i) an absolute cutoff on raw CMI (CMI$\geq 0.05$ bits), and (ii) rank-based rules computed on $\mathrm{nCMI}$ (top-10 edges overall, top quartile [top 25\%], and top-3 incoming edges per target). The absolute CMI cutoff retained more edges in the Xinxiang student sample than in UK Biobank, consistent with the datasets' differing entropy scales. In contrast, rank-based rules on $\mathrm{nCMI}$ produced more comparable densities and larger overlaps, thereby supporting interpretable cross-dataset comparisons.

\begin{table}[!ht]
\centering
\begin{threeparttable}
\caption{Overlap of statistically significant CMI edges across cohorts under alternative pruning rules. Rank-based rules use entropy-normalized CMI, $\mathrm{nCMI}=I(X;Y\mid\mathbf{Z})/H(Y)$. The Jaccard index is $|E_{\cap}|/|E_{\cup}|$ for the directed edge sets.}
\label{tab:edge_overlap_pruning}
\begin{tabular}{lrrrrr}
\toprule
Pruning rule & $|E_{\mathrm{Xinxiang}}|$ & $|E_{\mathrm{Biobank}}|$ & $|E_{\cap}|$ & $|E_{\cup}|$ & Jaccard \\
\midrule
All significant (FDR) & 54 & 56 & 54 & 56 & 0.96 \\
CMI $\geq 0.05$ bits & 10 & 6 & 6 & 10 & 0.60 \\
Top 10 by $\mathrm{nCMI}$ & 10 & 10 & 6 & 14 & 0.43 \\
Top 25\% by $\mathrm{nCMI}$ & 14 & 14 & 9 & 19 & 0.47 \\
Top 3 per target by $\mathrm{nCMI}$ & 24 & 24 & 17 & 31 & 0.55 \\
\bottomrule
\end{tabular}
\end{threeparttable}
\end{table}

\subsection{Depression-status contrast: all significant edges}
\label{app:dep_vs_nondep_alledges}

\begin{figure}[htbp]
\centering
\begin{tikzpicture}
  \node[anchor=south west,inner sep=0] (img) at (0,0)
    {\includegraphics[width=\textwidth]{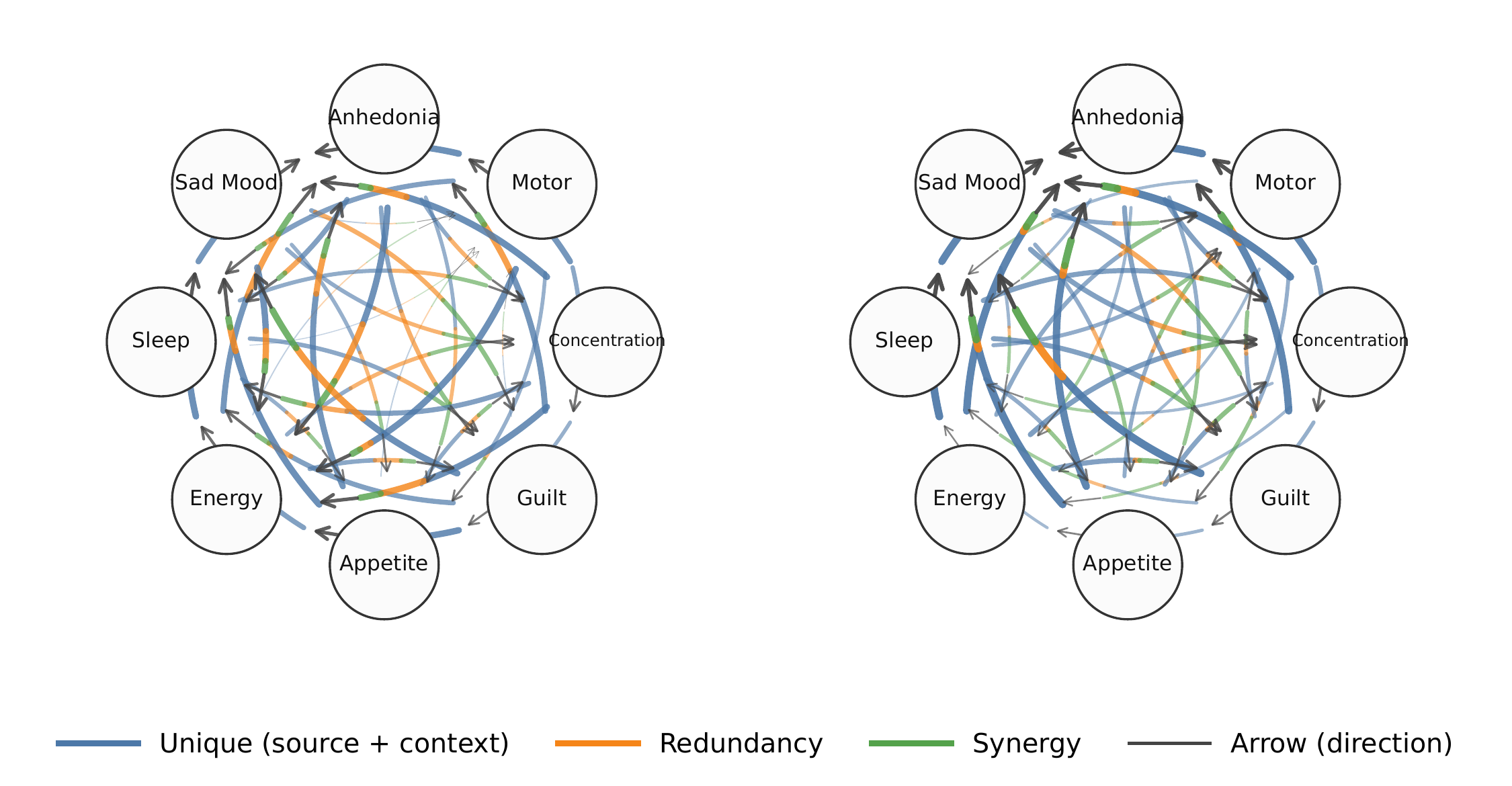}};
  \node[anchor=north west,font=\bfseries] at ([xshift=3pt,yshift=-3pt]img.north west) {(a)};
  \node[anchor=north west,font=\bfseries] at ([xshift=3pt,yshift=-3pt]img.north) {(b)};
\end{tikzpicture}
\caption{\apacap{ePID networks in non-depressed versus depressed UK Biobank respondents (all significant edges)}{Companion to main-text Fig.~\ref{fig:dep_vs_nondep}, showing all 56 significant directed edges rather than the top-three-per-target subset. \textbf{(a)} non-depressed (PHQ-9 total $<10$); \textbf{(b)} depressed (PHQ-9 total $\geq 10$); equal-$N$ matched at $n=8{,}879$ per arm, five-seed average, eight PHQ-9 symptoms (suicidality excluded). Segment colours (unique = blue, redundancy = orange, synergy = green) and the shared total-mutual-information thickness scale are as in Fig.~\ref{fig:dep_vs_nondep}. Over all 56 edges the redundancy-to-unique/synergy shift is larger than for the pruned display (mean shares: redundancy $0.26\rightarrow0.11$, unique $0.60\rightarrow0.66$, synergy $0.14\rightarrow0.23$). As in the main figure, the absolute synergy share is $N$-dependent and the panels should be read as a matched-$N$ contrast, not an absolute level.}}
\label{fig:dep_vs_nondep_alledges}
\end{figure}

Figure~\ref{fig:dep_vs_nondep_alledges} reproduces the depression-status contrast of main-text Figure~\ref{fig:dep_vs_nondep} over all 56 significant directed edges rather than the top-three-per-target subset. The redundancy-to-unique/synergy shift is larger on the full edge set (mean shares: redundancy $0.26 \rightarrow 0.11$, unique $0.60 \rightarrow 0.66$, synergy $0.14 \rightarrow 0.23$), consistent with the pruned main-text display but computed from the complete significant-edge network.

\FloatBarrier
\FloatBarrier
\section{IRI cross-subscale synergy and redundancy patterns}
\label{app:iri_bar_charts}

\begin{figure}[H]
  \centering
  \includegraphics[width=\textwidth]{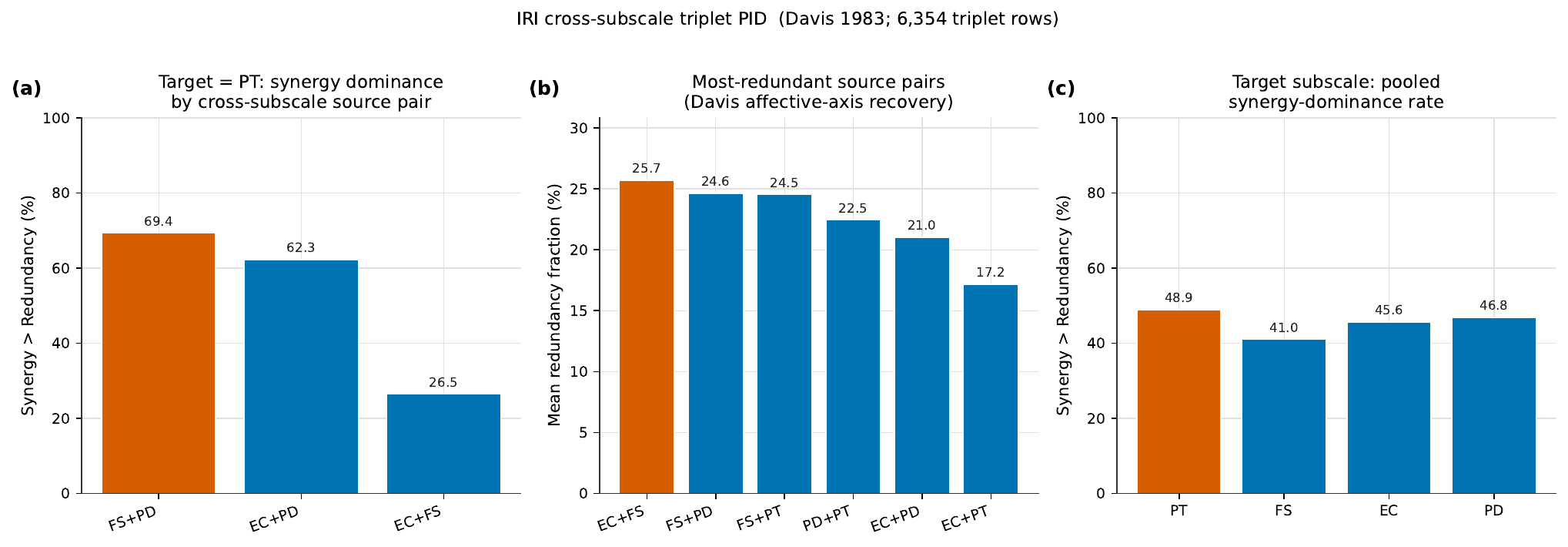}
  \caption{\apacap{IRI cross-subscale synergy and redundancy patterns}{Each panel is computed over informative triplets (total mutual information $I(X_i, X_j; X_k) \geq 0.01$ bits) in which the source pair and target item belong to three distinct IRI subscales. \textbf{(a)} Synergy-dominance rate (Synergy $>$ Redundancy) for triplets targeting a Perspective-Taking item, by cross-subscale source pair; FS$+$PD tops the ranking at 69.4\%, consistent with the main-text claim that mature cognitive empathy is informed synergistically by the affective extremes. \textbf{(b)} Mean redundancy fraction per source pair, pooled across all target subscales; EC$+$FS tops the ranking at 25.7\%, and all three pairs containing Fantasy rank above all three pairs without it. \textbf{(c)} Synergy-dominance rate per target subscale, pooled across all cross-subscale source pairs; PT ranks highest at 48.9\%, ahead of PD (46.8\%), EC (45.6\%) and FS (41.0\%), consistent with developmental accounts in which cognitive empathy emerges later than its affective counterpart and comes to integrate it~\citep{decety2021emergence}. PT $=$ Perspective Taking; FS $=$ Fantasy Scale; EC $=$ Empathic Concern; PD $=$ Personal Distress.}}
  \label{fig:iri_synergy_bars}
\end{figure}

Figure~\ref{fig:iri_synergy_bars} provides a visual summary of the cross-subscale triplet PID analysis whose headline percentages appear in main-text Section~\ref{subsec:iri_results}. The three panels report (a)~synergy-dominance by source pair for Perspective-Taking targets, (b)~redundancy ranking across source pairs pooled across targets, and (c)~synergy-dominance rate per target subscale. Percentages match those quoted in the main text under the same informative-pair filter ($I(X_i, X_j; X_k) \geq 0.01$ bits) and the same cross-subscale criterion (one item from each of three distinct subscales as $X_1$, $X_2$, target).

\FloatBarrier
\section{Illustrative example: stress, sleep, and concentration}
\label{app:stylised_example}

To illustrate how PC, CMI, and PID provide complementary information, we construct a hypothetical three-variable system: stress ($X_2$), sleep problems ($X_3$), and concentration ($X_1$), each on a five-level ordinal scale ($\{0,1,2,3,4\}$). Concentration closely tracks sleep quality under typical conditions ($X_1 = X_3$ when $X_3 \in \{0,1,2,3\}$), but when sleep is extremely poor ($X_3 = 4$), concentration depends non-monotonically on stress\footnote{Note that this equation can be rewritten with use of the modulo operator, but we omit this for ease of explanation.}:
\[
X_1 =
\begin{cases}
X_3, & \text{if $X_3 \in \{0,1,2,3\}$ (normal sleep)}, \\[6pt]
0, & \text{if $X_3=4,\,X_2=0$ (very poor sleep, no stress)}, \\[2pt]
4, & \text{if $X_3=4,\,X_2=1$ (very poor sleep, mild stress)}, \\[2pt]
3, & \text{if $X_3=4,\,X_2=2$ (very poor sleep, moderate stress)}, \\[2pt]
2, & \text{if $X_3=4,\,X_2=3$ (very poor sleep, high stress)}, \\[2pt]
1, & \text{if $X_3=4,\,X_2=4$ (very poor sleep, extreme stress)}.
\end{cases}
\]

We simulate $10{,}000$ samples and estimate the network using partial correlations, CMI, and PID (Figure~\ref{stylised-example}). The three methods reveal progressively richer structure:
\begin{itemize}
    \item \textbf{Partial correlation:} Detects only the $X_1$--$X_3$ edge. The stress--concentration link is missed because it is non-monotonic and emerges through a synergistic interaction with sleep.
    \item \textbf{CMI:} Detects both the $X_1$--$X_2$ and $X_1$--$X_3$ edges, revealing that stress influences concentration conditionally---its effect appears only when sleep is very poor.
    \item \textbf{PID} (using $I_{\min}$, a redundancy measure formalised in Section~\ref{subsec:pid_two_source})\textbf{:} Decomposes the joint influence into $71.5\%$ unique to sleep, $21.1\%$ synergistic between stress and sleep, $7.2\%$ redundant, and $0.1\%$ unique to stress. The synergy fraction quantifies precisely the interaction that PC misses and CMI detects only indirectly.
\end{itemize}

\begin{figure}[h!]
\centering
\begin{tikzpicture}[node distance=3cm, thick]
\tikzstyle{symptom} = [circle, draw, minimum size=1.2cm]

\node[symptom] (X1) {Concentration ($X_1$)};
\node[symptom, right of=X1, xshift=4cm] (X2) {Stress ($X_2$)};
\node[symptom, below of=X1, yshift=-3cm, xshift=2cm] (X3) {Sleep ($X_3$)};

\draw[blue, dotted, thick, bend right=35]
    (X1) to node[midway, left, blue, font=\tiny]{PC (0.709)} (X3);

\draw[red, dashed]
    (X1) -- node[midway, above, red, font=\tiny]{CMI (0.470 bits)} (X2);
\draw[red, dashed]
    (X1) -- node[midway, right, red, font=\tiny]{CMI (2 bits)}(X3);

\draw[green!50!black, thick, bend left=25]
    (X2) to node[midway, above, font=\tiny, green!50!black, text width=2.5cm, align=center]
    {Unique: 0.1\% \\ (0.00 bits)} (X1);

\draw[green!50!black, thick, bend right=35] (X3) to (X1);

\node[green!50!black, font=\tiny, text width=2.5cm, align=center]
    at ($(X3)!0.6!(X1) + (2.3,-0.9)$)
    {Unique: 71.5\% \\ (1.54 bits)};

\node (mid) at ($(X2)!0.5!(X3)$) {};
\draw[green!70!black, thick, bend left=20] (X2) to (mid.center);
\draw[green!70!black, thick, bend right=20] (X3) to (mid.center);
\draw[green!70!black, thick] (mid.center) -- (X1);

\node[green!70!black, font=\tiny, text width=2.5cm, align=center] at ($(mid)!0.5!(X1) + (3,-1.8)$)
    {Synergy: 21.1\% \\ (0.46 bits)};

\node[green!70!black, font=\tiny, text width=2.5cm, align=center] at ($(mid)!0.5!(X1) + (3,-2.3)$)
    {Redundancy: 7.2\% \\ (0.155 bits)};

\end{tikzpicture}
\caption{\apacap{The same three-variable system as seen by three methods}{PC (blue, dotted) only recovers the linear $X_1$--$X_3$ edge; CMI (red, dashed) additionally detects the conditional $X_1$--$X_2$ dependence; PID (green, solid) further splits the joint information (2.16 bits total) into unique, redundant, and synergistic atoms, making explicit the synergy that PC misses and CMI detects only indirectly. This example is constructed for methodological demonstration and does not represent an empirically identified mechanism.}}
\label{stylised-example}
\end{figure}

The deterministic structure of this example exaggerates the effect relative to empirical data. Nonetheless, the decomposition illustrates a general principle: when synergy is high, predictive models that accommodate interactions (e.g., random forests, which achieved perfect accuracy here) substantially outperform additive models (adjusted $R^2 = 0.45$ for linear regression). The information-theoretic decomposition thus provides principled guidance on when interaction terms are needed.

\section{Robustness of the ACIB embedding to its hyperparameters}
\label{app:acib_sensitivity}

\subsection{Question}
The embedding step at the heart of ePID (the agglomerative conditional information bottleneck, ACIB) has three free hyperparameters: the conditional-mutual-information \emph{loss tolerance} (default $5\%$), the embedding cardinality cap $K_{\max}$ (set to $12$ in the main analyses), and a Dirichlet \emph{smoothing} constant $\alpha$ (default $0.5$). A reviewer will reasonably ask whether the decompositions we report are robust to these settings, or artefacts of a particular tuning. This appendix answers that with two complementary sensitivity analyses, and in doing so also justifies the choice to embed only $n_{\mathrm{sources}}=5$ remainder symptoms rather than the full network.

\subsection{What we did}
\paragraph{Fidelity against ground truth (synthetic).}
On the synthetic networks used to validate the embedding (which have known, exact ground-truth PID atoms), we re-ran ACIB across a $3\times3$ grid of loss tolerance $\in\{2\%,5\%,10\%\}$ and $K_{\max}\in\{6,10,14\}$, plus a one-dimensional slice over the smoothing constant $\alpha\in\{0.1,0.5,1.0\}$. For each setting we recorded the total absolute error of the recovered two-source atoms against ground truth (measure $I_{\mathrm{mmi}}$, $N=5$ sources), the realised embedding cardinality, and the retained conditional mutual information (the information-bottleneck ``knee'').

\paragraph{Stability of the empirical signal (cohorts).}
On the two empirical instruments (the Xinxiang student sample and the IRI empathy network) we recomputed the per-edge synergy fraction across the same grid. We additionally contrasted the $n_{\mathrm{sources}}=5$ embedding protocol against a \emph{full-context} embedding that folds all remaining items into a single variable.

\subsection{What we found}
\begin{figure}[h]
\centering
\begin{minipage}{0.48\textwidth}\centering
\includegraphics[width=\linewidth]{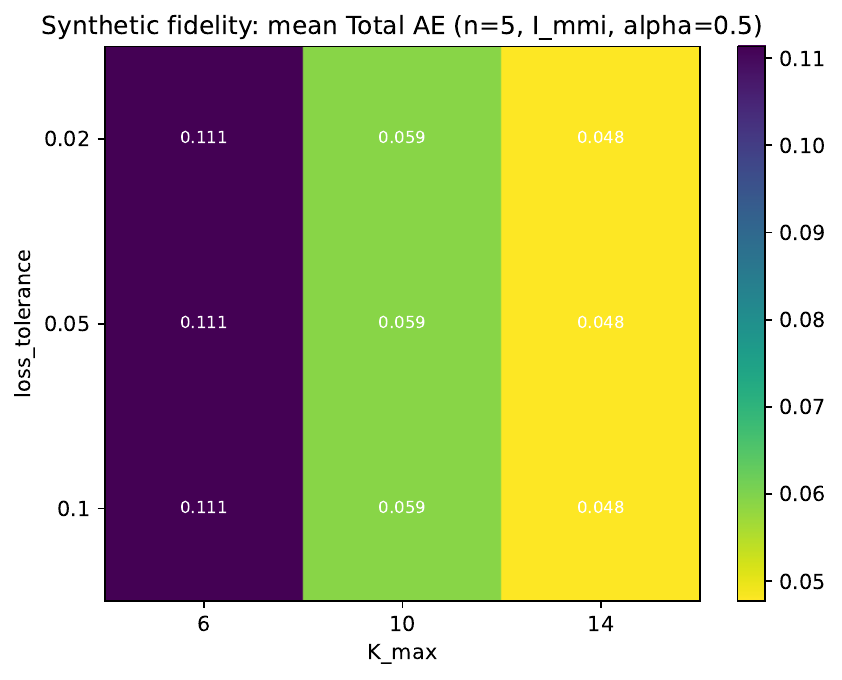}
\caption{Fidelity surface (mean total absolute error). $K_{\max}$ is the only active knob.}
\label{fig:acib_fidelity}
\end{minipage}\hfill
\begin{minipage}{0.48\textwidth}\centering
\includegraphics[width=\linewidth]{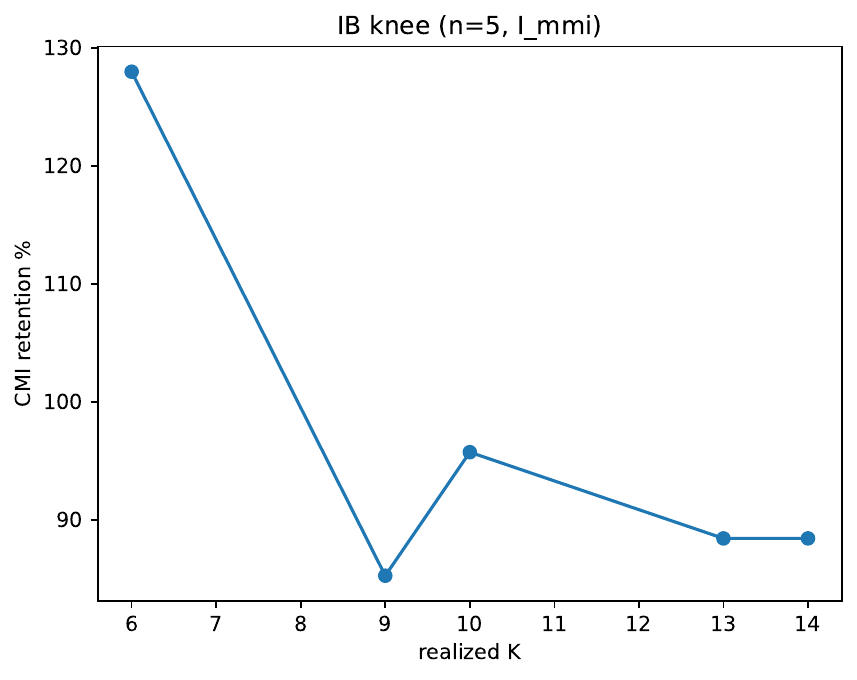}
\caption{Information-bottleneck retention plateaus near a realised cardinality of ten.}
\label{fig:acib_knee}
\end{minipage}
\end{figure}

\paragraph{(i) Fidelity depends only on $K_{\max}$, and the default is in the stable region.}
The approximation error is \emph{insensitive} to the loss tolerance: it is identical to four decimal places across $2\%$, $5\%$, and $10\%$ at every $K_{\max}$, and to the smoothing constant $\alpha$. The only consequential knob is $K_{\max}$: error falls monotonically as more clusters are allowed, with clearly diminishing returns (Table~\ref{tab:acib_fidelity}, Fig.~\ref{fig:acib_fidelity}). The information-bottleneck retention curve plateaus at a realised cardinality of roughly ten (Fig.~\ref{fig:acib_knee}). The cap used in the main analyses, $K_{\max}=12$, therefore sits above the realised cardinality and within this stable region: it is statistically indistinguishable from the neighbouring grid values $K_{\max}=10$ and $K_{\max}=14$, which differ from one another by only $0.011$ bits.

\begin{table}[h]
\centering
\caption{Mean total absolute error of the recovered atoms against ground truth ($N=5$, $I_{\mathrm{mmi}}$, $\alpha=0.5$). Rows (loss tolerance) are identical; columns ($K_{\max}$) carry all the variation.}
\label{tab:acib_fidelity}
\begin{tabular}{lccc}
\toprule
loss tolerance & $K_{\max}=6$ & $K_{\max}=10$ & $K_{\max}=14$ \\
\midrule
$2\%$  & 0.111 & 0.059 & 0.048 \\
$5\%$  & 0.111 & 0.059 & 0.048 \\
$10\%$ & 0.111 & 0.059 & 0.048 \\
\bottomrule
\end{tabular}
\end{table}

\paragraph{(ii) The empirical synergy contrast is not a hyperparameter artefact.}
Across the whole grid the per-edge synergy fraction is stable: $6.1$ to $6.8\%$ in the Xinxiang student sample and $25.5$ to $26.4\%$ in IRI. The headline contrast, namely depression networks redundancy-dominated and empathy networks synergy-dominated, therefore holds regardless of the embedding settings.

\begin{figure}[h]
\centering
\includegraphics[width=0.55\textwidth]{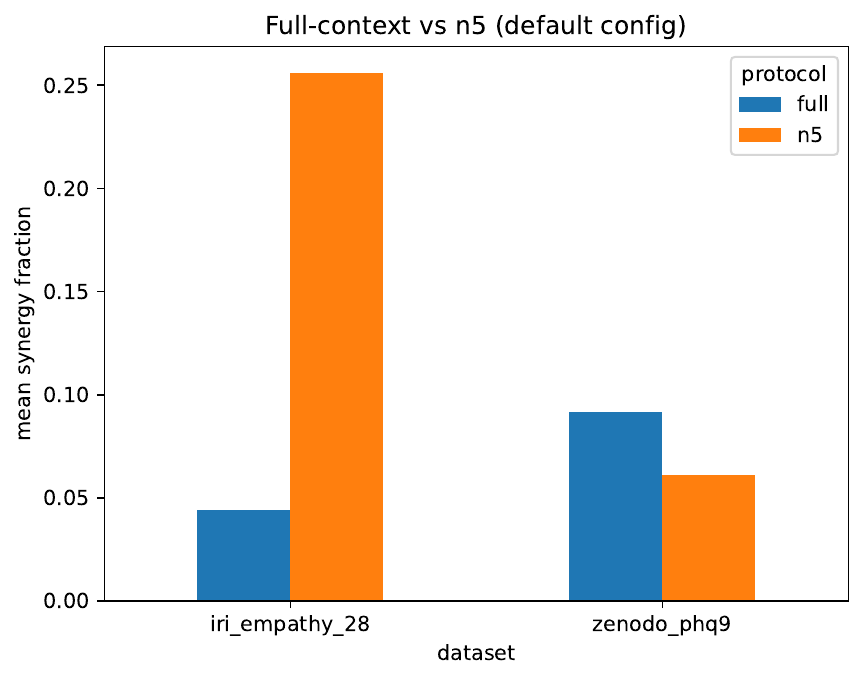}
\caption{Full-context versus $n_{\mathrm{sources}}=5$ synergy fraction. The 28-item IRI network saturates under full context (synergy collapses); the 9-item PHQ-9 does not.}
\label{fig:acib_saturation}
\end{figure}

\paragraph{(iii) Full-context embedding saturates, validating $n_{\mathrm{sources}}=5$.}
Folding \emph{all} remaining items into one embedding collapses the decomposition for the 28-item IRI network: synergy falls from $0.256$ under $n_{\mathrm{sources}}=5$ to $0.044$ under full context (Fig.~\ref{fig:acib_saturation}). For the 9-item PHQ-9 the change is modest ($0.061$ to $0.092$, no collapse). This is the expected saturation: when too many items are conditioned on at once, the residual conditional dependence approaches zero and the atoms degenerate into numerical noise. Limiting the embedding to a small remainder context avoids this regime.

\subsection{What we conclude}
The reported decompositions are robust to the ACIB hyperparameters. The loss tolerance and the smoothing constant have no measurable effect on fidelity; the only meaningful knob, $K_{\max}$, sits in the stable region at its main-analysis value of $12$, above the realised embedding cardinality. The empirical synergy estimates, and the redundancy-versus-synergy contrast between instruments, are stable across the grid. Finally, the saturation of the full-context embedding for large networks independently justifies the decision to embed only a bounded remainder context.

\subsection{Caveats}
\begin{itemize}
\item The synthetic sweep at $N=5$ was run for $I_{\mathrm{mmi}}$; the other measures ($I_{\min}$, $I_{\wedge}$) are covered at $N=3,4$ and in the full multi-dataset benchmark.
\item The cohort sensitivity used a modest sample of edges per configuration: it is a robustness and saturation \emph{demonstration}, not a precise re-estimate of the headline synergy levels (which are reported in the main text on the full samples).
\end{itemize}

\newpage

\bibliography{bibliography}


\end{document}